%% file: main_ARXIV.tex
\documentclass{article}

\usepackage{arxiv}
\usepackage[noblocks]{authblk}
\input{preamble}

\loadglsentries{acronyms.tex}

\title{Interpreting Temporal Graph Neural Networks \\ with Koopman Theory}
\author[1]{Michele Guerra}
\author[2]{Simone Scardapane}
\author[1,3]{Filippo Maria Bianchi}

\affil[1]{UiT The Arctic University of Norway, Department of Mathematics and Statistics, Norway.}
\affil[2]{Sapienza Università di Roma, Department of Information Engineering, Electronics and Telecommunications, Italy.}
\affil[3]{NORCE Norwegian Research Centre AS, Norway.}
\date{}

\begin{document}

\maketitle

\input{main_text}

\paragraph{Acknowledgments}
The authors gratefully acknowledge NVIDIA Corporation for the donation of two RTX A6000 that were used in this project.
This work was supported by the Norwegian Research Council projects 345017 \emph{RELAY: Relational Deep Learning for Energy Analytics}.

\bibliographystyle{abbrv}
\bibliography{sample}

\newpage
\appendix
\part*{Supplementary Materials}

\input{supplement}

\end{document}

%% file: preamble.tex
\usepackage{graphicx} 
\usepackage{amsmath}
\usepackage{amsthm}
\usepackage{amssymb}
\usepackage{bm}
\usepackage{xcolor}
\usepackage[nogroupskip,nonumberlist,acronym,toc=false,nopostdot=false]{glossaries-extra}
\setabbreviationstyle[acronym]{long-short}
\usepackage{caption}
\usepackage{subcaption}
\usepackage{hyperref}
\usepackage{booktabs}
\usepackage{multirow}
\usepackage{natbib}
\usepackage{standalone}
\usepackage{siunitx} 
\usepackage{cleveref} 
\usepackage{placeins} 
\usepackage[prependcaption]{todonotes}

\usepackage{xcolor}
\definecolor{darkgreen}{rgb}{0.0, 0.5, 0.0}
\definecolor{darkred}{rgb}{0.6, 0.0, 0.0}
\definecolor{gray}{rgb}{0.5, 0.5, 0.5}
\definecolor{myblue}{RGB}{0, 0, 180}
\definecolor{mygreen}{RGB}{0, 128, 0}
\definecolor{myred}{RGB}{196, 0, 0}
\usepackage{url}
\hypersetup{
    colorlinks=true,       
    linkcolor=myred,      
    citecolor=myblue,     
    filecolor=magenta,     
    urlcolor=mygreen         
}

\graphicspath{ {./images/} }

\theoremstyle{plain}
\newtheorem{theorem}{Theorem}[section]
\theoremstyle{definition}
\newtheorem{definition}[theorem]{Definition}

\newcommand{\timeidx}{t}
\newcommand{\stateidx}{d}
\newcommand{\sindyidx}{j}

\renewcommand{\cite}{\citep}

%% file: acronyms.tex
\newabbreviation{tg}{TG}{temporal graph}
\newabbreviation{stgnn}{STGNN}{spatiotemporal graph neural network}
\newabbreviation{gcrn}{GCRN}{graph convolutional recurrent network}
\newabbreviation{dcrnn}{DCRNN}{diffusion convolutional recurrent neural network}
\newabbreviation{gwn}{GWN}{graph WaveNet}
\newabbreviation{gcn}{GCN}{graph convolution network}
\newabbreviation{tcn}{TCN}{temporal convolution layer}
\newabbreviation{rnn}{RNN}{recurrent neural network}
\newabbreviation{lstm}{LSTM}{long short-term memory}
\newabbreviation{gru}{GRU}{gated recurrent unit}
\newabbreviation{dcl}{DCL}{diffusion convolutional layer}
\newabbreviation{mlp}{MLP}{multi-layer perceptron}
\newabbreviation{pca}{PCA}{principal component analysis}
\newabbreviation{dmd}{DMD}{dynamic mode decomposition}
\newabbreviation{ttdmd}{TT-DMD}{tensor-based dynamic mode decomposition}
\newabbreviation{svd}{SVD}{singular value decomposition}
\newabbreviation{sindy}{SINDy}{sparse identification of nonlinear dynamics}
\newabbreviation{msrc12}{MSRC-12}{Microsoft Research Cambridge-12}
\newabbreviation{auc}{AUC}{area under the curve}
\newabbreviation{mad}{MAD}{median absolute distance}
\newabbreviation{gnn}{GNN}{graph neural network}

%% file: main_text.tex
\begin{abstract}
    \Glspl{stgnn} have shown promising results in many domains, from forecasting to epidemiology. However, understanding the dynamics learned by these models and explaining their behaviour is significantly more difficult than for models that deal with static data.
    Inspired by Koopman theory, which allows a simple description of intricate, nonlinear dynamical systems, we introduce new explainability approaches for \glsentrylong{tg}s.
    Specifically, we present two methods to interpret the \gls{stgnn}'s decision process and identify the most relevant spatial and temporal patterns in the input for the task at hand.
    The first relies on \gls{dmd}, a Koopman-inspired dimensionality reduction method. The second relies on \gls{sindy}, a popular method for discovering governing equations of dynamical systems, which we use for the first time as a general tool for explainability. 
    On semi-synthetic dissemination datasets, our methods correctly identify interpretable features such as the times at which infections occur and the infected nodes. We also validate the methods qualitatively on a real-world human motion dataset, where the explanations highlight the body parts most relevant for action recognition.
\end{abstract}

\glsresetall

\section{Introduction}

Many complex phenomena can be described by the dynamics of items interacting with each other in space and time, leading to complex spatiotemporal relationships that are naturally modelled by \glspl{tg}~\cite{cini2023sparse}. Examples are roads and junctions in traffic dynamics~\cite{zhang2020spatio}, arms and legs during human motions~\cite{jain2016structural}, infections during social contacts~\cite{fritz2022combining}, social interactions during events~\cite{deng2019learning}, brain activity~\cite{chen2025explainable}, atmospheric events~\cite{marisca2024graph}, and many more.
\Glspl{gnn} have already proved effective on static graphs~\cite{wu2020comprehensive, khemani2024review}, but capturing both the spatial and temporal patterns in \glspl{tg} remains challenging. Recently, \glspl{stgnn} have emerged as powerful tools to handle this type of data~\cite{longa2023graph, micheli2022discrete, cini2023scalable, cini2024graph}.
Given their use in critical applications, model explainability for \glspl{stgnn} has become a primary concern~\cite{hassija2024interpreting}.

The field of explainability for static \glspl{gnn} has been very prolific in recent years, offering many methods that can be broadly categorised into factual and counterfactual approaches~\cite{kakkad2023survey, longa2025explaining}. 
The former approach aims to find important substructures in the input graph, for example, by employing perturbation methods such as GNNExplainer~\cite{ying2019gnnexplainer} or PGExplainer~\cite{luo2020parameterized, guerra2023explainability}. These are further classified into post hoc methods, which are applied to a trained \gls{gnn}, and self-interpretable models~\cite{spinelli2022meta, spinelli2023combining, azzolin2025beyond, azzolin2026gnn}.
Counterfactual methods, instead of identifying the most important subgraph, aim at finding the minimal change in the input graph that causes a change in the \gls{gnn}'s prediction, e.g.~CF-GNNExplainer~\cite{lucic2022cf}.
Explainability methods for \glspl{gnn} can be further classified into local and global approaches: the former, such as the aforementioned PGExplainer, provide an explanation for each input, while global explainers, such as GLGExplainer~\cite{azzolin2023global} or GNNInterpreter~\cite{wang2023gnninterpreter}, search for common patterns that explain the general behaviour of the model.

Despite preliminary work extending standard explainability techniques to \glspl{stgnn} for some specific applications in the energy and medical fields~\cite{verdone2024explainable, tang2023explainable, chen2025explainable, altieri2023explainable}, the presence of both spatial and temporal components combined with the black-box nature of neural networks still makes these models particularly difficult to interpret.

A promising direction to address these explainability challenges comes from the field of dynamical systems. In particular, Koopman theory reformulates a complicated nonlinear dynamical system into a simpler linear representation, at the cost of moving to a potentially infinite-dimensional state space.
By transforming the nonlinear dynamics into a linear framework, Koopman theory enables the use of robust, data-driven techniques to approximate the system's evolution directly from empirical measurements. This makes the approach particularly useful in many real-world scenarios where the explicit equations of the dynamical system are unknown, but abundant observation data is available.
Since deep learning models can be seen as dynamical systems~\cite{han2024from, Gravina2025}, Koopman theory has recently been applied to design interpretable deep learning architectures~\cite{lusch2018deep, mohr2021applications} or to perform post hoc analyses~\cite{naiman2023operator}. Particularly relevant to \glspl{tg} is the work in~\cite{melnyk2023understanding,melnyk2020graphkke}, which is, however, limited to the analysis of metastable states of the human microbiome. In~\cite{shi2025when}, inspired by Koopman theory, a new feature propagation mechanism for~\glspl{gnn} is proposed, but its effects on the model's interpretability are not explored, nor is the method extended to~\glspl{stgnn}.

We extend this line of work by studying how Koopman theory can help interpret a spatiotemporal model trained on complex inputs such as \glspl{tg}.
Analysing the model's embeddings with Koopman-inspired techniques such as \gls{dmd} allows us to recover both the temporal patterns and the subgraphs that have the largest influence on the model's decision-making process.
As an additional contribution, we propose using \gls{sindy} as an explainability method for \glspl{tg} for the first time. \Gls{sindy} is a popular algorithm originally introduced to discover governing equations for complex dynamics.

In our experiments, we demonstrate how the proposed methods correctly highlight important features of the input \gls{tg}. In particular, in dissemination processes, the explanations accurately locate the times at which infections occur and the nodes involved.
We further validate the methods qualitatively on a real-world human motion dataset~(\gls{msrc12}), where the explanations consistently highlight the body parts most relevant for action recognition.

\section{Background}
\label{sec:background}

\subsection{Koopman operator theory}

In~\cite{koopman1931hamiltonian}, Koopman proved how to translate a finite-dimensional nonlinear dynamical system into an infinite-dimensional linear one.
Consider a discrete\footnote{The description can be extended to the continuous case~\cite{mezic2021koopman}, but our application uses discrete time, so we will focus on the discrete case only.} dynamical system on a $D$-dimensional state space~$\mathcal{M}$
\begin{equation}
    \label{eq:dynamical-system}
    \bm{x}_{\timeidx+1} = F(\bm{x}_\timeidx),
\end{equation}
with state $\bm{x}\in\mathcal{M}$ and flow map $F:\mathcal{M}\to\mathcal{M}$.
Let $\varphi:\mathcal{M}\to\mathbb{C}$ be an \emph{observable} in the Hilbert space $L^2$, i.e.~$\varphi$ is measurable and the Lebesgue integral of the square of the absolute value of $\varphi$ is finite.
Then define the (discrete-time) Koopman operator $\kappa$ as
\begin{equation}
    \label{eq:koopman}
    \kappa\varphi(\bm{x}_\timeidx)=\varphi(F(\bm{x}_\timeidx))=\varphi(\bm{x}_{\timeidx+1}).
\end{equation}

The Koopman operator acts on the infinite-dimensional space of observables but has the benefit of being linear:
\begin{equation}
\begin{aligned}
    \kappa(a\varphi_1(\bm{x})+b\varphi_2(\bm{x}))&= a\varphi_1(F(\bm{x})) + b\varphi_2(F(\bm{x}))\\
    &=a \kappa\varphi_1(\bm{x}) + b \kappa\varphi_2(\bm{x}).
\end{aligned}
\end{equation}

\subsubsection{\Glsfmtlong{dmd}}
\label{sec:dmd}

\glslocalreset{dmd}
In some rare cases, it is possible to find a finite-dimensional subspace of $L^2$, so the Koopman operator, restricted to that subspace, is both finite-dimensional and linear, allowing the well-studied descriptions of linear systems~\cite{mezic2021koopman}.
In general, however, such an invariant subspace does not exist, so a finite-dimensional approximation of $\kappa$ must be sought instead. This has motivated a growing body of data-driven approaches, founded on Koopman theory, that have increased in popularity~\cite{brunton2021modern}. The idea is to approximate $\kappa$ using trajectories $(\bm{x}_\timeidx)_{\timeidx=1}^T$ collected from real dynamical systems, reinforced using a library of nonlinear functions, and then use the approximation to simulate and analyse the system.

One of the first classes of algorithms introduced to approximate $\kappa$ was~\gls{dmd}~\cite{schmid2010dynamic}. It was introduced in fluid dynamics and transport processes to extract relevant information directly from data, without necessarily knowing the governing equation of the dynamics. The link to the Koopman operator was only clarified later~\cite{kutz2016dynamic, arbabi2017ergodic}.

Suppose we have a dynamical system described by~\eqref{eq:dynamical-system}, from which we collect measurements $\bm{h}_\timeidx=\varphi(\bm{x}_\timeidx)\in\mathbb{R}^F$ at regularly spaced times. We can then build two matrices of snapshots
\begin{equation}
\begin{aligned}
\label{eq:snapshot-matrices}
    \bm{H} &= ( \bm{h}_1, \bm{h}_2, \dots, \bm{h}_{T-1}), \\
    \bm{H}' &= ( \bm{h}_2, \bm{h}_3, \dots, \bm{h}_T).
\end{aligned}
\end{equation}
Then the matrix $\bm{C} \in \mathbb{R}^{F\times F}$ given by $\bm{H}'\simeq \bm{C}\bm{H}$, namely $\bm{C}=\bm{H}'\bm{H}^\dagger$,\footnote{Where $\dagger$ is the Moore-Penrose pseudoinverse.} approximates the Koopman operator~\cite{kutz2016dynamic}.
Since the matrix $\bm{C}$ can be large, the~\gls{dmd} algorithm first computes an~\gls{svd} or~\gls{pca} of the data matrix $\bm{H}$, before producing a rank-reduced matrix $\bm{C}$~\cite{kutz2016dynamic}.
States $\bm{h}_\timeidx$ can be decomposed onto the basis of eigenvectors of $\bm{C}$, called \textit{\gls{dmd} modes}, and their projection $s^{(i)}(\timeidx)$ on the $i$-th mode contains useful information about the dynamics.

The original~\gls{dmd} algorithm described above has been further developed to extend its use and applicability to a wider range of contexts, beyond the fluid dynamics case, and to tackle some of its shortcomings. A review of~\gls{dmd} variations can be found in~\cite{schmid2022dynamic, brunton2021modern}.
For implementations in Python, we refer to~\cite{demo2018pydmd, ichinaga2024pydmd}.

\subsubsection{\Glsfmtlong{sindy}}

\glslocalreset{sindy}
In the context of discovering and approximating governing equations from data, another approach, alternative to~\gls{dmd}, was introduced under the name~\gls{sindy} in~\cite{brunton2016discovering}.
The idea is to approximate the dynamics in~\eqref{eq:dynamical-system} with a library of pre-determined nonlinear functions, only a few of which will be relevant~\cite{Brunton_Kutz_2022}.

Consider the matrices of snapshots of the system's state $\bm{x}_\timeidx\in\mathbb{R}^D$,
\begin{equation}
\label{eq:snapshot-states}
\begin{aligned}
    \bm{X} &= ( \bm{x}_1, \bm{x}_2, \dots, \bm{x}_{T-1})^\top \in \mathbb{R}^{(T-1)\times D} \\
    \bm{X}' &= ( \bm{x}_2, \bm{x}_3, \dots, \bm{x}_{T})^\top \in \mathbb{R}^{(T-1)\times D},
\end{aligned}
\end{equation}
and a library of $J$ candidate nonlinear functions,
\begin{equation}
    \Theta(\bm{X}) = (\mathbb{I}, \bm{X}^2, \dots, \sin(\bm{X}), \dots) \in \mathbb{R}^{(T-1) \times D \times J},
\end{equation}
where each function is applied to $\bm{X}$ element-wise.
The dynamical system~\eqref{eq:dynamical-system} can now be approximated with
\begin{equation}
    \label{eq:sindy}
    \bm{X}' = \Theta(\bm{X})\bm{\xi},
\end{equation}
where $\bm{\xi}\in\mathbb{R}^{J}$ is a sparse vector, obtained via sparse regression, that selects only a few of the most relevant terms of the library $\Theta$.

If we write $\bm{\xi} = ( \xi_1, \dots, \xi_J)^\top$, then equation~\eqref{eq:sindy} becomes
\begin{equation}
    \label{eq:sindy-explicit}
    x_{\timeidx+1,\stateidx} = \sum_{\sindyidx=1}^J\,\Theta(\bm{X})_{\timeidx,\stateidx,\sindyidx}\xi_\sindyidx,
\end{equation}
where each component $\xi_\sindyidx$ expresses the importance of the $\sindyidx$-th nonlinear function in $\Theta$ for the system's dynamics.
Crucially, the sparsity of $\bm{\xi}$ and the linear nature of the equation make the identified model directly interpretable: the non-zero coefficients reveal which interactions drive the dynamics, a property we will exploit for explainability in the context of~\glspl{stgnn}.

\subsection{Spatiotemporal models}

In the present work, we focus on \glspl{tg} classification task: the whole input \gls{tg} is processed by a \gls{stgnn} trained to predict its class $y$. This section first defines \glspl{tg}, then describes in detail \glspl{stgnn}.

\subsubsection{\Glsfmtlong{tg}}
\glslocalreset{tg}

\Glspl{tg} model data whose spatial relations are represented by a graph, and the node labels and topology are both time-dependent. For a formal treatment of~\glspl{tg} and recent reviews of methods and tasks applied to~\glspl{tg} and time series with dependencies expressed as a graph, we refer to~\cite{longa2023graph, cini2025graph, cini2023sparse}.

In this work, we rely on the following definition of~\gls{tg}.

\begin{definition}[Discrete time \gls{tg}]
    Given a set of $N$ nodes $\mathcal{V}$ and a set of edges $\mathcal{E}$, we define a~\glsentrylong{tg} as a sequence of graphs
    \begin{equation}
        \mathcal{G} := \left((\mathcal{V}_\timeidx, \mathcal{E}_\timeidx)\right)_{\timeidx=1}^T, 
    \end{equation}
    with $\mathcal{V}_\timeidx \subseteq \mathcal{V}$ and $\mathcal{E}_\timeidx \subseteq \mathcal{E}$.
    Each node $v\in\mathcal{V}_\timeidx$ is equipped with a feature vector $\bm{x}^v_\timeidx\in\mathbb{R}^D$.
\end{definition}

To simplify the notation, we will not consider features on edges, and the feature vector of the $n$-th node at time $\timeidx$ will be represented by $\bm{x}_{\timeidx,n}$. The adjacency matrix is
\begin{equation}
    (\bm{A}_\timeidx)_{n,m}=
    \begin{cases}
        1 \quad \text{if $(v_n,v_m)\in\mathcal{E}_\timeidx$} \\
        0 \quad \text{otherwise}
    \end{cases},
\end{equation}
and it varies with time since edges are not guaranteed to exist at all times. We will denote by $N_\mathcal{G}$ the number of~\glspl{tg} in a dataset.

\subsubsection{\Glsfmtlong{stgnn}}
\label{sec:stgnn}

To process~\glspl{tg}, we use a snapshot-based~\cite{longa2023graph}, time-and-space~\cite{cini2025graph} \gls{stgnn}, i.e.~a deep learning architecture where temporal and spatial processing cannot be factorised into two separate steps.

As representative \glspl{stgnn}, we consider three different models:
\begin{itemize}
    \item a~\gls{gcrn}~\cite{seo2018structured},
    \item a~\gls{dcrnn}~\cite{li2018diffusion},
    \item and a~\gls{gwn}~\cite{wu2019graph}.
\end{itemize}

\Glspl{gcrn} and~\glspl{dcrnn} use a spatial encoder at each time step intertwined with a~\gls{rnn} such as a~\gls{lstm}, to process temporal dependencies. Let $\mathcal{N}_\timeidx(v_n)$ represent the neighbourhood of the \mbox{$n$-th} node $v_n$ at time $\timeidx$, $\ell$ the layer index, and $\text{Enc}(\cdot)$ a spatial encoder. The model encodes an input $\bm{x}_{\timeidx,n}$ into an embedding $\bm{h}_{\timeidx,n,\ell}$ as follows:
\begin{equation}
\begin{aligned}
    \bm{u}_{\timeidx,n,\ell} &=
    \begin{cases}
    \bm{x}_{\timeidx,n} & \ell = 1 \\
    \bm{h}_{\timeidx,n,\ell-1} & \ell > 1
    \end{cases}, \\
    \bm{z}_{\timeidx,n,\ell} &= \text{Enc}(\{\bm{u}_{\timeidx,m,\ell}\}_{v_m\in\mathcal{N}_\timeidx(v_n)}, \bm{A}_\timeidx), \\
    \bm{i}_{\timeidx,n,\ell} &= \sigma(\bm{W}^\text{zi}\bm{z}_{\timeidx,n,\ell} + \bm{W}^\text{hi}\bm{h}_{\timeidx-1,n,\ell}), \\
    \bm{f}_{\timeidx,n,\ell} &= \sigma(\bm{W}^\text{zf}\bm{z}_{\timeidx,n,\ell} + \bm{W}^\text{hf}\bm{h}_{\timeidx-1,n,\ell}), \\
    \bm{g}_{\timeidx,n,\ell} &= \tanh(\bm{W}^\text{zg}\bm{z}_{\timeidx,n,\ell} + \bm{W}^\text{hg}\bm{h}_{\timeidx-1,n,\ell}), \\
    \bm{o}_{\timeidx,n,\ell} &= \sigma(\bm{W}^\text{zo}\bm{z}_{\timeidx,n,\ell} + \bm{W}^\text{ho}\bm{h}_{\timeidx-1,n,\ell}), \\
    \bm{c}_{\timeidx,n,\ell} &= \bm{f}_{\timeidx,n,\ell} \odot \bm{c}_{\timeidx-1,n,\ell} + \bm{i}_{\timeidx,n,\ell}\odot \bm{g}_{\timeidx,n,\ell}, \\
    \bm{h}_{\timeidx,n,\ell} &= \bm{o}_{\timeidx,n,\ell} \odot \tanh(\bm{c}_{\timeidx,n,\ell}).
\end{aligned}
\label{eq:gcrn}
\end{equation}

\Glspl{gcrn} use a~\gls{gcn}~\cite{kipf2017semisupervised} at each time step as the spatial encoder $\text{Enc}(\cdot)$.
\Glspl{dcrnn} use a~\gls{dcl}, which simulates, at each time step, a diffusion process with learnable transition probabilities over the graph according to the adjacency matrix~$\bm{A}_\timeidx$:
\begin{equation}
\label{eq:dcl}
    \bm{z}_{\timeidx,n,\ell} = \text{DCL}(\{\bm{u}_{\timeidx,m,\ell}\}_{m=1}^N) =\sum_{k=0}^{K-1} \sum_{m=1}^N  (\bm{P}^k_\timeidx)_{n,m} \bm{u}_{\timeidx,m,\ell} \bm{W}_k,
\end{equation}
where $\bm{P}_\timeidx=\bm{D}_\timeidx^{-1} \bm{A}_\timeidx$, with $\bm{D}_\timeidx$ the degree matrix, is the transition matrix of the diffusion process, and $K$ is the number of steps.

The third \gls{stgnn} we consider,~\gls{gwn}, has a different structure. Each layer consists of a (gated)~\gls{tcn} for the temporal part and a modified~\gls{dcl} for the spatial component: 
\begin{equation}
\label{eq:gwn}
\begin{aligned}
    \bm{z}_{\timeidx,n,\ell} &=\tanh(\bm{W}_1 \star \bm{u}_{:,n,\ell})_\timeidx \odot \sigma(\bm{W}_2 \star \bm{u}_{:,n,\ell})_\timeidx, \\
    \bm{h}_{\timeidx,n,\ell} &= \text{DCL}(\{\bm{z}_{\timeidx,m,\ell}\}_{m=1}^N) + \sum_{k=0}^{K} \sum_{m=1}^N  \tilde{\bm{A}}^k_{n,m} \bm{z}_{\timeidx,m,\ell} \tilde{\bm{W}}_k,
\end{aligned}
\end{equation}
where $\star$ is the dilated causal convolution operation, with dilation $d$ and temporal kernel $K_\texttt{t}$; to the~\gls{dcl} described in~\eqref{eq:dcl}, the~\gls{gwn} model adds an extra diffusion term with a trainable time-independent adjacency matrix $\bm{\tilde{A}}$.

For node-level tasks, the embedding for the $n$-th node is given by $\bm{h}_n\in\mathbb{R}^{FL}$, obtained by concatenating, along the layer dimension, the embeddings at the last time step of the \gls{stgnn} $\bm{h}_{T,n,\ell}\in\mathbb{R}^F$.\footnote{Another option is to simply take the last layer embedding $\bm{h}_n:=\bm{h}_{T,n,L}\in\mathbb{R}^F$.}
On the other hand, the output $y$ of a graph-level task is given by processing with a~\gls{mlp} the sum of all node embeddings $\bm{h}=\sum_{n=1}^{N}\,\bm{h}_n \in \mathbb{R}^{FL}$,
\begin{equation}
\label{eq:gcrn-output}
    y = \text{MLP}(\bm{h}) \in \mathbb{R}^C,
\end{equation}
where $C$ is the dimension of the desired output, e.g.~the number of classes.

For a classification task, the model is trained using a cross-entropy loss between the model's output $y$ and the class label $\hat{y}\in\{1,\dots,C\}$,
\begin{equation}
    \ell_{\text{ce}}(y,\hat{y}) = - \log \frac{\exp y_{\hat{y}}}{\sum_{c=1}^C\,\exp y_c}.
\end{equation}

\section{Methods}

\begin{figure*}[!ht]
    \centering
    \includegraphics[width=\textwidth, trim=0 1cm 1cm 0, clip]{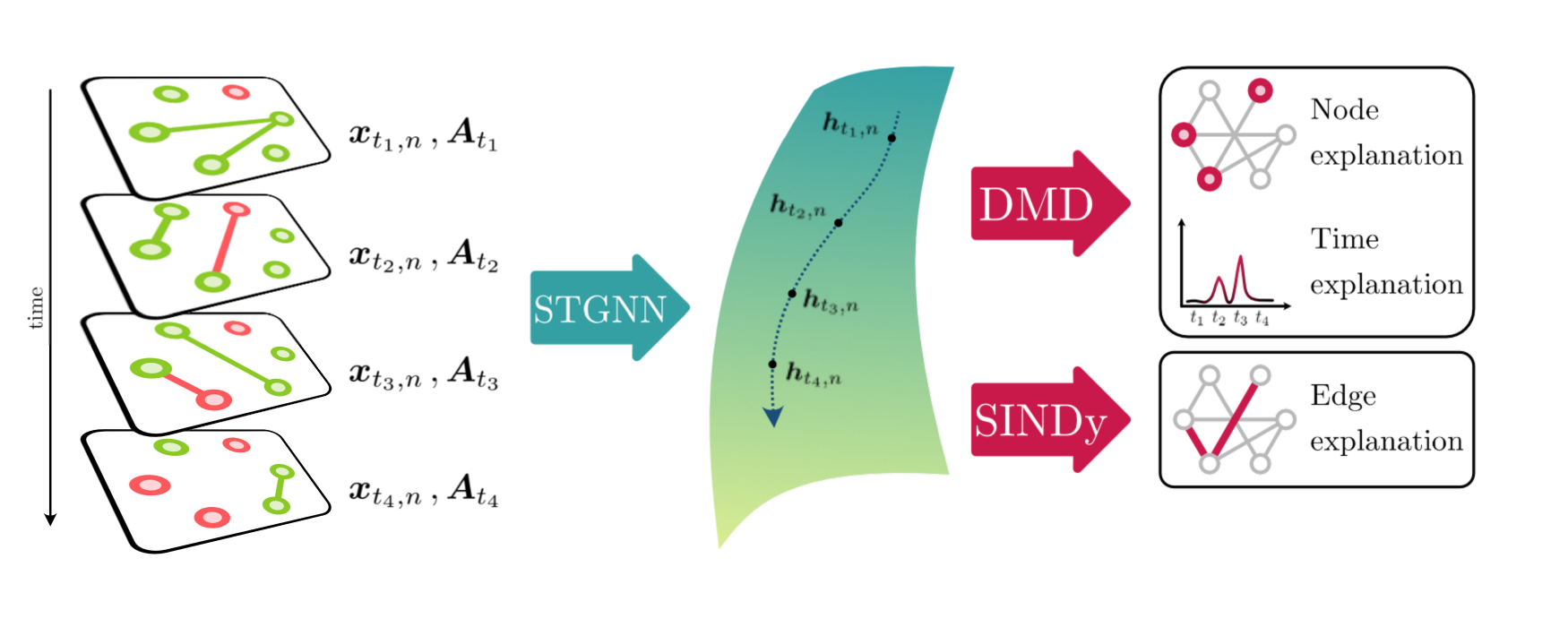}
    \caption{Overview of the proposed framework: the input \gls{tg} (\textit{left}) is processed via an \gls{stgnn}, which produces a trajectory of embeddings $\bm{h}_{\timeidx,n}$ (\textit{centre}), whose dynamics is analysed via \gls{dmd} and \gls{sindy} to produce spatial and temporal explanations (\textit{right}).}
    \label{fig:framework}
\end{figure*}

\subsection{Research hypotheses}
\label{sec:hypotheses}

Providing an instance-based explanation usually involves computing weights that highlight the most relevant parts of the input. In the case of~\glspl{tg}, that means finding a weight $w_\texttt{t}(\timeidx)$ for time step $\timeidx$, and spatial weights $w_\texttt{s}(n)$ for node $n$, or $w_\texttt{e}(n,m)$ for edge $(n,m)$.
To properly measure the performance of our explainability methods, we require ground truth for each of these quantities. We call $m_\texttt{t}(\timeidx)$ the time ground truth, $m_\texttt{s}(n)$ the spatial ground truth on nodes, and $m_\texttt{e}(n,m)$ the spatial ground truth on edges.

We propose two separate post hoc explainability methods, the first based on \gls{dmd} and the second on \gls{sindy}, that provide either temporal or spatial explanations for \glspl{tg}, or both.
We state below the hypotheses on how \gls{dmd} and \gls{sindy} can explain the \gls{stgnn} model:

\begin{enumerate}
    \item \label{hyp:time-gt} The input drives the dynamics of at least one \gls{dmd} state component $s^{(i)}(\timeidx)$ of the~\gls{tg}'s embedding, where $s^{(i)}(\timeidx)$ is the projection of the \gls{stgnn}'s embedding $\bm{h}_\timeidx$ onto the $i$-th \gls{dmd} mode (see \Cref{sec:expl-dmd}).
    Hence, a sudden change in $s^{(i)}(\timeidx)$ at time $\timeidx$ indicates that the model's internal state reacted to a task-relevant event in the input at that moment. We can therefore use the derivative $ds^{(i)}(\timeidx)/d\timeidx$ to compute the time weight~$w_\texttt{t}(\timeidx)$. 
    \item The projection $s_n^{(i)}(\timeidx)$ of a node embedding at each time step (in particular, the last one $T$) identifies whether that node is important for the output or not, and so it is a proxy for the spatial explanation~$w_\texttt{s}(n)$. \label{hyp:node-gt}
    \item \gls{sindy} fits the dynamics of each node's embedding as a sparse combination of library terms. By construction, the mixed terms in the library correspond to interactions between neighbouring nodes, i.e.~edges of the input~\gls{tg}. A large regression weight $\xi_\sindyidx$ for a term involving edge $(n,m)$ therefore indicates that this interaction strongly drives the embedding dynamics, and hence that the edge is important for the model's prediction. This allows us to use $\xi_\sindyidx$ to compute the edge explanation~$w_\texttt{e}(n,m)$. \label{hyp:edge-gt}
\end{enumerate}

\subsubsection{Explainability using~\glsfmtshort{dmd}}
\label{sec:expl-dmd}

We apply \gls{dmd} to analyse the trajectories of the \gls{stgnn}'s states and understand how inputs are processed by~\gls{stgnn} and affect their output.
To reduce the computational complexity of the analysis, a common first step~\cite{naiman2023operator} when working with these techniques consists in applying \gls{pca} or \gls{svd} to reduce the dimensionality of the embeddings from $FL$ to $f$. An alternative approach is to use \gls{ttdmd}~\cite{klus2018tensor,oseledets2011tensor}, described in more detail in \Cref{sec-app:ttdmd}.

Let $\bm{h}'_{\timeidx,n}\in\mathbb{R}^f$ and $\bm{h}'_\timeidx\in\mathbb{R}^f$ be the embeddings of the nodes and the whole \gls{tg}, respectively, projected onto the principal components.
To apply \gls{dmd} to the whole dataset, we use ridge regression to fit an operator $\bm{C}\in\mathbb{R}^{f\times f}$ on the training split of the dataset of~\glspl{tg} embeddings:
\begin{equation}
    \label{eq:tg-koopman-graph}
    \bm{h}'_{\timeidx+1} = \bm{C} \bm{h}'_{\timeidx}.
\end{equation}
We can then diagonalise $\bm{C}$, with eigenvalues $\lambda_i\in\mathbb{C}$ and eigenvectors $\bm{v}_i\in\mathbb{C}^f$, and study the dynamics along the eigenspaces $\langle \bm{v}_i \rangle$. We denote by $s^{(i)}(\timeidx) = \bm{v}_i^\top \bm{h}'_{\timeidx}$ the projection onto the $i$-th \gls{dmd} mode at time $\timeidx$ (ordered according to the magnitude of the corresponding eigenvalue $\lambda_i$). The analysis of $s^{(i)}(\timeidx)$ is then performed on the validation set, which allows us to evaluate the generalisation capability of the explainability framework.

On the other hand, to apply \gls{dmd} on the nodes we focus on a single \gls{tg} $\mathcal{G}$ at a time, and consider its nodes' embeddings $\bm{h}'_{\timeidx,n}$ to fit a matrix $\bm{C}_\mathcal{G}\in\mathbb{R}^{f\times f}$
\begin{equation}
    \label{eq:tg-koopman-node}    
    \bm{h}'_{\timeidx+1,n} = \bm{C}_\mathcal{G} \bm{h}'_{\timeidx,n}.
\end{equation}
As in the previous case, we diagonalise $\bm{C}_\mathcal{G}$ and compute the projection on the \gls{dmd} modes for nodes, $s_n^{(i)}(\timeidx)$.
Unlike before, however, we cannot split the nodes of the \gls{tg} $\mathcal{G}$ into training and validation sets. Thus, we fit and analyse $\bm{C}_\mathcal{G}$ on all nodes.

When $|\lambda_i|<1$, we expect both $s^{(i)}(\timeidx)$ and $s_n^{(i)}(\timeidx)$ to decay to 0 and contribute little to the final state (see hypotheses~\ref{hyp:time-gt} and~\ref{hyp:node-gt}).
Conversely, when $|\lambda_i|\simeq 1$, the corresponding component $s^{(i)}(\timeidx)$ neither grows nor decays, but persists throughout the sequence. In this regime, the component's evolution is not dominated by transient effects and instead tracks the non-trivial spatiotemporal structure of the input, making it a meaningful proxy for the model's decision process. For this reason, we select \gls{dmd} modes with $|\lambda_i|\simeq 1$ and define the time weight
\begin{equation}
    w_\texttt{t}^{(i)}(\timeidx):=\left| \frac{ds^{(i)}(\timeidx)}{d\timeidx} \right|,
    \label{eq:w_t}
\end{equation}
and the node weight as the distance from the average,
\begin{equation}
    w_\texttt{s}^{(i)}(n):=\left| s_n^{(i)}(T) - \frac{1}{|\mathcal{V}|} \sum_m\,s_m^{(i)}(T) \right|.
    \label{eq:w_n}
\end{equation}
Hypotheses~\ref{hyp:time-gt} and~\ref{hyp:node-gt} can be tested by comparing equations~\eqref{eq:w_t} and~\eqref{eq:w_n} with $m_\texttt{t}(\timeidx)$ and $m_\texttt{s}(n)$, respectively.

The same approach allows us to define a combined explanation for the entire \gls{tg} $\mathcal{G}$, which jointly accounts for spatial and temporal patterns.
Indeed, we can extend the definition~\eqref{eq:w_n} by considering all time steps and not just the last one. We therefore define
\begin{equation}
\label{eq:w_G}
    w_\mathcal{G}^{(i)}(\timeidx,n):=\left| s_n^{(i)}(\timeidx) - \frac{1}{|\mathcal{V}|} \sum_m\,s_m^{(i)}(\timeidx) \right|,
\end{equation}
such that $w_\texttt{s}^{(i)}(n)=w_\mathcal{G}^{(i)}(T,n)$.
We refer to \Cref{par:st-explanations} for a discussion on the results obtained with this approach.

\subsubsection{Explainability using~\glsfmtshort{sindy}}

As a data-driven method for approximating governing equations, \gls{sindy} can also learn and store useful information about the dynamics of $\bm{h}'_{\timeidx,n}$. 
In our case, the matrices defined in equations~\eqref{eq:snapshot-states} become
\begin{equation}
\begin{aligned}
    \bm{H}_n = (\bm{h}'_{1,n},\bm{h}'_{2,n},\dots,\bm{h}'_{T-1,n}) \in \mathbb{R}^{T-1,f} \\
    \bm{H}'_n = (\bm{h}'_{2,n},\bm{h}'_{3,n},\dots,\bm{h}'_{T,n}) \in \mathbb{R}^{T-1,f},
\end{aligned}
\end{equation}
where $n=1,\dots,N$.
The difficult and somewhat arbitrary part of \gls{sindy} is the choice of the library of nonlinearities $\Theta$. In our case, though, we can take advantage of this flexibility to enforce a bias in the reconstructed dynamics.
Since the \gls{gcn} and \gls{dcl} components aggregate information from neighbouring nodes, the next embedding of a node is by construction a nonlinear function of its own and its neighbours' current embeddings. 
We can therefore align the \gls{sindy} library with this known inductive bias, and restrict the candidate terms to monomials involving the embeddings of neighbouring node pairs. 
Concretely, the library for node $n$ contains only those terms involving the node itself and its neighbours, namely
\begin{equation}
\begin{aligned}
    \Theta(\bm{H}_n) = (&\bm{H}_n^2, \bm{H}_n^3, \bm{H}_n \bm{H}_{m_1}, \dots, \bm{H}_n  \bm{H}_{m_1}^2, \\
                    &\dots,\bm{H}_n \bm{H}_{m_2}, \dots, \bm{H}_n  \bm{H}_{m_2}^2,\dots),    
\end{aligned}
\end{equation}
where all operations are performed element-wise, and the node indices $m_i$ refer to nodes that are connected to the $n$-th node at least once, i.e.~all those $m_i$ for which there exists a time $\timeidx$ such that $(\bm{A}_\timeidx)_{n, m_i}=1$.
While we could consider higher-order terms, such as $\bm{H}_n^2 \bm{H}_m^2$, and nodes more than one hop away from $n$, for the sake of simplicity, we consider monomials up to degree 3 and only one-hop neighbourhoods. We treat the order of the monomials, $d_\text{SINDy}$, as a hyperparameter.

After introducing an extra index for the node dimension, equation~\eqref{eq:sindy-explicit} becomes
\begin{equation}
    h'_{\timeidx+1,n,\stateidx} = \sum_{\sindyidx=1}^J\,\Theta(\bm{H}_n)_{\timeidx,\stateidx,\sindyidx}\,\xi_{n,\sindyidx},
\end{equation}
where the index $\sindyidx=1,\dots,J$ spans the monomials of $\Theta$, each corresponding by construction to an edge of the input \gls{tg}, including self-loops.
We can interpret $\xi_{n,\sindyidx}$ as a weight that measures how strongly the $\sindyidx$-th term contributes to the dynamics of the $n$-th node embedding, $\bm{h}'_n$. This allows us to define a weight for each edge~$(n,m)$ as
\begin{equation}
    \label{eq:sindy-weights}
    w_\texttt{e}(n,m) := \sum_{n'=1}^N \sum_{\sindyidx\sim(n,m)} \, |\xi_{n',\sindyidx}|.
\end{equation}
In equation~\eqref{eq:sindy-weights}, the inner sum runs over all monomial weights $\xi_{n',\sindyidx}$ that refer to the same edge $(n,m)$, which expresses how important the edge $(n,m)$ is for the \mbox{$n'$-th} node. The inner sum is needed because different monomials can refer to the same edge, e.g.~the terms $\bm{H}_n \bm{H}_m^2$ and $\bm{H}_n^2 \bm{H}_{m}$ both relate to $(n,m)$. The outer sum runs over all nodes and therefore measures how important edge $(n,m)$ is for the whole~\gls{tg}. 
The edge weight $w_\texttt{e}(n,m)$ can then be compared with the ground truth $m_\texttt{e}(n,m)$ to test hypothesis~\ref{hyp:edge-gt}.

\section{Experiments}
\label{sec:experiments}

To ensure reproducibility, we make our code and experimental setup available in a public repository.\footnote{\href{https://anonymous.4open.science/r/Koopman-TGNN-Interpretability-11D7}{GitHub repository}}

\subsection{Datasets}

We test the hypotheses above on two types of datasets: the first one consists of semi-synthetic datasets for binary classification tasks with explainability ground truth, while the second is a real-world action classification dataset with no explainability ground truth.

In the first class of datasets, we have~\glspl{tg} whose time-varying topologies describe different types of social interactions. 
The time-varying node labels $x_{\timeidx,n}\in \{0,1\}$ are produced via a dissemination process simulated with the susceptible-infected model~\cite{Oettershagen2020dissemination}. \Glspl{tg} of class $1$ correspond to dissemination processes.
In~\glspl{tg} of class $0$, instead, the infected nodes found via the same dissemination process are shuffled randomly at each time step.
In each dataset, the two classes are balanced.

While our methods are general and applied to both these semi-synthetic datasets and a real-world dataset (see below), the availability of ground truth in the semi-synthetic datasets allows us to quantitatively measure explainability performance.
Specifically, the time ground truth $m_\texttt{t}(\timeidx)$ counts the infections occurring at each time step $\timeidx$ between adjacent nodes. That is, a time step $\timeidx$ contributes to $m_\texttt{t}(\timeidx)$ if $x_{\timeidx,n}x_{\timeidx,m}=0$ and $x_{\timeidx+1,n}x_{\timeidx+1,m}=1$ for some adjacent nodes $n,m$.
The spatial ground truth on nodes $m_\texttt{s}(n)$ indicates which nodes have been infected: $m_\texttt{s}(n)=1$ if node $n$ has been infected, $m_\texttt{s}(n)=0$ otherwise.
The spatial ground truth on edges $m_\texttt{e}(n,m)$ is computed by finding the edges that transmit the infection: $m_\texttt{e}(n,m)=1$ if there has been an infection between nodes $n$ and $m$, $0$ otherwise.

As mentioned, we also test our approach on the real-world dataset \gls{msrc12}~\cite{fothergill2012instructing}, consisting of sequences of 12 human movements, where the node labels $\bm{x}_{\timeidx,n}\in\mathbb{R}^3$ represent the 3D joint coordinates. When training a~\gls{stgnn} on this dataset, each input is augmented by performing random rotations in 3D space.
Unlike the semi-synthetic datasets, in~\gls{msrc12} the topology does not change over time, and there is no explainability ground truth, so we provide only qualitative results for our explainability methods. Moreover, since the sequences have different lengths, each one needs to be padded along the time dimension before forming batches.

See Appendix~\ref{sec-app:datasets} for further details on each dataset.

\subsection{Metrics}
\label{sec:metrics}

For those datasets that come with a ground truth, we can define metrics to assess quantitatively whether the hypotheses formulated in \Cref{sec:hypotheses} hold. To test hypothesis~\ref{hyp:time-gt}, we measure the agreement between the time ground truth $m_\texttt{t}(\timeidx)$ and the time weight $w_\texttt{t}^{(i)}(\timeidx)$.
Projections on \gls{dmd} modes can be quite noisy, while the time ground truth $m_\texttt{t}(\timeidx)$ is very sharp, being $0$ almost everywhere and positive only at a few sparse time steps. This poses a significant challenge when comparing two time signals, common in fields such as anomaly detection~\cite{wagner2026formally, kim2022towards}. To overcome this, we consider a regularised version of the time ground truth, obtained by convolving $m_\texttt{t}(\timeidx)$ with a uniform filter to make it smoother before computing the F1 metric, and we apply a threshold in different ways:
\begin{itemize}
    \item \emph{F1 with thresholds}. The F1 score between the time ground truth and the time steps $\timeidx$ such that $w_\texttt{t}^{(i)}(\timeidx)>\delta$, with $\delta$ being a threshold:
    \begin{description}
        \item[MAX] $\delta=\delta' \cdot \max_\timeidx w_\texttt{t}^{(i)}(t)$, i.e.~the threshold is a fraction of the maximum value of $w_\texttt{t}^{(i)}(t)$;
        \item[AVG] $\delta=\mu_{w_\texttt{t}} + \sigma_{w_\texttt{t}}$, where $\mu_{w_\texttt{t}}$ and $\sigma_{w_\texttt{t}}$ are the time average and standard deviation respectively;
        \item[MAD] $\delta= \bar{w}_\texttt{t} + k\cdot \text{median}(|w_\texttt{t}-\bar{w}_\texttt{t}|)$, where $k$ is a hyperparameter (we use $k=3$), and $\bar{w}_\texttt{t}=\text{median}(w_\texttt{t})$.
    \end{description}
    \item \emph{F1 with window average}. As above, but in addition, we first take a running average of $w_\texttt{t}^{(i)}(\timeidx)$ to reduce noise, with window size $\omega$.
\end{itemize}

For hypothesis~\ref{hyp:node-gt}, we compare $w_\texttt{s}^{(i)}(n)$ from equation~\eqref{eq:w_n} with the explanation ground truth $m_\texttt{s}(n)$. Since identifying the explanation is a binary classification problem at the node level, we can measure the~\gls{auc} score between $w_\texttt{s}^{(i)}(n)$ and $m_\texttt{s}(n)$, a metric often called \emph{plausibility} in the literature~\cite{longa2025explaining}. We denote it $\text{AUC}_\mathcal{G}$. We refer to~\cite{fontanesi2025bridging} for a discussion on the challenges of explaining~\glspl{gnn} even in the presence of a ground truth.

For hypothesis~\ref{hyp:edge-gt}, we use weights $w_\texttt{e}(n,m)$ from equation~\eqref{eq:sindy-weights}, and compare them with $m_\texttt{e}(n,m)$ via an \gls{auc} score, called~$\displaystyle\smash{\text{AUC}_\text{edge}}$.

Standard explainability metrics for~\glspl{gnn} in the absence of ground truth do exist --- e.g.~\emph{faithfulness}, \emph{fidelity}, and \emph{comprehensiveness}~\cite{agarwal2023evaluating, fontanesi2024explaining, longa2025explaining, azzolin2025reconsidering} ---, but their extension to the \glspl{tg} domain is not trivial and remains underexplored (see~\citet{dileo2025evaluating} for an early attempt in the context of link prediction). Although we acknowledge that a systematic adaptation of these metrics to temporal graphs is important, it lies beyond the scope of this work; therefore, we leave it as a direction for future research. For this reason, the performance of the proposed methods on the \gls{msrc12} dataset is explored only qualitatively.

\subsection{Results}

Before testing the proposed explainability tools, we tune the \gls{stgnn} hyperparameters to maximise classification accuracy. The accuracies obtained with the best hyperparameters are reported in \Cref{tab:accuracies}.

Since the F1 scores depend on the threshold $\delta$ and the window size $\omega$, we perform a grid search on these parameters too. We refer to Appendix~\ref{sec-app:hyperparameters} for details.

\begingroup
\renewcommand{\arraystretch}{1}
\begin{table*}
    \caption{Accuracies of the best-performing \glspl{stgnn}, averaged over 5 runs.}
    \centering
    \small
    \[
    \begin{array}{ccccccc}
        \toprule
        \text{\glsfmtshort{stgnn}} & \text{Facebook} & \text{Infectious} & \text{DBLP}& \text{Highschool} & \text{Tumblr} & \text{\glsfmtshort{msrc12}} \\
        \cmidrule(r){1-1} \cmidrule(l){2-7}
        \text{\glsfmtshort{gcrn}}   & 0.95 \pm 0.02 & 0.97 \pm 0.03 & 0.991 \pm 0.008 & 0.87 \pm 0.11 & 0.96 \pm 0.01 & 0.87 \pm 0.23 \\ 
        \text{\glsfmtshort{dcrnn}}  & 0.95 \pm 0.01 & 0.93 \pm 0.07 & 0.988 \pm 0.005 & 0.96 \pm 0.06 & 0.85 \pm 0.09 & 0.963 \pm 0.009 \\ 
        \text{\glsfmtshort{gwn}}    & 0.95 \pm 0.01 & 0.86 \pm 0.04 & 0.95  \pm 0.01  & 0.89 \pm 0.04 & 0.96 \pm 0.02 & 0.909 \pm 0.008 \\ 
        \bottomrule
    \end{array}
    \]
    \label{tab:accuracies}
\end{table*}
\endgroup

\paragraph{Time explanations}
\label{sec:time-explanations}

\begin{figure*}
    \centering
    \begin{subfigure}[b]{\textwidth}
        \centering
        \includegraphics[width=\textwidth]{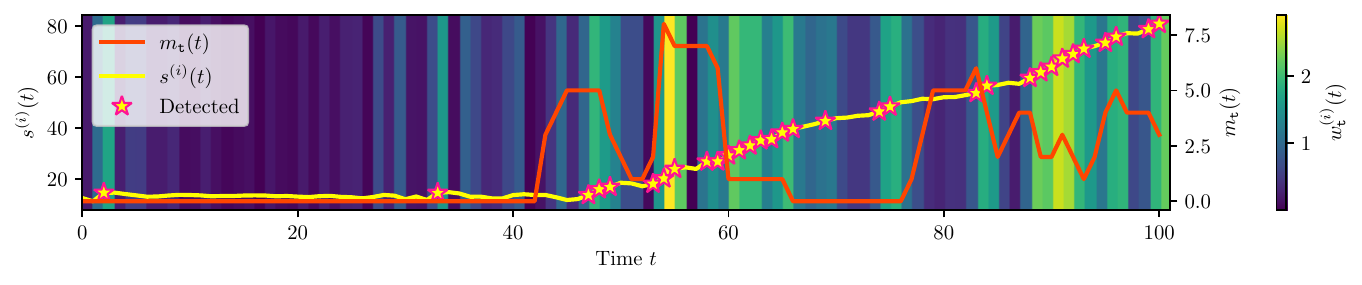}
        \caption{Time explanation via threshold. The F1 score is $0.68$, affected by some false negatives.}
    \end{subfigure}
    \begin{subfigure}[b]{\textwidth}
        \centering
        \includegraphics[width=\textwidth]{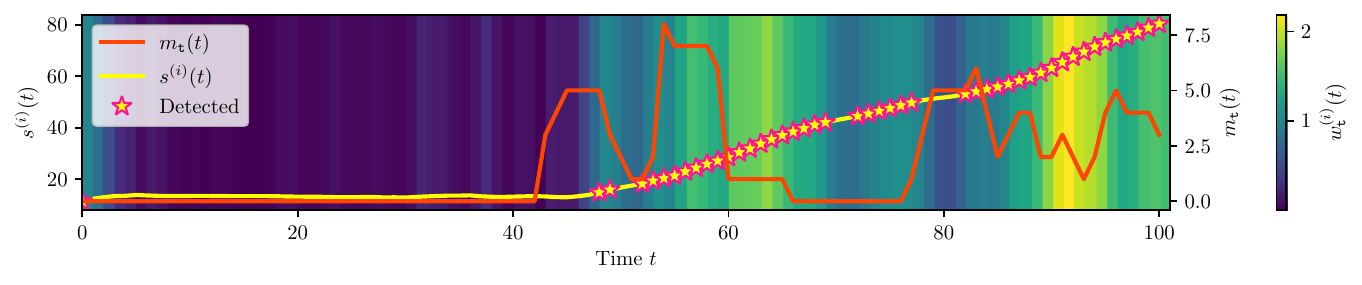}
        \caption{Time explanation via window average. The F1 score is $0.81$.}
    \end{subfigure}
    \caption{Examples of time explanations for the Facebook dataset and \gls{gcrn} model. The red line represents the smoothed ground truth $m_\texttt{t}(\timeidx)$, the yellow line is the relevant component $s^{(i)}(\timeidx)$, the background colour scale shows the explanation weight $w_\texttt{t}^{(i)}(t)$, the stars highlight those times $t$ where $w_\texttt{t}^{(i)}(t)>\delta$.}
    \label{fig:time-example}
\end{figure*}

To test hypothesis~\ref{hyp:time-gt}, we report in Table~\ref{tab:time-results} the F1 scores comparing the time weights $w_\texttt{t}^{(i)}$ with the time ground truth $m_\texttt{t}$.
We also report two baseline values. The first, $\text{F1naïf}$, is obtained from a naïf explainer that outputs $w_\texttt{t}^{(i)}(\timeidx)=1$ for all $\timeidx$. 
As expected, $\text{F1naïf}$ is always very low, evidencing the difficulty of comparing this type of signal~\cite{wagner2026formally,kim2022towards}. The second baseline, $\text{F1sal}$, is computed using a saliency map as explanation (more details are given in \Cref{sec-app:saliency}) \cite{Simonyan14a}. The F1 scores found with our methods are comparable to or better than those found with saliency. In \Cref{fig:time-example}, we report two examples of time explanations from the Facebook dataset: in the top panel, the detection is obtained by thresholding $w_\texttt{t}^{(i)}(\timeidx)$ directly, while in the bottom panel we first apply a window average to $w_\texttt{t}^{(i)}(\timeidx)$.

Results associated with the \gls{gwn} highlight some limitations of our explainability method. As discussed in \Cref{sec:stgnn}, equation~\eqref{eq:gwn}, a \gls{gwn} uses a dilated causal convolution operation with dilation $d$ and temporal kernel $K_\texttt{t}$, which makes the time sequence of embeddings $\displaystyle\smash{( \bm{h}_{\timeidx,n})_{\timeidx=1}^{T_\text{out}}}$ shorter than the time sequence of inputs $(\bm{x}_{\timeidx,n})_{\timeidx=1}^{T_\text{in}}$. The relationship between $T_\text{out}$ and $T_\text{in}$,
\begin{equation}
    T_{\text{out}} = T_{\text{in}} - (K_\texttt{t}-1) \sum_{\ell=0}^{L-1} d^{(\ell \bmod 2)},
\end{equation}
depending on the choice of hyperparameters, can significantly affect the length of the sequence of states to which we apply our methods, especially if the dataset consists of short input sequences. In choosing the hyperparameters, we traded off some accuracy to prevent $T_\text{out}$ from becoming too short, which in turn negatively affects the effectiveness of the explanations. This is particularly apparent for the Infectious dataset.

\begingroup
\renewcommand{\arraystretch}{1}
\begin{table*}
    \caption{Results of experiments to test hypothesis~\ref{hyp:time-gt}. The averages and standard deviations are computed over 5 runs. Methods scoring the highest mean value are reported in \textbf{bold}.}
    \centering
    \small
    \[
    \begin{array}{clccccc}
        \toprule
        & \text{Metrics} & \text{Facebook} & \text{Infectious} & \text{DBLP} & \text{Highschool} & \text{Tumblr} \\
        \cmidrule(r){1-1} \cmidrule(lr){2-2} \cmidrule(l){3-7}
        \multirow{3}{*}{\rotatebox[origin=c]{90}{\glsfmtshort{gcrn}}}  %
        & \text{F1}     & \bm{0.33 \pm 0.03}    & \bm{0.43 \pm 0.14}    & \bm{0.59 \pm 0.13}    & \bm{0.33 \pm 0.15}    & 0.23 \pm 0.03         \\
        \cmidrule(lr){2-2} \cmidrule(l){3-7}
        & \text{F1naïf} & 0.004 \pm 0.001       & 0.023 \pm 0.008       & 0.0006\pm0.0006       & 0.017 \pm 0.008       & 0.010 \pm 0.002       \\
        & \text{F1sal}  & 0.28 \pm 0.06         & 0.19 \pm 0.11         & 0.54 \pm 0.22         & 0.21 \pm 0.13         & \bm{0.37 \pm 0.09}    \\
        \cmidrule(r){1-1} \cmidrule(lr){2-2} \cmidrule(l){3-7}
        \multirow{3}{*}{\rotatebox[origin=c]{90}{\glsfmtshort{dcrnn}}}  %
        & \text{F1}     & 0.30 \pm 0.06         & \bm{0.54 \pm 0.03}    & 0.47 \pm 0.16         & 0.41 \pm 0.09         & 0.26 \pm 0.06         \\
        \cmidrule(lr){2-2} \cmidrule(l){3-7}
        & \text{F1naïf} & 0.003 \pm 0.001       & 0.02 \pm 0.01         & 0.0006 \pm 0.0006     & 0.02 \pm 0.01         & 0.010 \pm 0.002       \\
        & \text{F1sal}  & \bm{0.45 \pm 0.02}    & 0.07 \pm 0.03         & \bm{0.65 \pm 0.13}    & \bm{0.49 \pm 0.11}    & \bm{0.38 \pm 0.02}    \\
        \cmidrule(r){1-1} \cmidrule(lr){2-2} \cmidrule(l){3-7}
        \multirow{3}{*}{\rotatebox[origin=c]{90}{\glsfmtshort{gwn}}}  %
        & \text{F1}     & 0.36 \pm 0.02         & \bm{0.33 \pm 0.04}    & \bm{0.51 \pm 0.05}    & \bm{0.35 \pm 0.10}    & 0.31 \pm 0.02         \\
        \cmidrule(lr){2-2} \cmidrule(l){3-7}
        & \text{F1naïf} & 0.019 \pm 0.005       & 0.29 \pm 0.06         & 0.0006 \pm 0.0006     & 0                     & 0.043 \pm 0.007       \\
        & \text{F1sal}  & \bm{0.59 \pm 0.02}    & 0.27 \pm 0.05         & 0.44 \pm 0.03         & 0.19 \pm 0.05         & \bm{0.55 \pm 0.01}    \\
        \bottomrule
    \end{array}
    \]
    \label{tab:time-results}
\end{table*}
\endgroup

\begin{figure*}[ht!]
    \centering
    \begin{subfigure}[b]{0.49\textwidth}
        \centering
        \includegraphics[width=\linewidth]{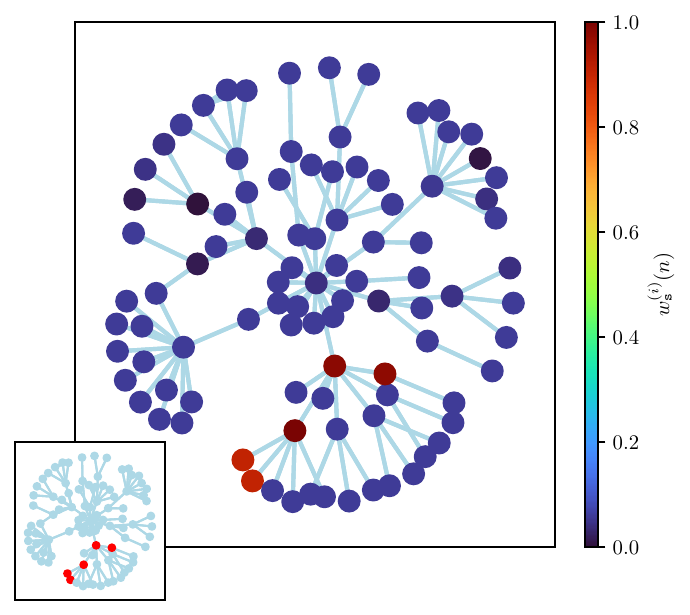}
        \caption{Node mask with $\text{AUC}_\mathcal{G}=1$.}
        \label{fig:node-mask-example}
    \end{subfigure}
    \begin{subfigure}[b]{0.49\textwidth}
        \centering
        \includegraphics[width=\linewidth]{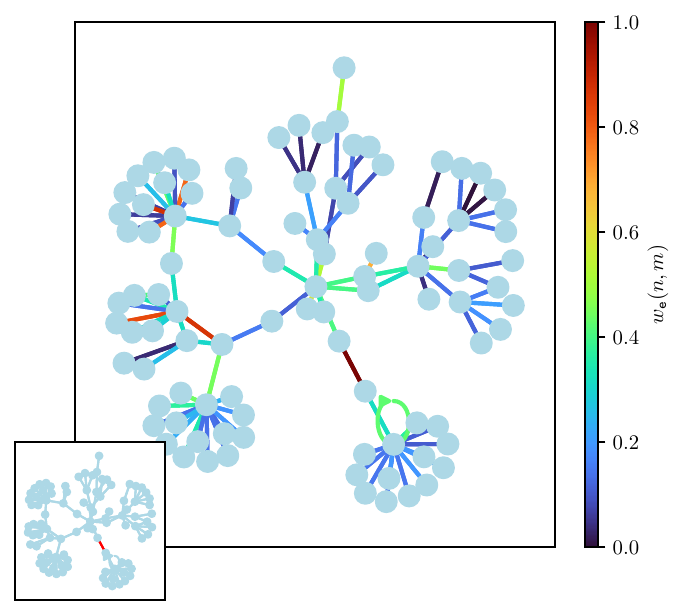}
        \caption{Edges mask with $\text{AUC}_\text{edge}=1$.}
        \label{fig:edge-mask-example}
    \end{subfigure}
    \caption{Spatial explanations from the Facebook dataset. The colour scale on the nodes or edges represents the explanation weights $w_\texttt{s}^{(i)}(n)$ (for nodes) and $w_\texttt{e}(n,m)$ (for edges). The ground truth is reported in the corner.}
    \label{fig:topological-expl}
\end{figure*}

\paragraph{Spatial explanations}
To evaluate hypotheses~\ref{hyp:node-gt} and~\ref{hyp:edge-gt}, we report the AUC scores in Table~\ref{tab:spatial-results}. We also report the AUC scores of the node explanations provided by the saliency map (see \Cref{sec-app:saliency} for more details).
All proposed methods that provide spatial explanations perform consistently well, with some differences across datasets. Even when compared with saliency, our methods almost always perform better than the baseline, demonstrating their effectiveness. 
In Figure~\ref{fig:topological-expl}, we report an example of an explanation on both nodes and edges.

Although the analysis is instance-based, we can leverage our methods to infer something more general about the model's behaviour.
For example, we notice that the weight $w_\mathcal{G}^{(i)}(\timeidx,n)$ effectively recognises whether the $n$-th node is infected or not at time $\timeidx$. This means that the quantity $\sum_n\,w_\mathcal{G}^{(i)}(\timeidx,n)$ is proportional to the number of infected nodes at each time step $\timeidx$. In other words, it reveals a behaviour of the \gls{stgnn} that transcends the specific input, namely that it learns to count infected nodes. 
Even though this information alone is not sufficient to tell the two classes apart, it is an implicit feature that emerges as the model learns to solve the task at hand. Therefore, we argue that the proposed tools can also help interpret model behaviour, not only the input data.

\begingroup
\renewcommand{\arraystretch}{1}
\begin{table*}
    \caption{Results of experiments to test hypotheses~\ref{hyp:node-gt} and~\ref{hyp:edge-gt}. The averages and standard deviations are computed over 5 runs.}
    \centering
    \small
    \[
    \begin{array}{clccccc}
        \toprule
        & \text{Metrics} & \text{Facebook} & \text{Infectious} & \text{DBLP} & \text{Highschool} & \text{Tumblr} \\
        \cmidrule(r){1-1} \cmidrule(lr){2-2} \cmidrule(l){3-7}
        \multirow{3}{*}{\rotatebox[origin=c]{90}{\glsfmtshort{gcrn}}}%
        & \text{AUC}_\text{edge}%
        & 0.923 \pm 0.002 & 0.71 \pm 0.08 & 0.85 \pm 0.03 & 0.73 \pm 0.09 & 0.72 \pm 0.04 \\
        \cmidrule(lr){2-2} \cmidrule(l){3-7}
        & \text{AUC}_\mathcal{G}%
        & 0.849 \pm 0.005 & 0.74 \pm 0.09 & 0.63 \pm 0.03 & 0.66 \pm 0.20 & 0.88 \pm 0.02 \\
        & \text{AUC}_\text{sal}%
        & 0.44 \pm 0.02   & 0.50 \pm 0.25 & 0.81 \pm 0.14 & 0.52 \pm 0.17 & 0.89 \pm 0.05 \\
        \cmidrule(r){1-1} \cmidrule(lr){2-2} \cmidrule(l){3-7}
        \multirow{3}{*}{\rotatebox[origin=c]{90}{\glsfmtshort{dcrnn}}}%
        & \text{AUC}_\text{edge}%
        & 0.84 \pm 0.01   & 0.66 \pm 0.05 & 0.83 \pm 0.02 & 0.67 \pm 0.10 & 0.81 \pm 0.05 \\
        \cmidrule(lr){2-2} \cmidrule(l){3-7}
        & \text{AUC}_\mathcal{G}%
        & 0.859 \pm 0.003 & 0.67 \pm 0.01 & 0.72 \pm 0.04 & 0.62 \pm 0.20 & 0.80 \pm 0.09 \\
        & \text{AUC}_\text{sal}%
        & 0.36 \pm 0.02   & 0.67 \pm 0.37 & 0.16 \pm 0.09 & 0.43 \pm 0.11 & 0.82 \pm 0.09 \\
        \cmidrule(r){1-1} \cmidrule(lr){2-2} \cmidrule(l){3-7}
        \multirow{3}{*}{\rotatebox[origin=c]{90}{\glsfmtshort{gwn}}}%
        & \text{AUC}_\text{edge}%
        & 0.77 \pm 0.03     & 0.59 \pm 0.02   & 0.70 \pm 0.04     & 0.61 \pm 0.05 & 0.68 \pm 0.04 \\
        \cmidrule(lr){2-2} \cmidrule(l){3-7}
        & \text{AUC}_\mathcal{G}%
        & 0.83 \pm 0.02     & 0.69 \pm 0.05   & 0.82 \pm 0.03     & 0.56 \pm 0.11 & 0.73 \pm 0.04 \\
        & \text{AUC}_\text{sal}%
        & 0.15 \pm 0.07     & 0.37 \pm 0.30   & 0.25 \pm 0.24     & 0.12 \pm 0.06 & 0.22 \pm 0.07 \\
        \bottomrule
    \end{array}
    \]
    \label{tab:spatial-results}
\end{table*}
\endgroup

\paragraph{Combined spatiotemporal explanations}
\label{par:st-explanations}

\begin{figure}
    \centering
    \includegraphics[width=1\linewidth]{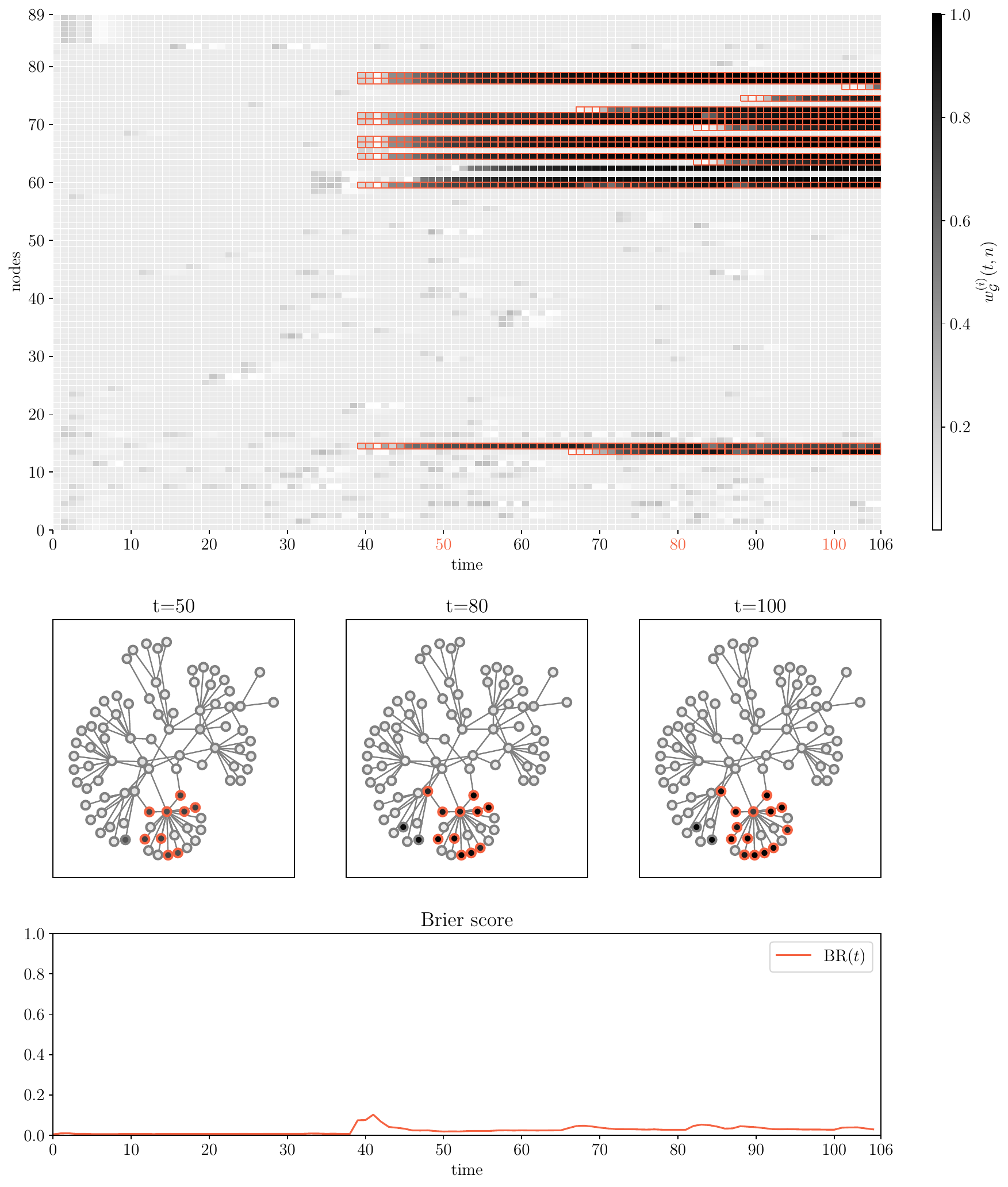}
    \caption{\textit{Top}: The colour scale represents the explanation $w_\mathcal{G}^{(i)}(\timeidx,n)$ for each time step ($x$ axis) and each node ($y$ axis). The red squares mark the entries for which $m_\texttt{st}(\timeidx,n)=1$. \textit{Middle}: The panel shows the \gls{tg} $\mathcal{G}$ at three times, $\timeidx=50,~80,~100$. Nodes in the ground truth are highlighted in red. \textit{Bottom}: The panel shows the value of the Brier score $\text{BS}(\timeidx)$.}
    \label{fig:spatiotemporal-expl}
\end{figure}

We can combine the two approaches and use the spatiotemporal weight $w_\mathcal{G}^{(i)}(\timeidx,n)$ defined in~\eqref{eq:w_G}, comparing it with a spatiotemporal ground truth $m_\texttt{st}(\timeidx,n)$.

To assess the agreement with the ground truth qualitatively, we refer to \Cref{fig:spatiotemporal-expl}, which shows an example of a spatiotemporal explanation for the \gls{gcrn} model from the Facebook dataset. The colour scale in the background represents $w_\mathcal{G}^{(i)}(\timeidx,n)$, and the red boxes indicate the ground truth $m_\texttt{st}^{(i)}(\timeidx,n)$.

To provide a more quantitative measure of the agreement between the spatiotemporal explanation $w_\mathcal{G}^{(i)}(\timeidx,n)$ and the mask $m_\texttt{st}(\timeidx,n)$, one option is to use the Brier score, defined as
\begin{equation}
    \text{BS}(\timeidx):=\frac{1}{|\mathcal{V}|}\sum_{n=1}^{|\mathcal{V}|}\,\left(\frac{w_\mathcal{G}^{(i)}(\timeidx,n)}{\max(w_\mathcal{G}^{(i)}(\timeidx,n))} - m_\texttt{st}(\timeidx,n)\right)^2.
\end{equation}
We choose the Brier score to measure accuracy because it correctly accounts for imbalanced classes and it also provides an easily interpretable outcome, where $\text{BS}(\timeidx)=0$ is the best value and $\text{BS}(\timeidx)=1$ is the worst.
The Brier score is depicted at the bottom of Figure~\ref{fig:spatiotemporal-expl}: the bumps in the plot correspond to the region with more disagreement between the prediction and the ground truth.
In particular, we notice that there is a delay before the explanation registers the infection of a node, and two nodes are false positives, but the Brier score is consistently close to 0.

\paragraph{Qualitative explanations}
\label{par:msrc12-expl}

\begin{figure*}
    \renewcommand*\thesubfigure{\arabic{subfigure}}
    \centering
    \begin{subfigure}[b]{0.05\textwidth}
        \centering
        \includegraphics[height=2.8cm]{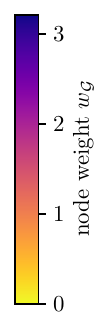}
        \caption*{}
    \end{subfigure}
    \begin{subfigure}[b]{0.078\textwidth}
        \centering
        \includegraphics[width=\linewidth]{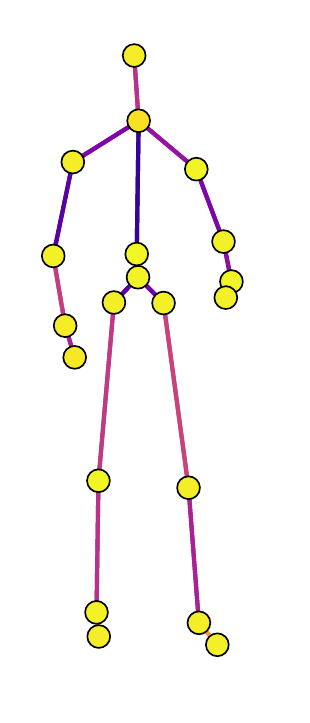}
        \caption{}
    \end{subfigure}
    \begin{subfigure}[b]{0.078\textwidth}
        \centering
        \includegraphics[width=\linewidth]{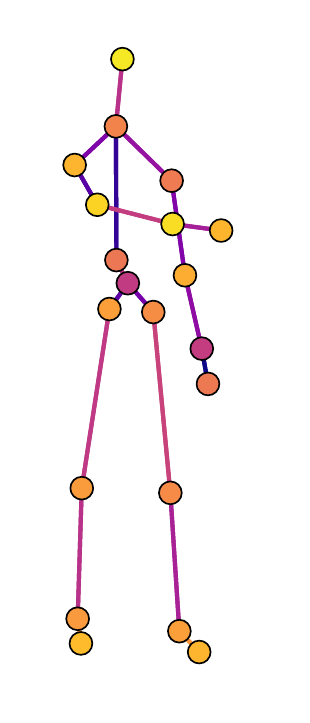}
        \caption{}
    \end{subfigure}
    \begin{subfigure}[b]{0.078\textwidth}
        \centering
        \includegraphics[width=\linewidth]{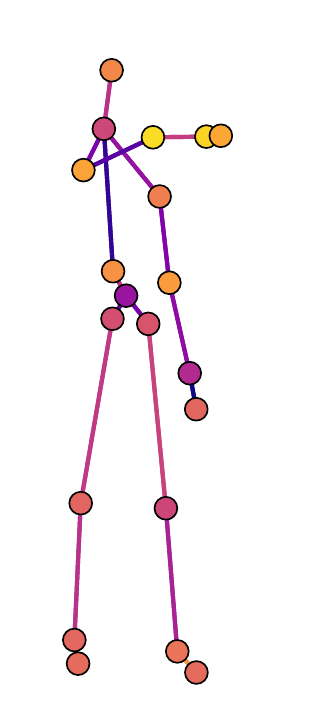}
        \caption{}
    \end{subfigure}
    \begin{subfigure}[b]{0.078\textwidth}
        \centering
        \includegraphics[width=\linewidth]{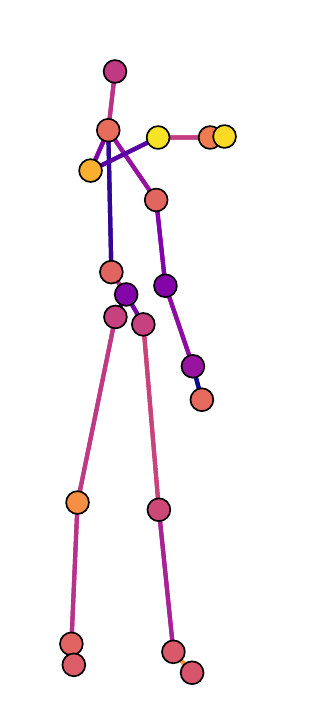}
        \caption{}
    \end{subfigure}
    \begin{subfigure}[b]{0.078\textwidth}
        \centering
        \includegraphics[width=\linewidth]{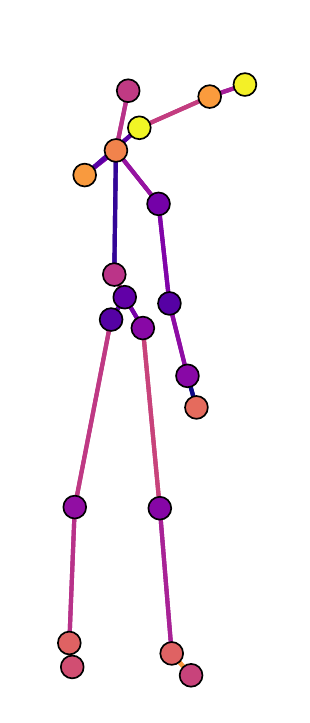}
        \caption{}
    \end{subfigure}
    \begin{subfigure}[b]{0.078\textwidth}
        \centering
        \includegraphics[width=\linewidth]{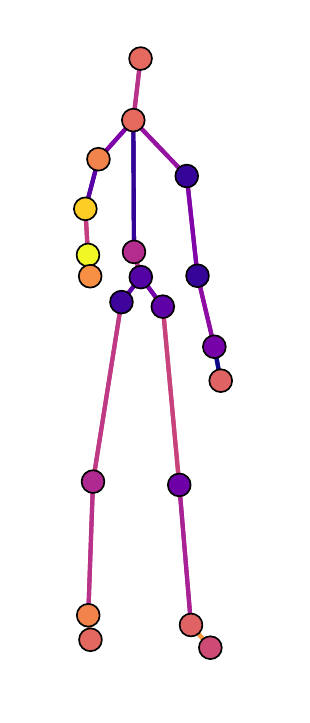}
        \caption{}
    \end{subfigure}
    \begin{subfigure}[b]{0.078\textwidth}
        \centering
        \includegraphics[width=\linewidth]{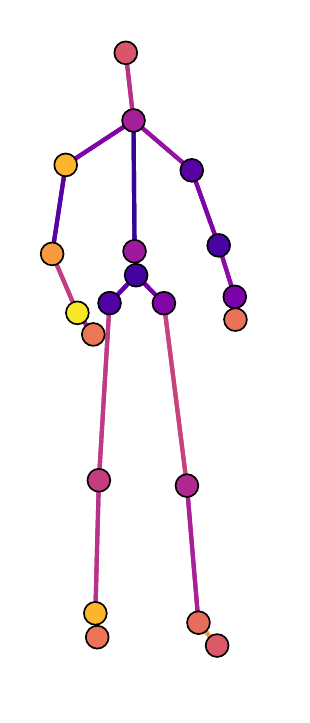}
        \caption{}
    \end{subfigure}
    \begin{subfigure}[b]{0.078\textwidth}
        \centering
        \includegraphics[width=\linewidth]{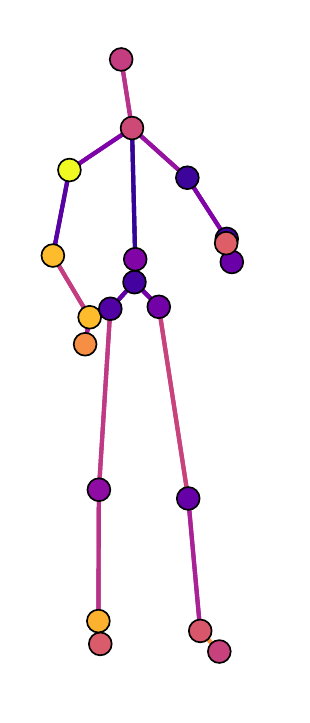}
        \caption{}
    \end{subfigure}
    \begin{subfigure}[b]{0.078\textwidth}
        \centering
        \includegraphics[width=\linewidth]{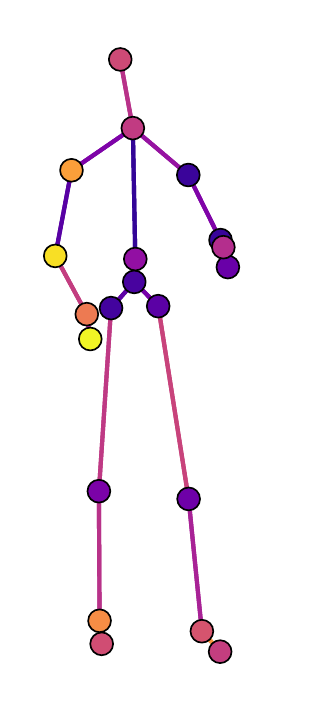}
        \caption{}
    \end{subfigure}
    \begin{subfigure}[b]{0.078\textwidth}
        \centering
        \includegraphics[width=\linewidth]{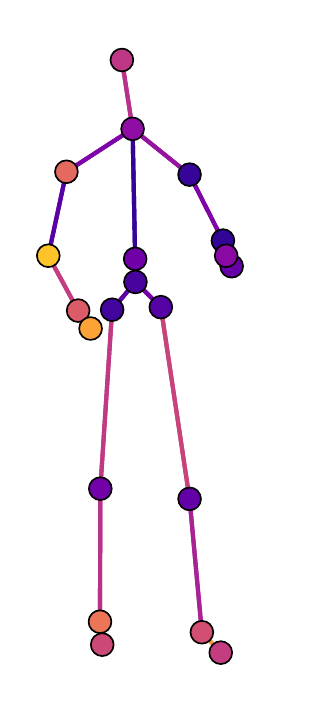}
        \caption{}
    \end{subfigure}
    \begin{subfigure}[b]{0.05\textheight}
        \centering
        \includegraphics[height=2.8cm]{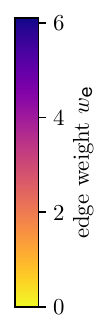}
        \caption*{}
    \end{subfigure}
    \caption{One example from \gls{msrc12} dataset, corresponding to class ``change weapon''.}
    \label{fig:msrc12-change-weapon}

    \vspace{0.5cm}
    \setcounter{subfigure}{0}
    
    \begin{subfigure}[b]{0.05\textwidth}
        \centering
        \includegraphics[height=2.8cm]{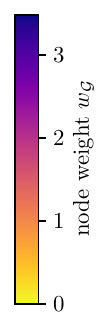}
        \caption*{}
    \end{subfigure}
    \begin{subfigure}[b]{0.078\textwidth}
        \centering
        \includegraphics[width=\linewidth]{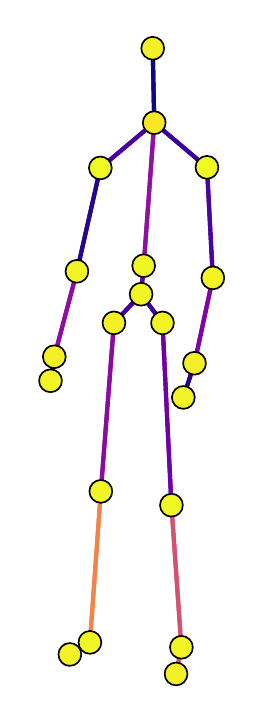}
        \caption{}
    \end{subfigure}
    \begin{subfigure}[b]{0.078\textwidth}
        \centering
        \includegraphics[width=\linewidth, trim=0 0cm 0 0, clip]{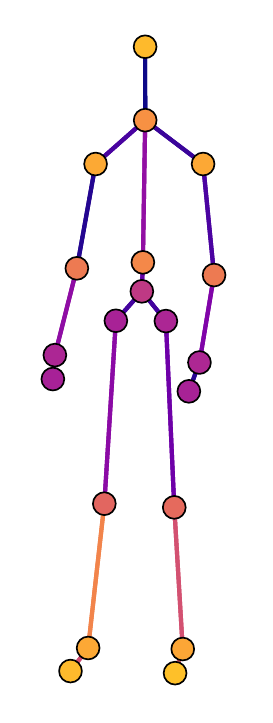}
        \caption{}
    \end{subfigure}
    \begin{subfigure}[b]{0.078\textwidth}
        \centering
        \includegraphics[width=\linewidth, trim=0 0.5cm 0 0, clip]{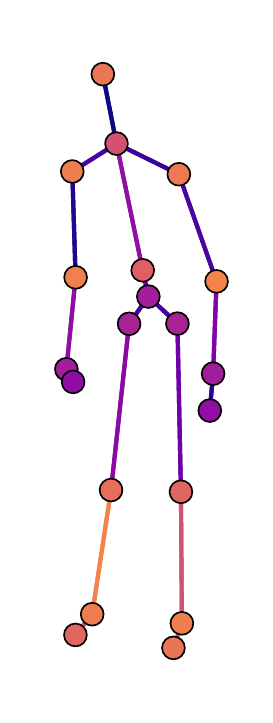}
        \caption{}
    \end{subfigure}
    \begin{subfigure}[b]{0.078\textwidth}
        \centering
        \includegraphics[width=\linewidth, trim=0 1cm 0 0, clip]{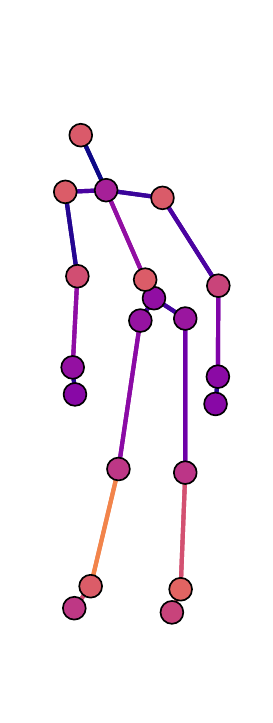}
        \caption{}
    \end{subfigure}
    \begin{subfigure}[b]{0.078\textwidth}
        \centering
        \includegraphics[width=\linewidth, trim=0 1cm 0 0, clip]{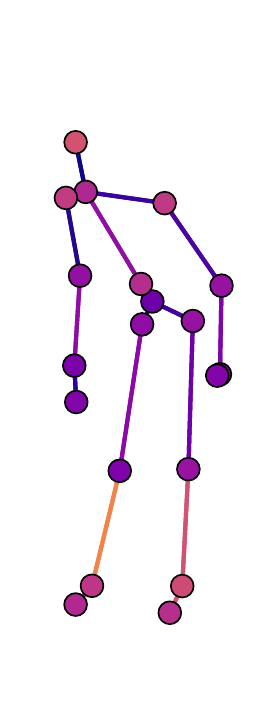}
        \caption{}
    \end{subfigure}
    \begin{subfigure}[b]{0.078\textwidth}
        \centering
        \includegraphics[width=\linewidth, trim=0 1.1cm 0 0, clip]{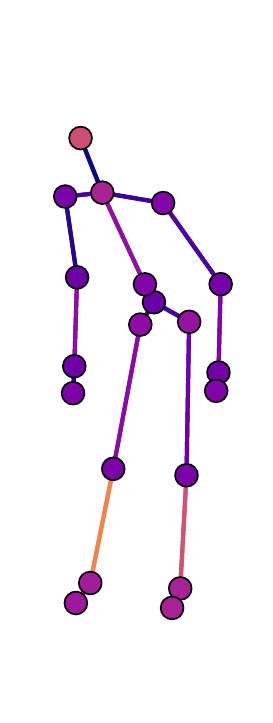}
        \caption{}
    \end{subfigure}
    \begin{subfigure}[b]{0.078\textwidth}
        \centering
        \includegraphics[width=\linewidth, trim=0 0.8cm 0 0, clip]{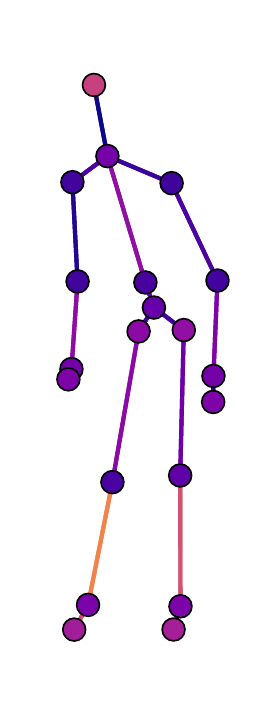}
        \caption{}
    \end{subfigure}
    \begin{subfigure}[b]{0.078\textwidth}
        \centering
        \includegraphics[width=\linewidth, trim=0 0.1cm 0 0, clip]{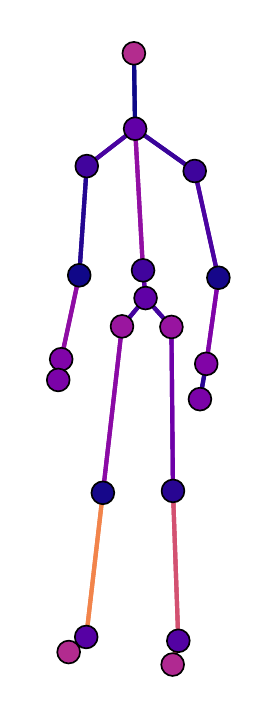}
        \caption{}
    \end{subfigure}
    \begin{subfigure}[b]{0.078\textwidth}
        \centering
        \includegraphics[width=\linewidth, trim=0 0.1cm 0 0, clip]{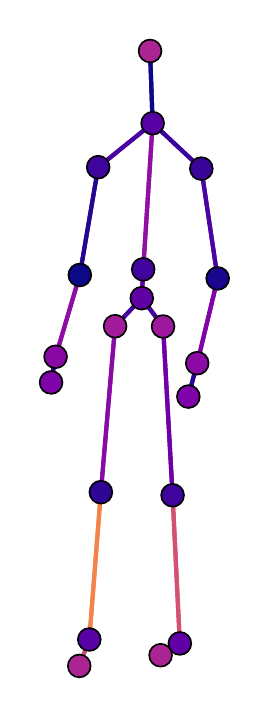}
        \caption{}
    \end{subfigure}
    \begin{subfigure}[b]{0.078\textwidth}
        \centering
        \includegraphics[width=\linewidth, trim=0 0.1cm 0 0, clip]{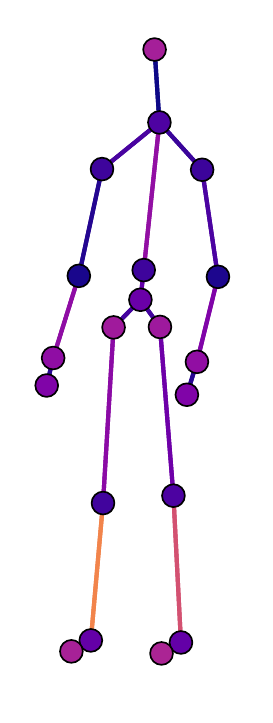}
        \caption{}
    \end{subfigure}
    \begin{subfigure}[b]{0.05\textheight}
        \centering
        \includegraphics[height=2.8cm]{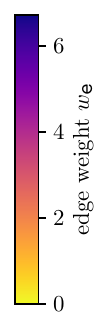}
        \caption*{}
    \end{subfigure}
    \caption{One example from \gls{msrc12} dataset, corresponding to class ``take a bow''.}
    \label{fig:msrc12-bow}

    \vspace{0.5cm}
    \setcounter{subfigure}{0}
    
    \begin{subfigure}[b]{0.05\textwidth}
        \centering
        \includegraphics[height=2.8cm]{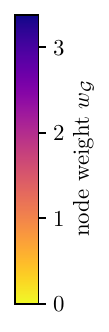}
        \caption*{}
    \end{subfigure}
    \begin{subfigure}[b]{0.078\textwidth}
        \centering
        \includegraphics[width=\linewidth]{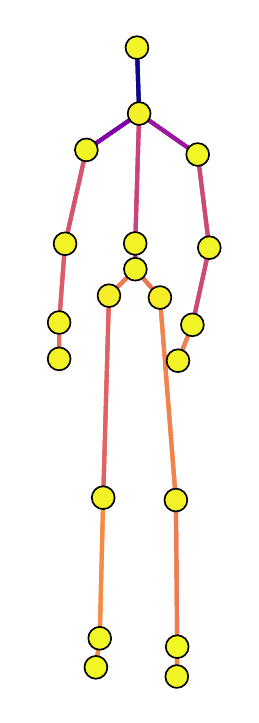}
        \caption{}
    \end{subfigure}
    \begin{subfigure}[b]{0.078\textwidth}
        \centering
        \includegraphics[width=\linewidth, trim=0 0.2cm 0 0, clip]{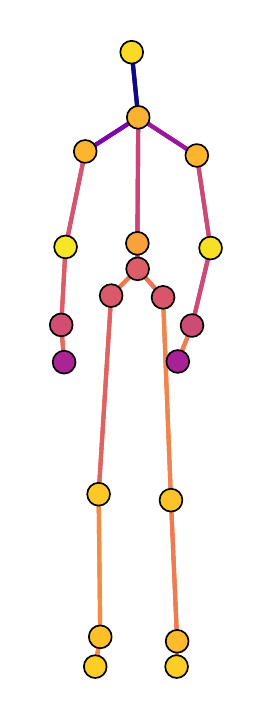}
        \caption{}
    \end{subfigure}
    \begin{subfigure}[b]{0.078\textwidth}
        \centering
        \includegraphics[width=\linewidth, trim=0 1.8cm 0 0, clip]{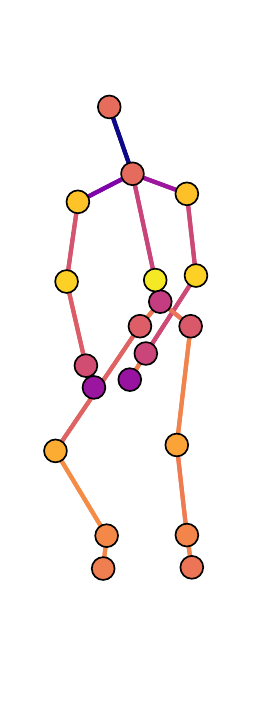}
        \caption{}
    \end{subfigure}
    \begin{subfigure}[b]{0.078\textwidth}
        \centering
        \includegraphics[width=\linewidth, trim=0 2.6cm 0 0, clip]{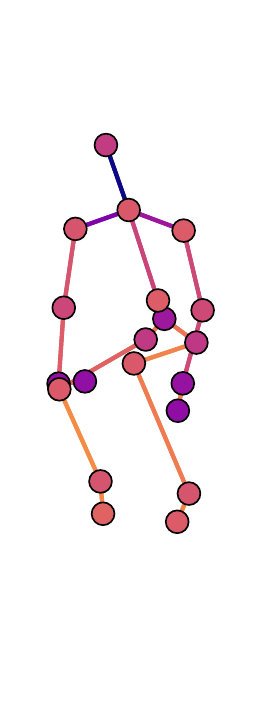}
        \caption{}
    \end{subfigure}
    \begin{subfigure}[b]{0.078\textwidth}
        \centering
        \includegraphics[width=\linewidth, trim=0 2.7cm 0 0, clip]{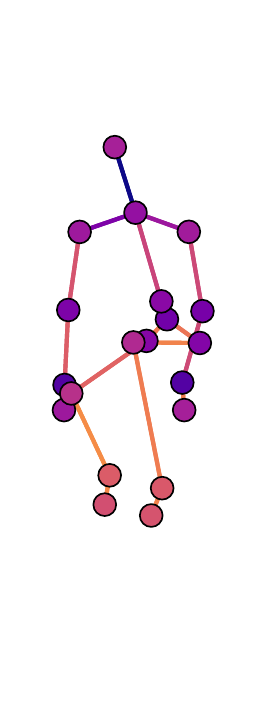}
        \caption{}
    \end{subfigure}
    \begin{subfigure}[b]{0.078\textwidth}
        \centering
        \includegraphics[width=\linewidth, trim=0 2.7cm 0 0, clip]{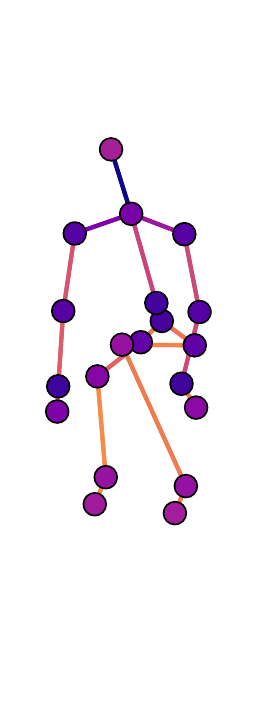}
        \caption{}
    \end{subfigure}
    \begin{subfigure}[b]{0.078\textwidth}
        \centering
        \includegraphics[width=\linewidth, trim=0 2.7cm 0 0, clip]{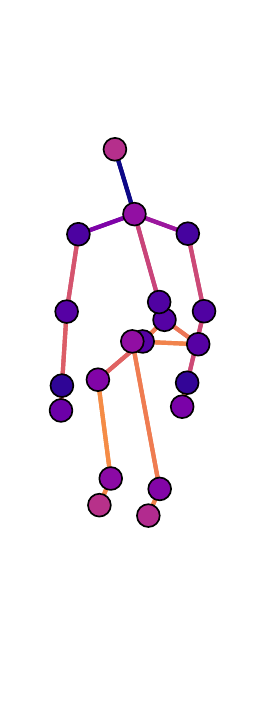}
        \caption{}
    \end{subfigure}
    \begin{subfigure}[b]{0.078\textwidth}
        \centering
        \includegraphics[width=\linewidth, trim=0 2.1cm 0 0, clip]{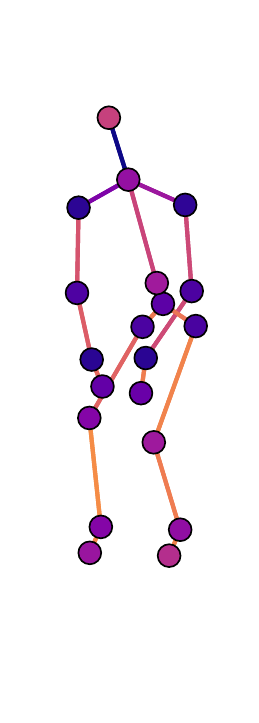}
        \caption{}
    \end{subfigure}
    \begin{subfigure}[b]{0.078\textwidth}
        \centering
        \includegraphics[width=\linewidth, trim=0 0.3cm 0 0, clip]{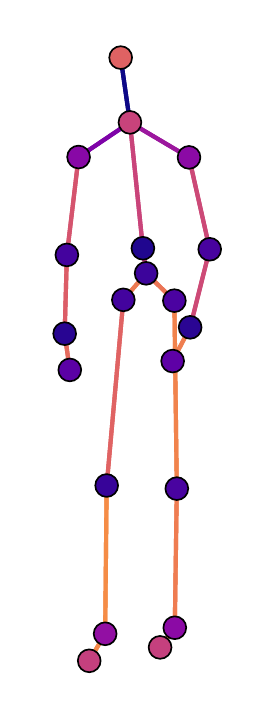}
        \caption{}
    \end{subfigure}
    \begin{subfigure}[b]{0.078\textwidth}
        \centering
        \includegraphics[width=\linewidth, trim=0 0.2cm 0 0, clip]{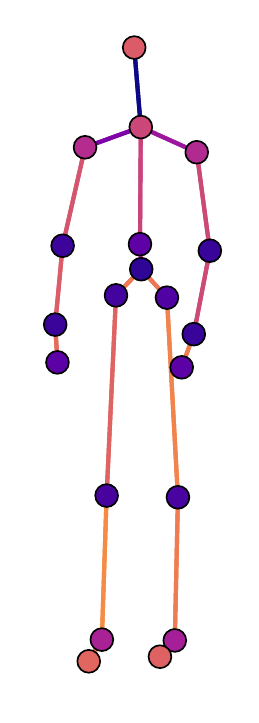}
        \caption{}
    \end{subfigure}
    \begin{subfigure}[b]{0.05\textheight}
        \centering
        \includegraphics[height=2.8cm]{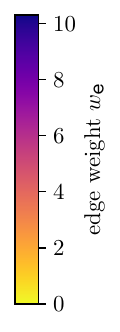}
        \caption*{}
    \end{subfigure}
    \caption{One example from \gls{msrc12} dataset, corresponding to class ``crouch''.}
    \label{fig:msrc12-duck}
\end{figure*}

For the \gls{msrc12} dataset, given the lack of a ground truth (see the discussion in \Cref{sec:metrics}), we rely on a qualitative analysis of the explanations provided by our methods. In \Cref{fig:msrc12-change-weapon,fig:msrc12-bow,fig:msrc12-duck}, we plot 10 frames sampled from three sequences, representing the actions ``change weapon'', ``take a bow'' and ``crouch'', respectively. The colour scale on the left represents the node weight $w_\mathcal{G}(\timeidx,n)$ from equation~\eqref{eq:w_G}, while the colour scale on the right represents the edge weight $w_\texttt{e}(n,m)$ from equation~\eqref{eq:sindy-weights}, whose value is constant in time. For these particular examples, we use \gls{gcrn} as the model, \gls{ttdmd} to compute $w_\mathcal{G}(\timeidx,n)$, and \gls{sindy} with degree 3 to compute $w_\texttt{e}(n,m)$. All the reported instances are correctly classified by the model. Other examples from the remaining classes are reported in \Cref{sec-app:msrc12}.

Here are some qualitative comments on the figures:
\begin{itemize}
    \item \emph{Class ``change weapon''}. In \Cref{fig:msrc12-change-weapon}, we see that the movement involves only the arms: one reaches behind the back to pick up the weapon, and the other holds it. The node explanation shows that the model focuses more on the arm on the right side of the figure, which is consistent with what we see in other samples of the same class. One possible explanation is that the movement of a single arm provides enough discriminative power, given that other classes, e.g.~``protest the music'', also involve arm movements only. The pelvis nodes are also important to the model because of their rotational motion. The edge weights mainly highlight the arms as well.
    \item \emph{Class ``take a bow''}. In \Cref{fig:msrc12-bow}, the explanation is less localised, with more importance given to the upper body and the knees, both in terms of nodes and edges. This makes sense, since the bowing movement involves almost all nodes: other classes involving the movement of most of the body (e.g.~``crouch'' and ``kick'') have more localised explanations (see \Cref{fig:msrc12-duck,fig:msrc12-kick}), so focusing extendendly on more nodes holds enough discriminative power.
    \item \emph{Class ``crouch''}. In \Cref{fig:msrc12-duck}, the model focuses on the lumbar region, the hips and the knees; the arms seem to play a role too, because in other samples in this class the subjects bend their arms and rest their hands on the knees, although that is not the case in this specific instance. The legs have the highest scores in the edge explanation.
\end{itemize}

\section{Conclusion}

In this work, we introduced a Koopman-theoretic perspective on explainability for \glspl{stgnn}. 
By treating the internal embeddings of the model as observables of an underlying dynamical system, we showed that data-driven tools from dynamical systems, namely~\gls{dmd} and~\gls{sindy}, can reveal when relevant events occur, which nodes are most responsible for the prediction, and which interactions are the most influential. 
The experiments on semi-synthetic dissemination datasets, together with the qualitative analysis on~\gls{msrc12}, indicate that this perspective is effective across different \gls{stgnn} architectures and can recover meaningful temporal, spatial, and edge-level explanations.

A key takeaway is that, although \glspl{stgnn} are nonlinear models operating on highly structured inputs, their learned latent dynamics still contain enough regularity for Koopman-inspired analyses to be informative. 
This suggests that explainability for temporal graph models can benefit from dynamical-systems tools, and not only from perturbation-based or saliency-based approaches. 
More broadly, the proposed framework offers a way to connect the internal representations of~\glspl{stgnn} with interpretable phenomena in the input domain, which may be especially relevant in scientific applications where understanding the evolution of the system is as important as obtaining an accurate prediction.

Several directions remain open. 
First, the field would benefit from more principled evaluation protocols for explainability on \glspl{tg}, especially in settings without explanation ground truth. 
Second, it would be natural to extend this perspective beyond classification to tasks such as forecasting and link prediction, and to study whether the same dynamical structures remain explanatory there. 
Third, understanding how dynamical priors should be incorporated into model design and training remains an open question, since the relationship between more structured latent dynamics and better explanations is not yet fully understood.
We hope that this work will encourage further interaction between Koopman theory, system identification, and explainability for graph-based temporal learning.

\nocite{viswanath2009evolution,isella2011s,leskovec2009meme}


%% file: supplement.tex
\section{\Glsfmtlong{ttdmd}}
\label{sec-app:ttdmd}

As mentioned in Section~\ref{sec:dmd}, the standard \gls{dmd} algorithm relies on the vectorisation of the snapshots $\bm{h}_\timeidx$ to build the matrices in~\eqref{eq:snapshot-matrices}, incurring the curse of dimensionality.

Instead of relying on dimensionality reduction techniques like \gls{pca} to mitigate this issue, an alternative is to use a more efficient algebraic representation that leaves the original dimensions intact. One such method is \gls{ttdmd}~\cite{klus2018tensor}, which overcomes high computational costs by exploiting the tensor-train format~\cite{oseledets2011tensor} for the snapshot tensors~\eqref{eq:snapshot-matrices}. Importantly, \gls{ttdmd} does not perform dimensionality reduction; rather, it factorises the data to make full-dimensional computations tractable.

In the tensor-train representation, a tensor $\bm{X}$ of order $d$ is decomposed into a tensor product of $d$ tensors of order at most 3, called \emph{cores}:
\begin{equation}
\label{eq:tt-format}
    \bm{X} = \sum^{r_0}_{k_0=1}\cdots \sum^{r_d}_{k_d =1}\bm{X}^{(1)}_{k_0,:,k_1} \otimes \bm{X}^{(2)}_{k_1,:,k_2} \otimes \dots \otimes \bm{X}^{(d-1)}_{k_{d-2},:,k_{d-1}}\otimes \bm{X}^{(d)}_{k_{d-1},i_d ,k_d},
\end{equation}
or, focusing on the entries,
\begin{equation}
    \bm{X}_{i_1 ,\dots,i_d} =\sum^{r_0}_{k_0=1}\cdots \sum^{r_d}_{k_d =1}\bm{X}^{(1)}_{k_0,i_1,k_1} \cdot \bm{X}^{(2)}_{k_1,i_2,k_2} \cdot \dots \cdot \bm{X}^{(d-1)}_{k_{d-2},i_{d-1},k_{d-1}}\cdot \bm{X}^{(d)}_{k_{d-1},i_d ,k_d}.
\end{equation}
The advantage of this representation is twofold: on one hand, it is possible to rewrite the \gls{dmd} algorithm so that it takes advantage of the tensor-train format. On the other hand, each dimension of the original tensor is stored on a different core and therefore retains its meaning, unlike the standard vectorisation approach, which mixes dimensions.

This approach is related to the \emph{tensor PCA} introduced in~\cite{bianchi2021reservoir} in the context of reservoir computing, where dimensionality reduction is applied to the feature mode of the reservoir states tensor while preserving the temporal structure.

In our setting, we transform the \gls{stgnn}'s embeddings $\bm{h}_\timeidx \in \mathbb{R}^{N\times F}$ into the tensor-train format, and then we apply \gls{ttdmd}.

\section{Description of datasets}
\label{sec-app:datasets}

\begin{table}[b]
    \caption{Description of datasets.}
    \centering
    \begin{tabular}{cccccc}
    \toprule
        Dataset & $N_\mathcal{G}$ & $T$ & $|\mathcal{V}|$ & $|\mathcal{E}|$ & $C$ \\
    \cmidrule(r){1-1} \cmidrule(l){2-6}
        Facebook & 995 & 106 & 71--100 & 176--362 & 2 \\
        Infectious & 200 & 50 & 50 & 218--1010 & 2 \\
        DBLP & 755 & 48 & 50--60 & 96--380 & 2 \\
        Highschool & 180 & 205 & 26--60 & 302--1178 & 2 \\
        Tumblr & 373 & 91 & 25--99 & 96--380 & 2 \\
        \gls{msrc12} & 6243 & 14--493 & 20 & 58 & 12 \\
    \bottomrule
    \end{tabular}
    \label{tab:dataset-description}
\end{table}

The semi-synthetic datasets employed in the experiments consist of \glspl{tg} whose time-varying topologies describe different types of social interactions. The \textit{Facebook} dataset is based on the activity of the New Orleans Facebook community over three months~\cite{viswanath2009evolution}.
The \textit{Infectious} dataset is based on face-to-face contacts between visitors of the SocioPattern project~\cite{isella2011s}.
The \textit{DBLP} dataset is based on co-author graphs from the DBLP database, with publication year used as the timestamp.
The \textit{Highschool} dataset is based on a contact network from the SocioPattern project, describing interactions between high school students over seven days.
The \textit{Tumblr} dataset is based on a graph of quoting interactions between Tumblr users~\cite{leskovec2009meme}.

The original~\gls{msrc12} dataset consists of 594 sequences in which 30 people perform 12 actions, captured at a~\qty[mode=text]{30}{\hertz} sampling rate. Since in each sequence the action is performed multiple times, we preprocess the dataset so that each input sequence corresponds to a single action.

Table~\ref{tab:dataset-description} reports the details of each dataset, such as the number of \glspl{tg} $N_\mathcal{G}$, the range of the length of the temporal sequences $T$, the minimum and maximum number of nodes $|\mathcal{V}|$ and edges $|\mathcal{E}|$, and the number of classes $C$.

\section{Hyperparameters and implementation details}
\label{sec-app:hyperparameters}

The presented methods depend on several hyperparameters. Some of these are used to define the architecture and the training of the \gls{stgnn}, others are involved in the explainability methods. Table~\ref{tab:candidate-hyperparameters} shows all possible values, and \Cref{tab:hyperparameters-gcrnn,tab:hyperparameters-dcrnn,tab:hyperparameters-gwn} report the optimal hyperparameter configurations used for each dataset.
In order to find the best values, we perform a grid search.
For those parameters related to the model's architecture and training, we select the values that yield the highest performance in terms of classification accuracy. For the parameters related to the explainability methods, we use the F1 score defined in Section~\ref{sec:metrics} as the validation metric.

\begingroup
\renewcommand{\arraystretch}{1}
\begin{table}[b]
    \centering
    \caption{Candidate values for hyperparameters.}
    \begin{tabular}{cll}
    \toprule
        & Parameter & Candidates \\
    \cmidrule(r){1-1} \cmidrule(lr){2-2} \cmidrule(lr){3-3}
        \multirow{8}{*}{\rotatebox[origin=c]{90}{STGN par.}} 
        & RNN type          & LSTM, GRU \\ 
        & Activation        & identity, linear, ReLU, leaky ReLU, $\tanh$ \\
        & $F$               & $16$, $32$, $64$ \\ 
        & $L$               & $1, \dots, 10$ \\ 
        & $K$ (\glsfmtshort{dcrnn}, \glsfmtshort{gwn}) & $2,3,4,5$ \\ 
        & $K_\texttt{t}$ (\glsfmtshort{gwn})         & $2,3,4,5$ \\ 
        & Dilation $d$ (\glsfmtshort{gwn})           & $1,2,4$ \\ 
        & MLP layers        & $1, \dots, 5$ \\ 
    \cmidrule(r){1-1} \cmidrule(lr){2-2} \cmidrule(lr){3-3} 
        \multirow{7}{*}{\rotatebox[origin=c]{90}{Expl.~par.}} 
        & Dim.~red.         & PCA, SVD, \glsfmtshort{ttdmd} \\ 
        & $f$               & $10$, $16$, $32$, $64$ \\ 
        & F1 type           & threshold (THR), window (WIN) \\ 
        & Threshold type    & MAX, AVG or MAD \\ 
        & $\omega$          & $2,\dots,6$ \\ 
        & $d_\text{SINDy}$  & $2, 3$ \\ 
        & Mode $i$          & $0, 1$ \\ 
    \bottomrule
    \end{tabular}
    \label{tab:candidate-hyperparameters}
\end{table}
\endgroup

\begin{table}[b]
    \centering
    \small
    \caption{Selected hyperparameters for~\glsfmtshort{gcrn}.}
    \begin{tabular}{ccccccccc}
    \toprule
         & Parameter & Facebook & Infectious & DBLP & Highschool & Tumblr & \glsfmtshort{msrc12} \\
    \cmidrule(r){1-1} \cmidrule(lr){2-2} \cmidrule(l){3-8}
        \multirow{5}{*}{\rotatebox[origin=c]{90}{\glsfmtshort{stgnn} par.}} 
        & \glsfmtshort{rnn} type     & \glsfmtshort{lstm} & LSTM & LSTM & LSTM & LSTM & LSTM \\
        & Activation   & linear & linear & ReLU & ReLU & linear & identity \\
        & $F$          & 64   & 64   & 64   & 32   & 64   & 32 \\
        & $L$          & 2    & 2    & 4    & 3    & 3    & 3 \\
        & \glsfmtshort{mlp} layers%
                       & 1    & 1    & 3    & 4    & 1    & 1 \\
    \cmidrule(r){1-1} \cmidrule(lr){2-2} \cmidrule(l){3-8}
        \multirow{7}{*}{\rotatebox[origin=c]{90}{Expl.~par.}} 
        & Dim.~red.    & PCA  & TT-DMD & TT-DMD & TT-DMD & PCA & TT-DMD \\
        & $f$          & 10   & 10   & 10   & 10   & 10   & 10 \\
        & F1 type      & THR  & THR  & THR  & THR  & WIN  & --- \\
        & Thr.~type    & AVG  & AVG  & MAD  & MAD  & AVG  & --- \\
        & $\omega$     & 5    & 5    & 5    & 5    & 5    & 5 \\
        & $d_\text{SINDy}$ & 2 & 3   & 2    & 3    & 3    & 3 \\
        & Mode $i$     & 0    & 0    & 0    & 0    & 0    & 0 \\
    \bottomrule
    \end{tabular}
    \label{tab:hyperparameters-gcrnn}
\end{table}

\begin{table}[b]
    \centering
    \small
    \caption{Selected hyperparameters for~\glsfmtshort{dcrnn}.}
    \begin{tabular}{ccccccccc}
    \toprule
         & Parameter & Facebook & Infectious & DBLP & Highschool & Tumblr & \glsfmtshort{msrc12} \\
    \cmidrule(r){1-1} \cmidrule(lr){2-2} \cmidrule(l){3-8}
        \multirow{6}{*}{\rotatebox[origin=c]{90}{\glsfmtshort{stgnn} par.}} 
        & \glsfmtshort{rnn} type & LSTM & LSTM & LSTM & LSTM & LSTM & LSTM \\
        & Activation   & ReLU  & $\tanh$ & ReLU & linear & linear & leaky ReLU \\
        & $F$          & 32    & 32   &  64  & 16   & 32   & 64 \\
        & $K$          & 2     & 2    &  2   & 2    & 5    & 5 \\
        & $L$          & 1     & 1    &  1   & 1    & 1    & 2 \\
        & \glsfmtshort{mlp} layers & 1 & 3 & 2 & 2  & 3    & 1 \\
    \cmidrule(r){1-1} \cmidrule(lr){2-2} \cmidrule(l){3-8}
        \multirow{7}{*}{\rotatebox[origin=c]{90}{Expl.~par.}} 
        & Dim.~red.    & PCA & TT-DMD & TT-DMD & TT-DMD & PCA  & TT-DMD \\
        & $f$          & 10    & 10   & 10   & 10   & 10   & 10 \\
        & F1 type      & THR   & THR  & THR  & THR  & WIN  & ---  \\
        & Thr.~type    & AVG   & AVG  & MAD  & MAD  & AVG  & --- \\
        & $\omega$     & 5     & 5    & 5    & 5    & 5    & 5 \\
        & $d_\text{SINDy}$ & 2 & 3    & 3    & 3    & 3    & 3 \\
        & Mode $i$     & 0     & 0    & 0    & 0    & 0    & 0 \\
    \bottomrule
    \end{tabular}
    \label{tab:hyperparameters-dcrnn}
\end{table}

\begin{table}[b]
    \centering
    \small
    \caption{Selected hyperparameters for~\glsfmtshort{gwn}.}
    \begin{tabular}{ccccccccc}
    \toprule
         & Parameter & Facebook & Infectious & DBLP & Highschool & Tumblr & \glsfmtshort{msrc12} \\
    \cmidrule(r){1-1} \cmidrule(lr){2-2} \cmidrule(l){3-8}
        \multirow{7}{*}{\rotatebox[origin=c]{90}{\glsfmtshort{stgnn} par.}} 
        & Activation   & leaky ReLU & linear & leaky ReLU & linear & leaky ReLU & ReLU \\
        & $F$          & 16   & 32   & 64   & 64   & 64   & 32 \\
        & $L$          & 8    & 5    & 4    & 4    & 5    & 5 \\
        & $K$          & 3    & 4    & 4    & 2    & 4    & 3 \\
        & $K_\texttt{t}$ & 5  & 3    & 2    & 5    & 5    & 5 \\
        & $d$          & 2    & 4    & 1    & 4    & 4    & 4 \\
        & \glsfmtshort{mlp} layers %
                       & 3    & 3    & 3    & 3    & 2    & 1 \\
    \cmidrule(r){1-1} \cmidrule(lr){2-2} \cmidrule(l){3-8}
        \multirow{7}{*}{\rotatebox[origin=c]{90}{Expl.~par.}} 
        & Dim.~red.    & PCA  & PCA  & PCA  & TT-DMD  & PCA  & TT-DMD \\
        & $f$          & 10   & 10   & 10   & 10   & 10   & 10 \\
        & F1 type      & WIN  & THR  & THR  & WIN  & WIN  & --- \\
        & Thr.~type    & AVG  & MAD  & MAD  & AVG  & AVG  & --- \\
        & $\omega$     & 5    & 5    & 5    & 5    & 5    & 5 \\
        & $d_\text{SINDy}$ & 3 & 3   & 3    & 3    & 3    & 3 \\
        & Mode $i$     & 0    & 0    & 0    & 0    & 0    & 0 \\
    \bottomrule
    \end{tabular}
    \label{tab:hyperparameters-gwn}
\end{table}

\section{Saliency baselines}
\label{sec-app:saliency}

To strengthen the empirical comparison, we compare our results with explanations produced by a saliency map~\cite{Simonyan14a}.
We use a standard saliency map to find the nodes and times of the input~\gls{tg} that are most relevant to the model. The saliency map provides a saliency attribute
\begin{equation}
    \label{eq:saliency-attribute}
    S(n,\timeidx) := \left| \nabla_{\bm{x}_{\timeidx,n}} y \right|,
\end{equation}
where $y$ is the output of the model~\eqref{eq:gcrn-output}, and $\bm{x}_{\timeidx,n}$ is the input label of the $n$-th node. Notice that we only consider node labels, so we don't compute a saliency attribute for the input's edges.
We can define a temporal explanation as $w_\texttt{t}(\timeidx)=\sum_n S(n,\timeidx)$ and, as done before, we can measure the F1 score by highlighting those time steps $\timeidx$ such that $w_\texttt{t}(\timeidx)>\delta$, where we use \gls{mad} as threshold.
The use of \gls{mad} is necessary due to the shape of $w_\texttt{t}(\timeidx)$, which shows some peaks much more prominent than others, and the other methods would hinder weaker, but relevant, peaks.

The spatial explanation is defined as
\begin{equation}
    w_\texttt{s}(n) := \max_\timeidx\,\left| S(n,\timeidx) \right|.
\end{equation}

The measured values of the F1 scores and $\text{AUC}$ are reported in \Cref{tab:time-results,tab:spatial-results}, referred to as $\text{F1sal}$ and $\text{AUC}_\text{sal}$, and they offer a baseline for the metrics $\text{F1}$ and $\text{AUC}_\mathcal{G}$ respectively.

\section{Further examples from \glsfmtshort{msrc12}}
\label{sec-app:msrc12}

\begin{figure*}
    \renewcommand*\thesubfigure{\arabic{subfigure}}
    \centering
    \begin{subfigure}[b]{0.05\textwidth}
        \centering
        \includegraphics[height=2.8cm]{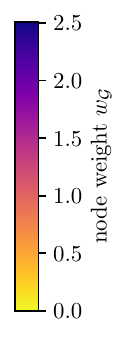}
        \caption*{}
    \end{subfigure}
    \begin{subfigure}[b]{0.077\textwidth}
        \centering
        \includegraphics[width=\linewidth, trim=1.5cm 1.5cm 1.5cm 0, clip]{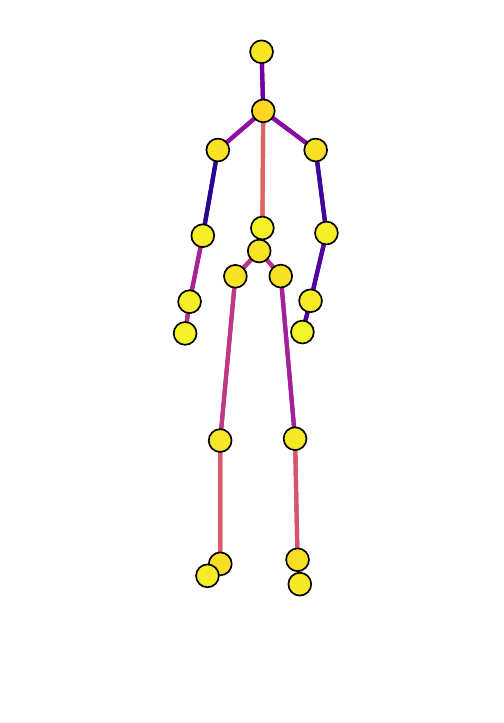}
        \caption{}
    \end{subfigure}
    \begin{subfigure}[b]{0.077\textwidth}
        \centering
        \includegraphics[width=\linewidth, trim=1.5cm 1.5cm 1.5cm 0, clip]{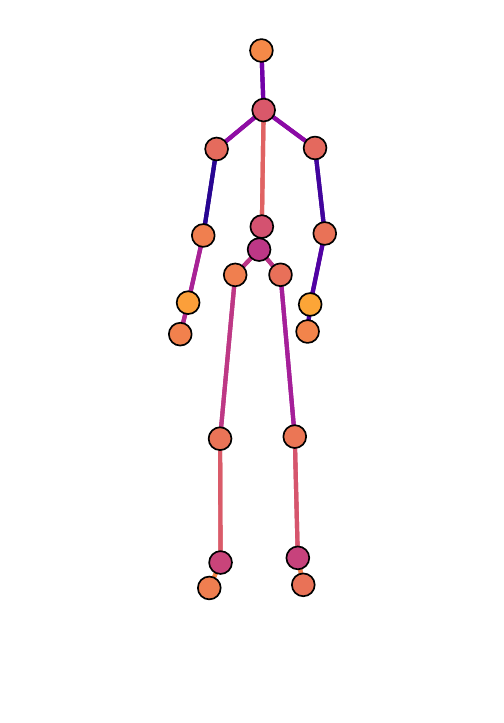}
        \caption{}
    \end{subfigure}
    \begin{subfigure}[b]{0.077\textwidth}
        \centering
        \includegraphics[width=\linewidth, trim=1.5cm 1.4cm 1.5cm 0, clip]{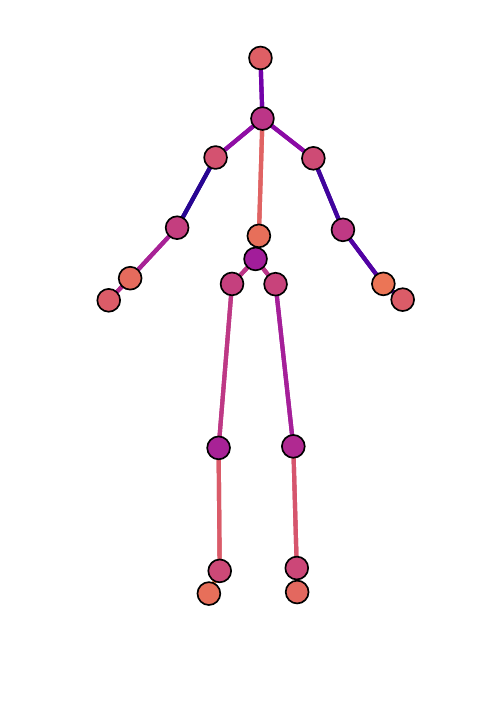}
        \caption{}
    \end{subfigure}
    \begin{subfigure}[b]{0.096\textwidth}
        \centering
        \includegraphics[width=\linewidth, trim=0.7cm 0.6cm 0.7cm 0, clip]{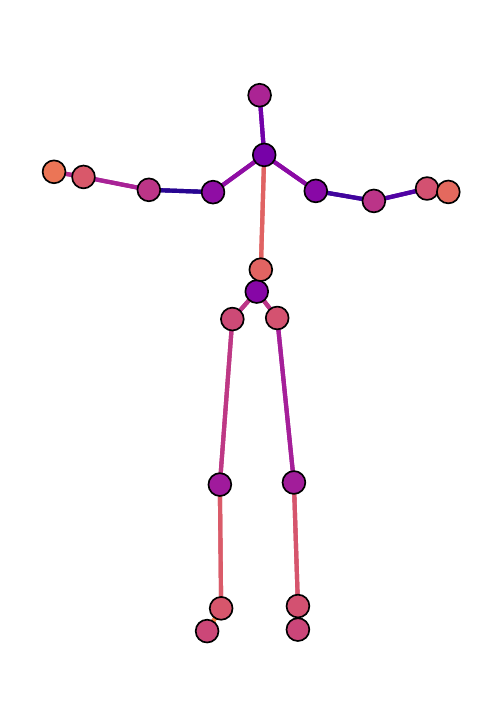}
        \caption{}
    \end{subfigure}
    \begin{subfigure}[b]{0.077\textwidth}
        \centering
        \includegraphics[width=\linewidth, trim=1.5cm 0cm 1.4cm 0, clip]{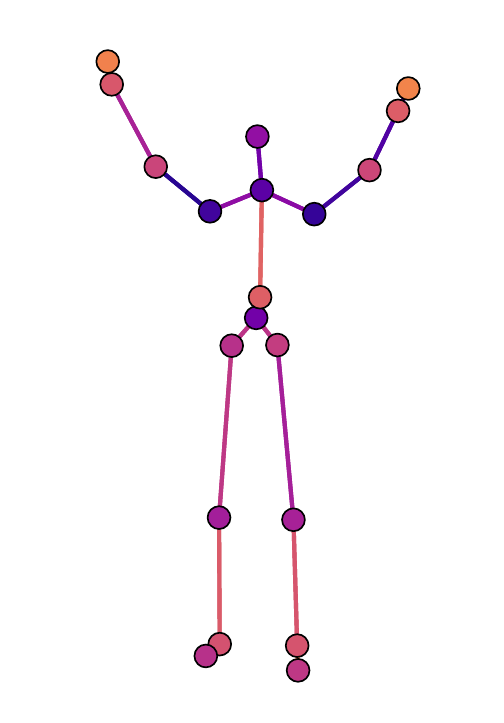}
        \caption{}
    \end{subfigure}
    \begin{subfigure}[b]{0.077\textwidth}
        \centering
        \includegraphics[width=\linewidth, trim=1.5cm 0cm 1.5cm 0, clip]{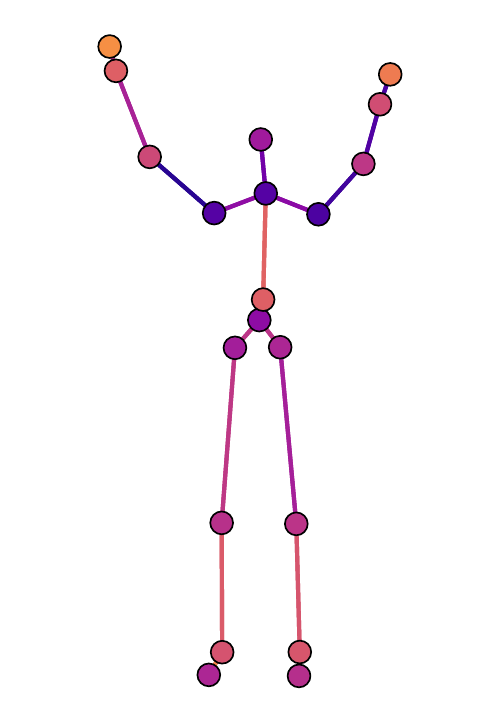}
        \caption{}
    \end{subfigure}
    \begin{subfigure}[b]{0.077\textwidth}
        \centering
        \includegraphics[width=\linewidth, trim=1.5cm 0.5cm 1.5cm 0, clip]{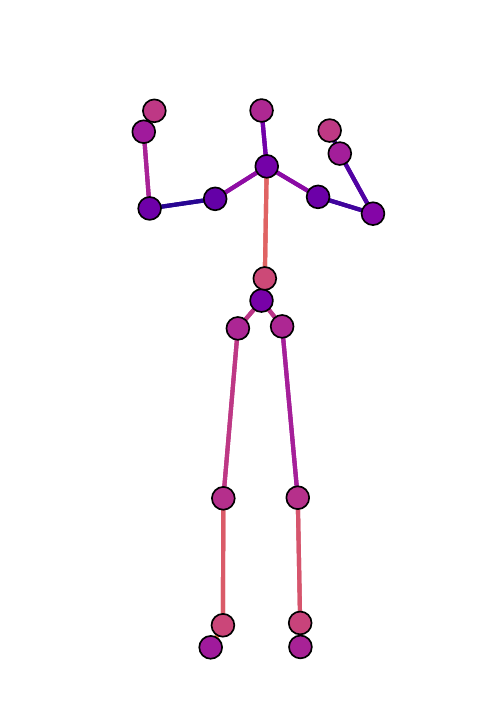}
        \caption{}
    \end{subfigure}
    \begin{subfigure}[b]{0.077\textwidth}
        \centering
        \includegraphics[width=\linewidth, trim=1.5cm 1.3cm 1.5cm 0, clip]{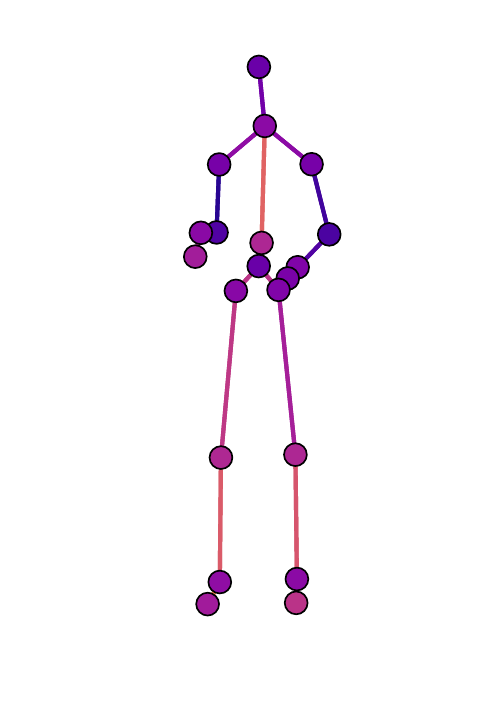}
        \caption{}
    \end{subfigure}
    \begin{subfigure}[b]{0.077\textwidth}
        \centering
        \includegraphics[width=\linewidth, trim=1.5cm 1.5cm 1.5cm 0, clip]{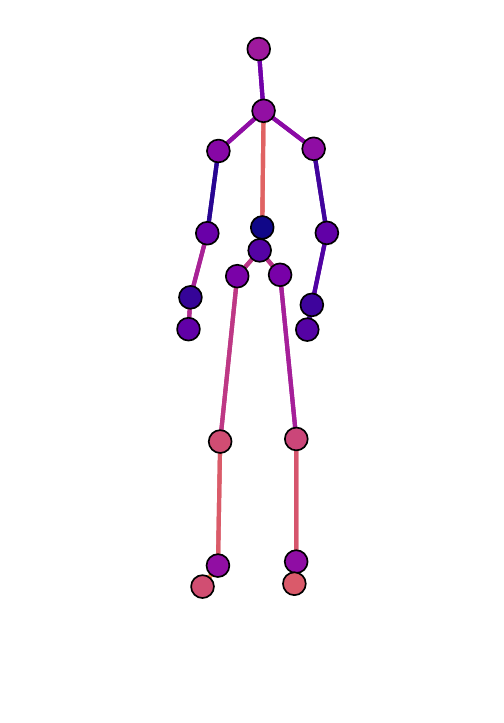}
        \caption{}
    \end{subfigure}
    \begin{subfigure}[b]{0.077\textwidth}
        \centering
        \includegraphics[width=\linewidth, trim=1.5cm 1.5cm 1.5cm 0, clip]{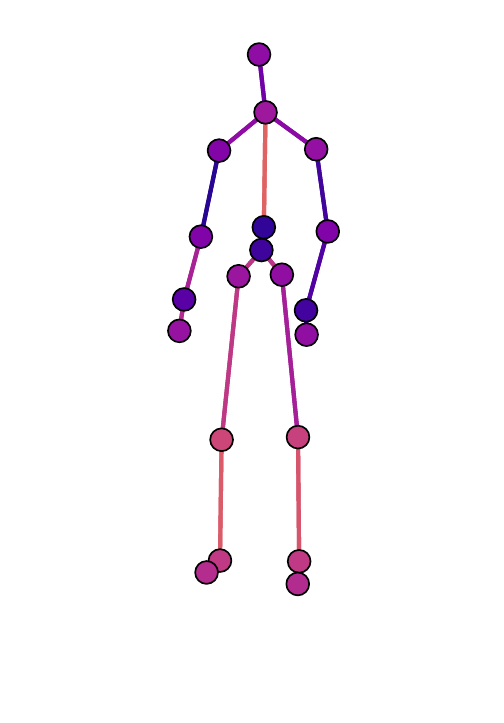}
        \caption{}
    \end{subfigure}
    \begin{subfigure}[b]{0.05\textheight}
        \centering
        \includegraphics[height=2.8cm]{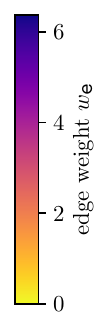}
        \caption*{}
    \end{subfigure}
    \caption{One example from the \gls{msrc12} dataset, corresponding to the class ``raise volume of music''.}
    \label{fig:msrc12-raise-volume}

    \vspace{0.5cm}
    \setcounter{subfigure}{0}
    
    \begin{subfigure}[b]{0.05\textwidth}
        \centering
        \includegraphics[height=2.8cm]{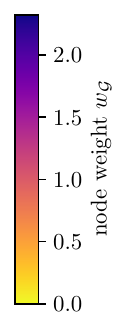}
        \caption*{}
    \end{subfigure}
    \begin{subfigure}[b]{0.078\textwidth}
        \centering
        \includegraphics[width=\linewidth, trim=1cm 0cm 0.5cm 0, clip]{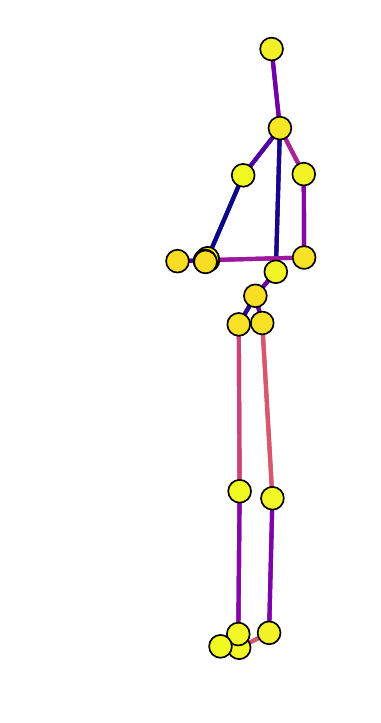}
        \caption{}
    \end{subfigure}
    \begin{subfigure}[b]{0.078\textwidth}
        \centering
        \includegraphics[width=\linewidth, trim=1cm 0cm 0.5cm 0, clip]{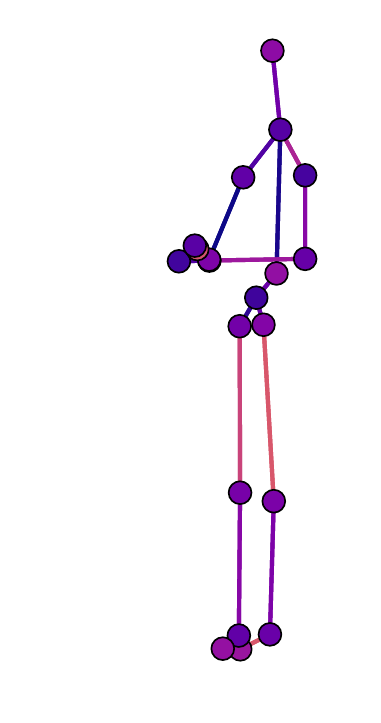}
        \caption{}
    \end{subfigure}
    \begin{subfigure}[b]{0.078\textwidth}
        \centering
        \includegraphics[width=\linewidth, trim=1cm 0cm 0.5cm 0, clip]{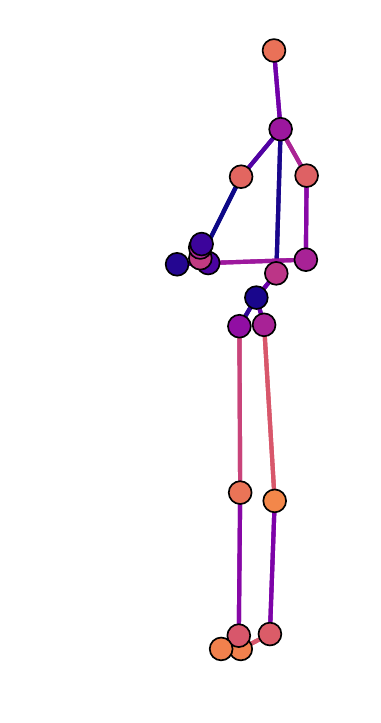}
        \caption{}
    \end{subfigure}
    \begin{subfigure}[b]{0.078\textwidth}
        \centering
        \includegraphics[width=\linewidth, trim=1cm 0cm 0.5cm 0, clip]{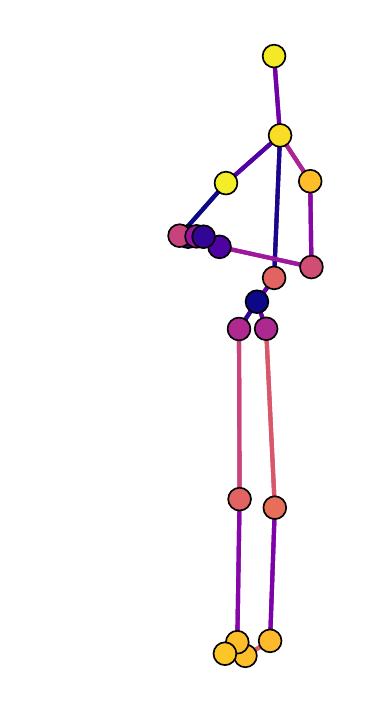}
        \caption{}
    \end{subfigure}
    \begin{subfigure}[b]{0.078\textwidth}
        \centering
        \includegraphics[width=\linewidth, trim=1cm 0cm 0.5cm 0, clip]{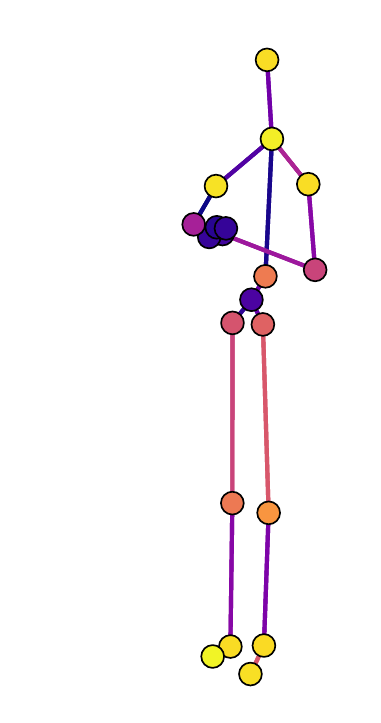}
        \caption{}
    \end{subfigure}
    \begin{subfigure}[b]{0.078\textwidth}
        \centering
        \includegraphics[width=\linewidth, trim=1cm 0cm 0.5cm 0, clip]{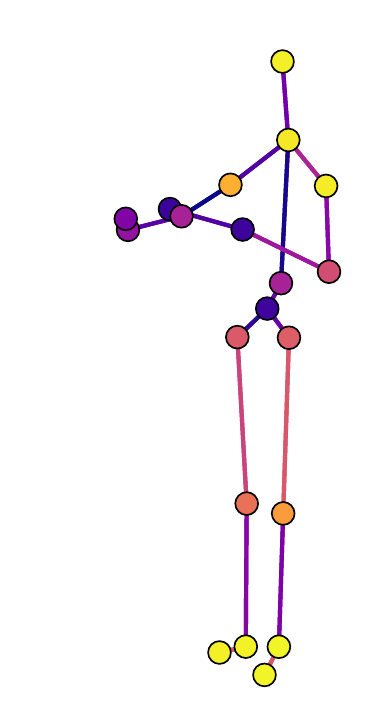}
        \caption{}
    \end{subfigure}
    \begin{subfigure}[b]{0.078\textwidth}
        \centering
        \includegraphics[width=\linewidth, trim=1cm 0cm 0.5cm 0, clip]{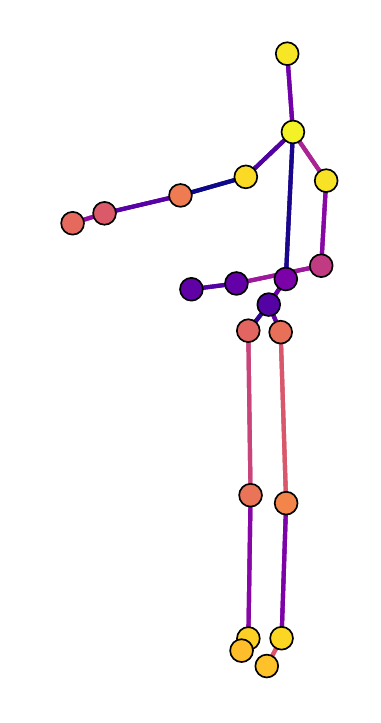}
        \caption{}
    \end{subfigure}
    \begin{subfigure}[b]{0.078\textwidth}
        \centering
        \includegraphics[width=\linewidth, trim=0.9cm 0cm 0.5cm 0, clip]{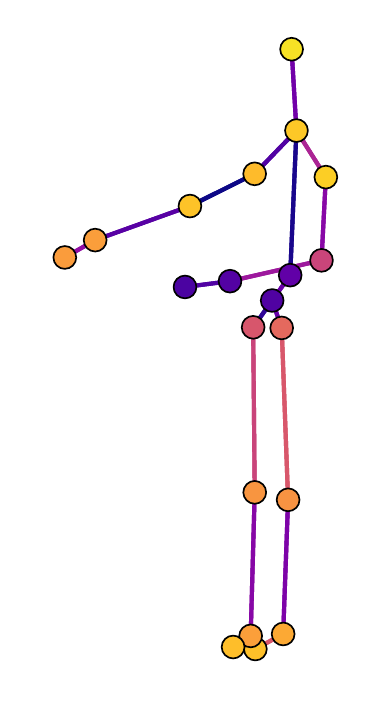}
        \caption{}
    \end{subfigure}
    \begin{subfigure}[b]{0.078\textwidth}
        \centering
        \includegraphics[width=\linewidth, trim=1cm 0cm 0.5cm 0, clip]{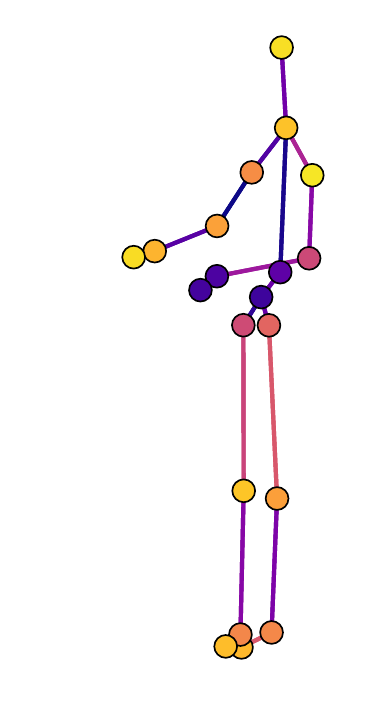}
        \caption{}
    \end{subfigure}
    \begin{subfigure}[b]{0.078\textwidth}
        \centering
        \includegraphics[width=\linewidth, trim=1cm 0cm 0.5cm 0, clip]{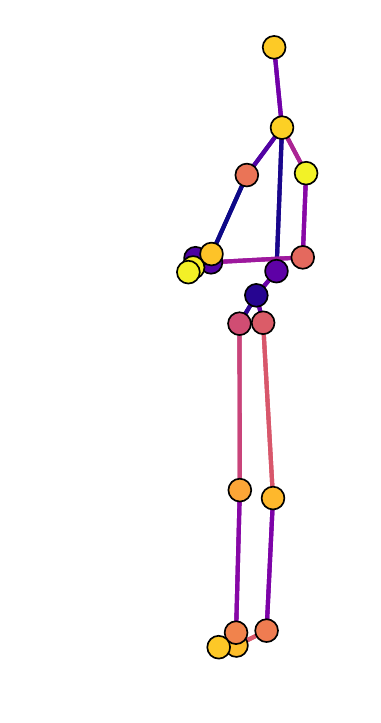}
        \caption{}
    \end{subfigure}
    \begin{subfigure}[b]{0.05\textheight}
        \centering
        \includegraphics[height=2.8cm]{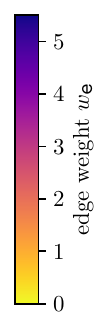}
        \caption*{}
    \end{subfigure}
    \caption{One example from the \gls{msrc12} dataset, corresponding to the class ``navigate to next menu''.}
    \label{fig:msrc12-push-right}

    \vspace{0.5cm}
    \setcounter{subfigure}{0}
    
    \begin{subfigure}[b]{0.05\textwidth}
        \centering
        \includegraphics[height=2.8cm]{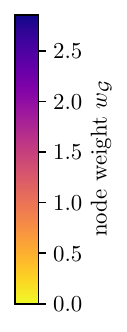}
        \caption*{}
    \end{subfigure}
    \begin{subfigure}[b]{0.078\textwidth}
        \centering
        \includegraphics[width=\linewidth, trim=0 1cm 0 0, clip]{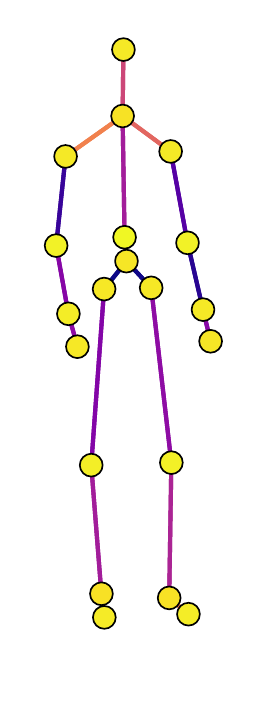}
        \caption{}
    \end{subfigure}
    \begin{subfigure}[b]{0.078\textwidth}
        \centering
        \includegraphics[width=\linewidth, trim=0 1cm 0 0, clip]{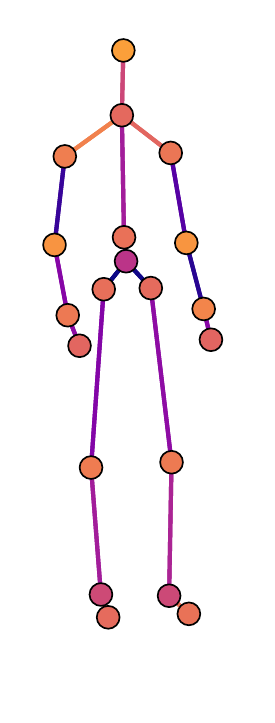}
        \caption{}
    \end{subfigure}
    \begin{subfigure}[b]{0.078\textwidth}
        \centering
        \includegraphics[width=\linewidth, trim=0 0.9cm 0 0, clip]{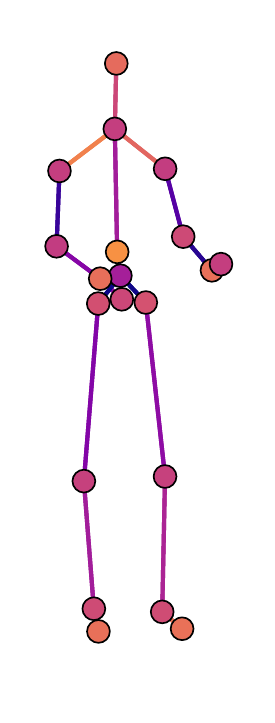}
        \caption{}
    \end{subfigure}
    \begin{subfigure}[b]{0.078\textwidth}
        \centering
        \includegraphics[width=\linewidth, trim=0 0cm 0 0, clip]{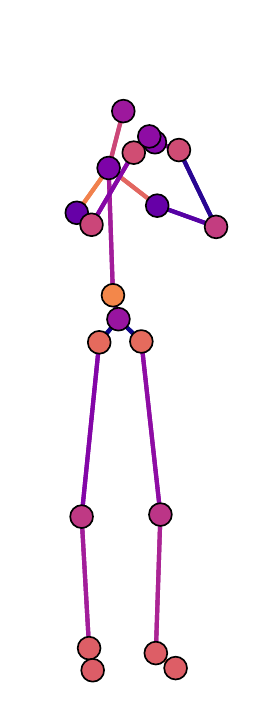}
        \caption{}
    \end{subfigure}
    \begin{subfigure}[b]{0.078\textwidth}
        \centering
        \includegraphics[width=\linewidth, trim=0 0cm 0 0, clip]{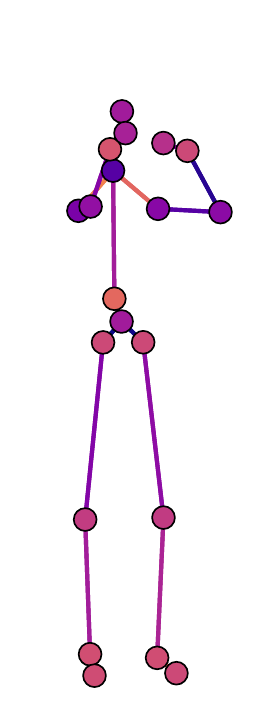}
        \caption{}
    \end{subfigure}
    \begin{subfigure}[b]{0.078\textwidth}
        \centering
        \includegraphics[width=\linewidth, trim=0 0cm 0 0, clip]{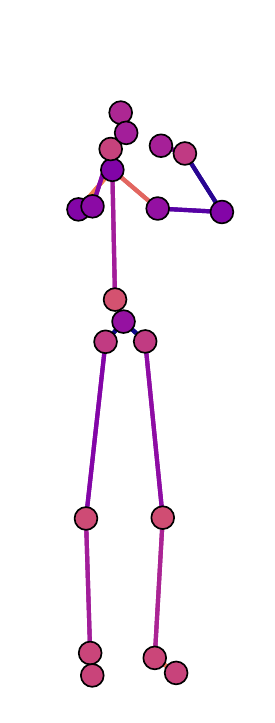}
        \caption{}
    \end{subfigure}
    \begin{subfigure}[b]{0.078\textwidth}
        \centering
        \includegraphics[width=\linewidth, trim=0 0cm 0 0, clip]{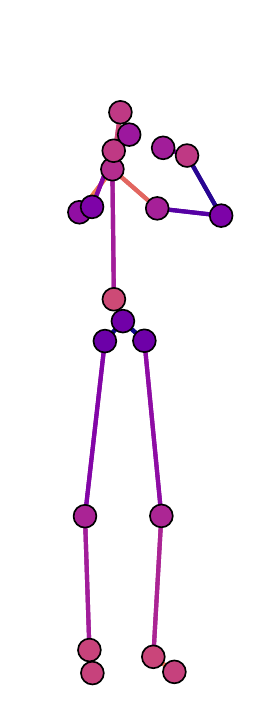}
        \caption{}
    \end{subfigure}
    \begin{subfigure}[b]{0.078\textwidth}
        \centering
        \includegraphics[width=\linewidth, trim=0 0.7cm 0 0, clip]{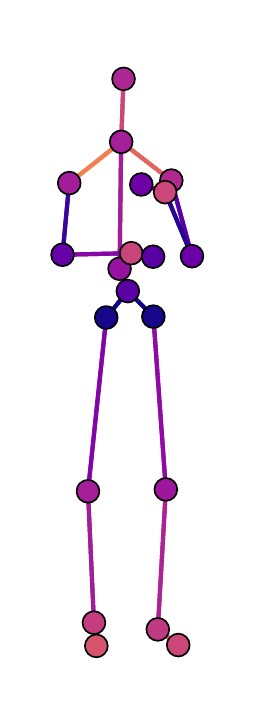}
        \caption{}
    \end{subfigure}
    \begin{subfigure}[b]{0.078\textwidth}
        \centering
        \includegraphics[width=\linewidth, trim=0 1cm 0 0, clip]{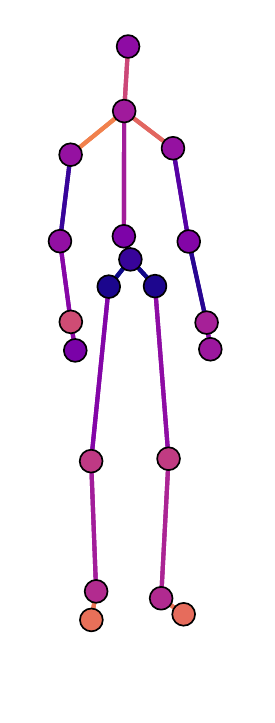}
        \caption{}
    \end{subfigure}
    \begin{subfigure}[b]{0.078\textwidth}
        \centering
        \includegraphics[width=\linewidth, trim=0 1cm 0 0, clip]{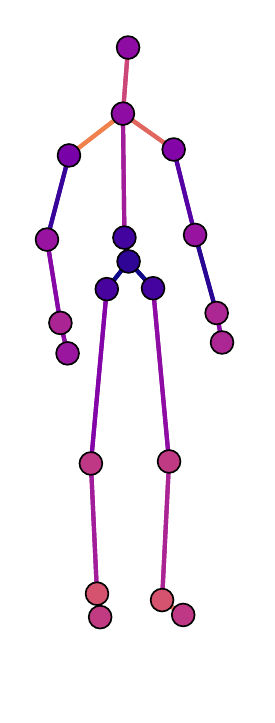}
        \caption{}
    \end{subfigure}
    \begin{subfigure}[b]{0.05\textheight}
        \centering
        \includegraphics[height=2.8cm]{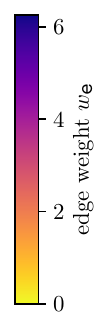}
        \caption*{}
    \end{subfigure}
    \caption{One example from the \gls{msrc12} dataset, corresponding to the class ``put on goggles''.}
    \label{fig:msrc12-goggles}

    \vspace{0.5cm}
    \setcounter{subfigure}{0}
    
    \begin{subfigure}[b]{0.05\textwidth}
        \centering
        \includegraphics[height=2.8cm]{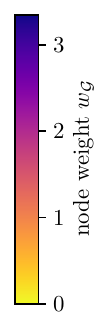}
        \caption*{}
    \end{subfigure}
    \begin{subfigure}[b]{0.073\textwidth}
        \centering
        \includegraphics[width=\linewidth, trim=1.8cm 0.9cm 1.8cm 0, clip]{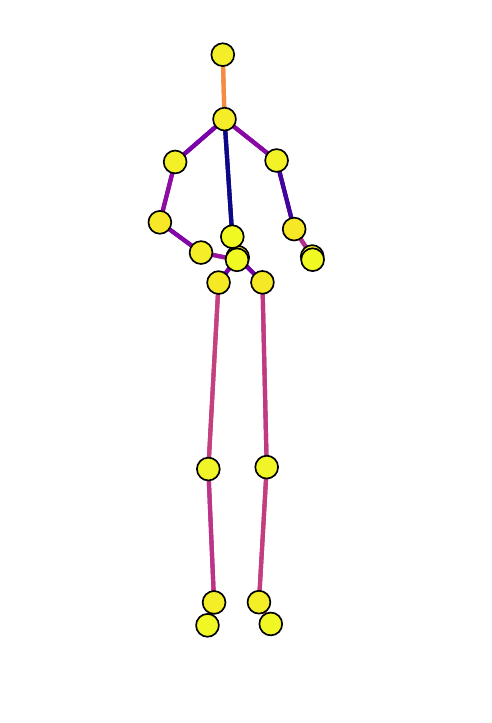}
        \caption{}
    \end{subfigure}
    \begin{subfigure}[b]{0.073\textwidth}
        \centering
        \includegraphics[width=\linewidth, trim=1.8cm 0.9cm 1.8cm 0, clip]{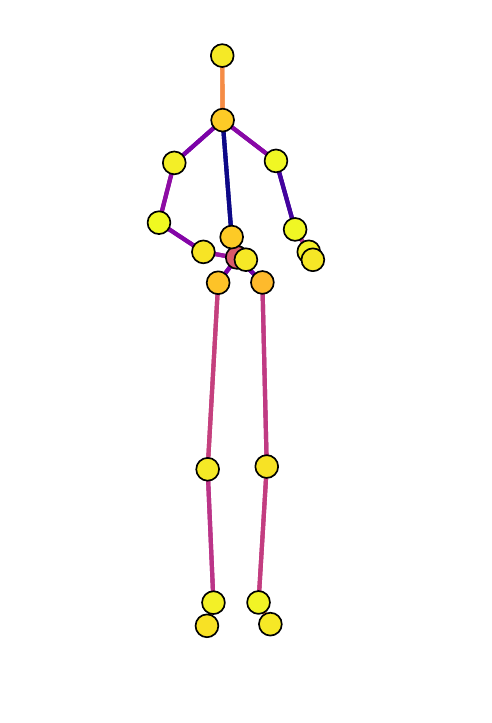}
        \caption{}
    \end{subfigure}
    \begin{subfigure}[b]{0.073\textwidth}
        \centering
        \includegraphics[width=\linewidth, trim=1.8cm 0.9cm 1.8cm 0, clip]{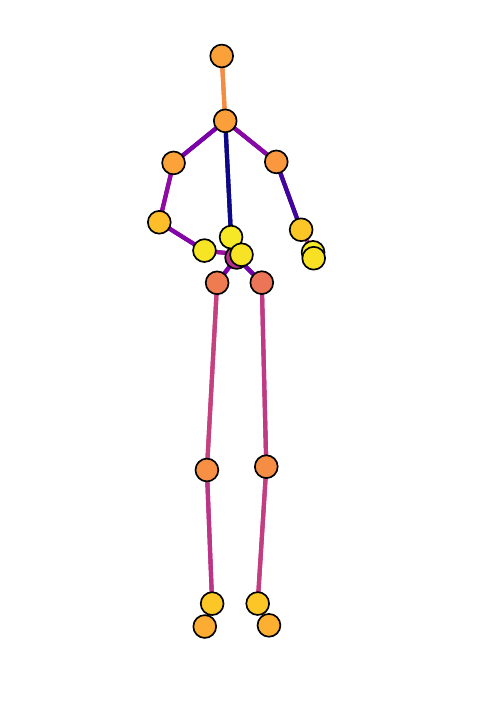}
        \caption{}
    \end{subfigure}
    \begin{subfigure}[b]{0.073\textwidth}
        \centering
        \includegraphics[width=\linewidth, trim=1.8cm 0.7cm 1.8cm 0, clip]{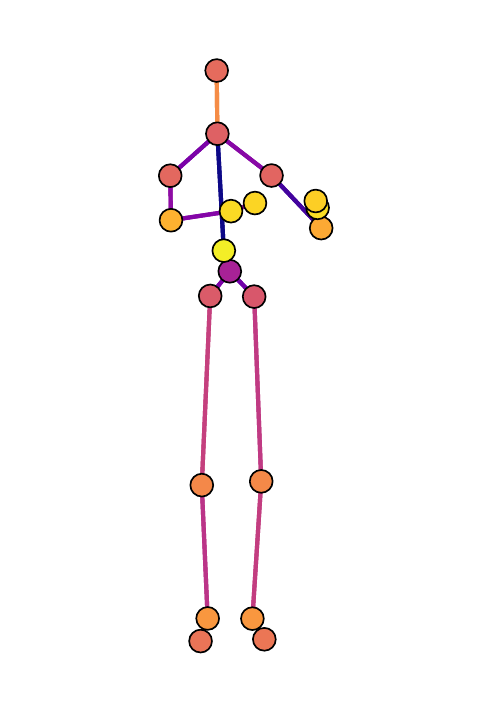}
        \caption{}
    \end{subfigure}
    \begin{subfigure}[b]{0.073\textwidth}
        \centering
        \includegraphics[width=\linewidth, trim=1.8cm 0cm 1.8cm 0, clip]{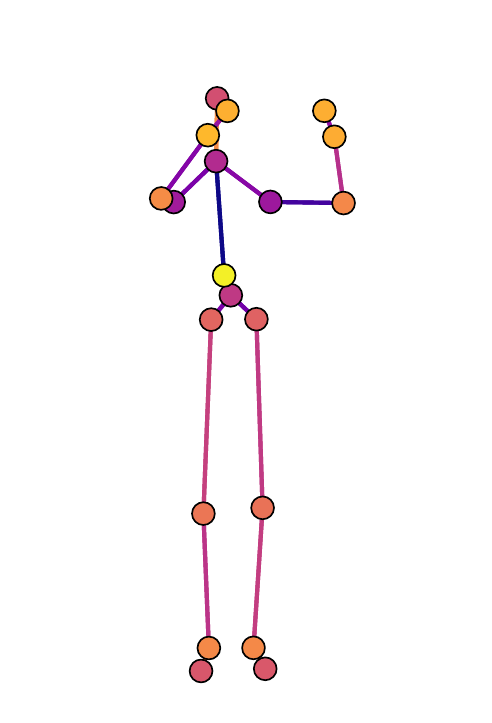}
        \caption{}
    \end{subfigure}
    \begin{subfigure}[b]{0.091\textwidth}
        \centering
        \includegraphics[width=\linewidth, trim=1.5cm 0cm 1cm 0, clip]{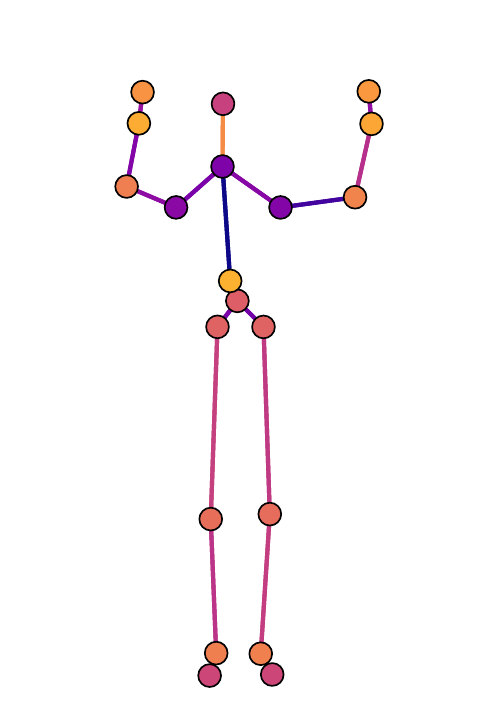}
        \caption{}
    \end{subfigure}
    \begin{subfigure}[b]{0.103\textwidth}
        \centering
        \includegraphics[width=\linewidth, trim=0.8cm 0.1cm 0.7cm 0, clip]{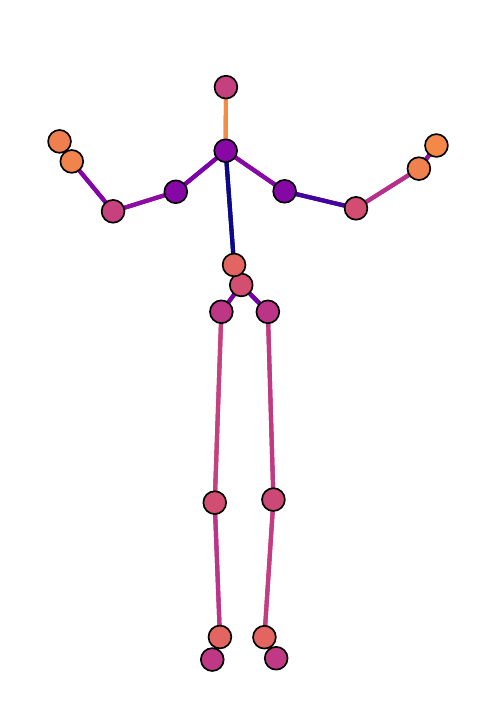}
        \caption{}
    \end{subfigure}
    \begin{subfigure}[b]{0.103\textwidth}
        \centering
        \includegraphics[width=\linewidth, trim=0.8cm 0.8cm 0.8cm 0, clip]{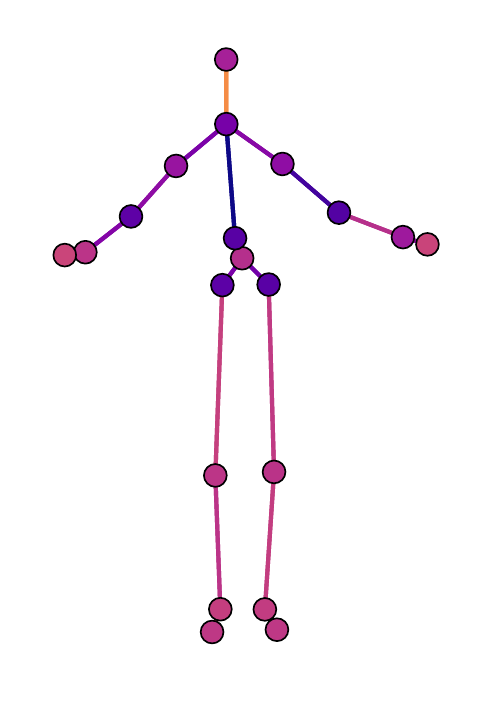}
        \caption{}
    \end{subfigure}
    \begin{subfigure}[b]{0.073\textwidth}
        \centering
        \includegraphics[width=\linewidth, trim=1.8cm 1cm 1.8cm 0, clip]{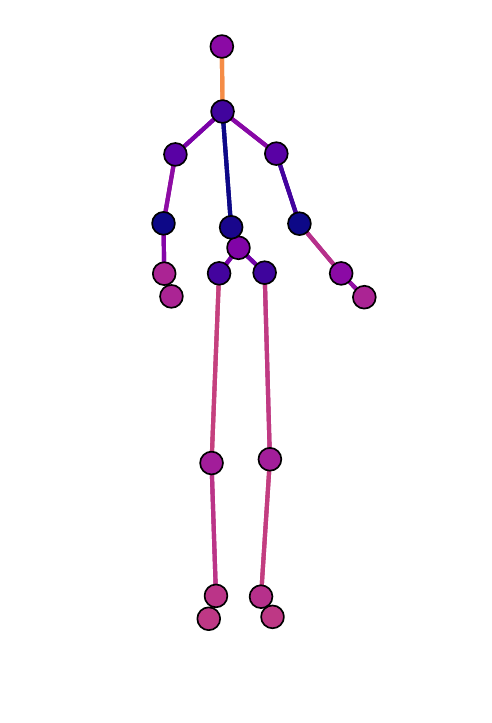}
        \caption{}
    \end{subfigure}
    \begin{subfigure}[b]{0.073\textwidth}
        \centering
        \includegraphics[width=\linewidth, trim=1.8cm 1cm 1.8cm 0, clip]{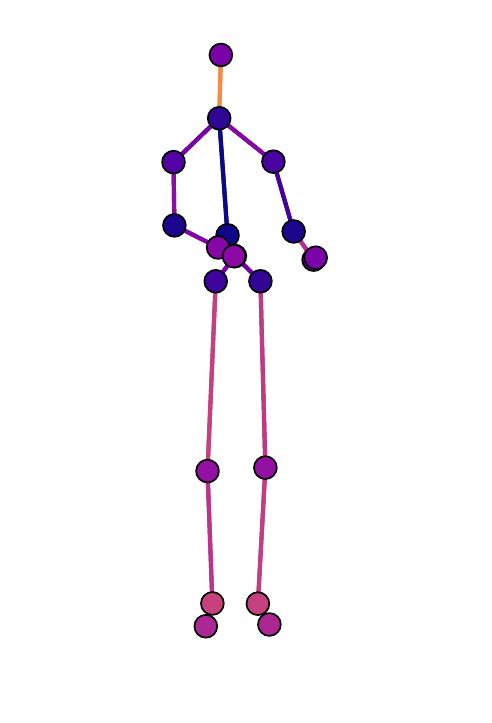}
        \caption{}
    \end{subfigure}
    \begin{subfigure}[b]{0.05\textheight}
        \centering
        \includegraphics[height=2.8cm]{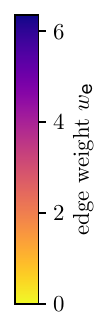}
        \caption*{}
    \end{subfigure}
    \caption{One example from the \gls{msrc12} dataset, corresponding to the class ``wind up the music''.}
    \label{fig:msrc12-wind-up}
\end{figure*}

\begin{figure*}
    \renewcommand*\thesubfigure{\arabic{subfigure}}
    \centering
    \begin{subfigure}[b]{0.05\textwidth}
        \centering
        \includegraphics[height=2.8cm]{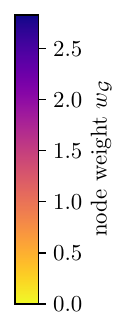}
        \caption*{}
    \end{subfigure}
    \begin{subfigure}[b]{0.078\textwidth}
        \centering
        \includegraphics[width=\linewidth, trim=0.8cm 1.1cm 1cm 0, clip]{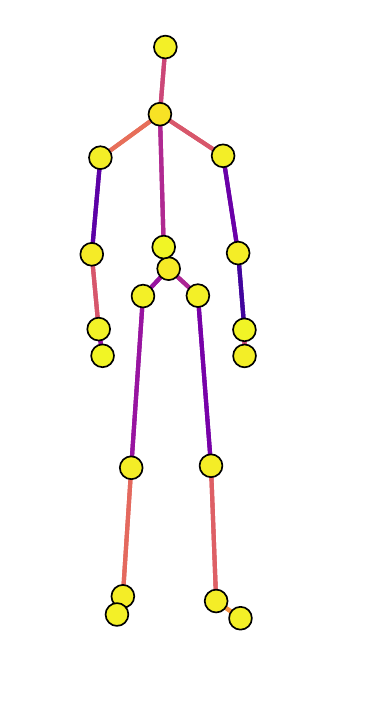}
        \caption{}
    \end{subfigure}
    \begin{subfigure}[b]{0.078\textwidth}
        \centering
        \includegraphics[width=\linewidth, trim=0.8cm 1.1cm 1cm 0, clip]{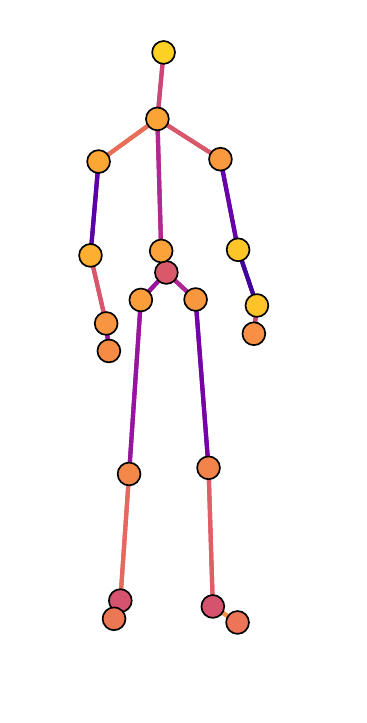}
        \caption{}
    \end{subfigure}
    \begin{subfigure}[b]{0.078\textwidth}
        \centering
        \includegraphics[width=\linewidth, trim=0.8cm 0.7cm 1cm 0, clip]{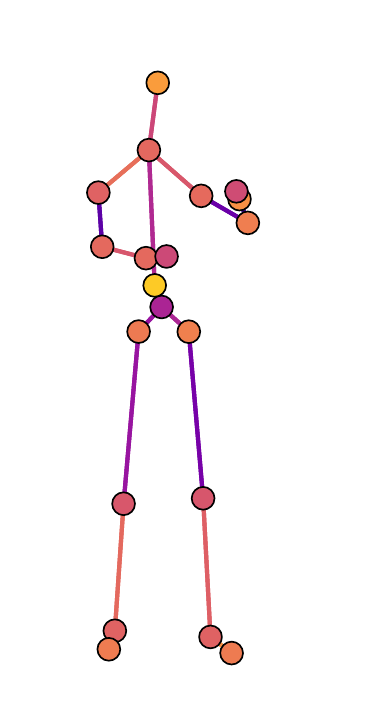}
        \caption{}
    \end{subfigure}
    \begin{subfigure}[b]{0.078\textwidth}
        \centering
        \includegraphics[width=\linewidth, trim=0.8cm 0.2cm 1cm 0, clip]{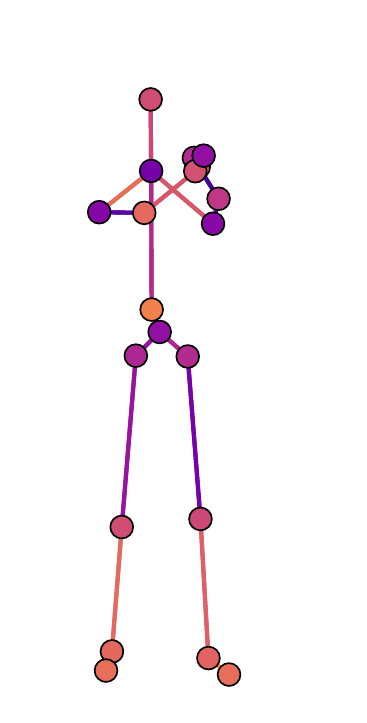}
        \caption{}
    \end{subfigure}
    \begin{subfigure}[b]{0.078\textwidth}
        \centering
        \includegraphics[width=\linewidth, trim=0.8cm 0.2cm 1cm 0, clip]{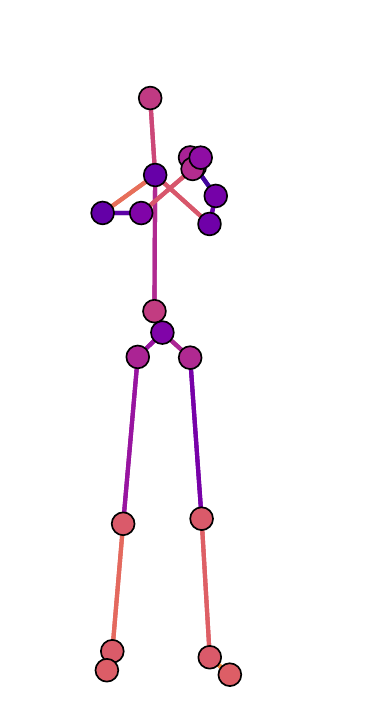}
        \caption{}
    \end{subfigure}
    \begin{subfigure}[b]{0.078\textwidth}
        \centering
        \includegraphics[width=\linewidth, trim=0.8cm 0.2cm 1cm 0, clip]{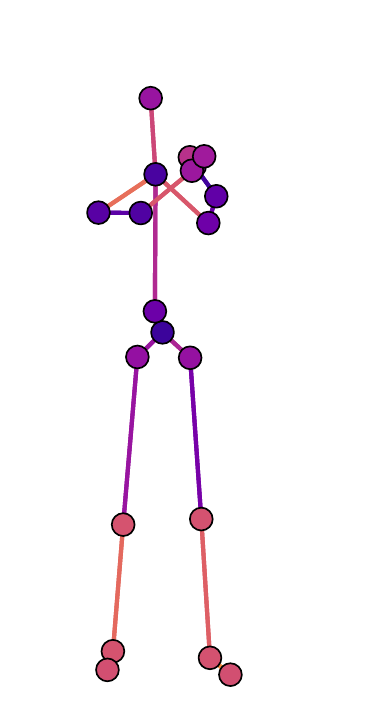}
        \caption{}
    \end{subfigure}
    \begin{subfigure}[b]{0.078\textwidth}
        \centering
        \includegraphics[width=\linewidth, trim=0.8cm 0.2cm 1cm 0, clip]{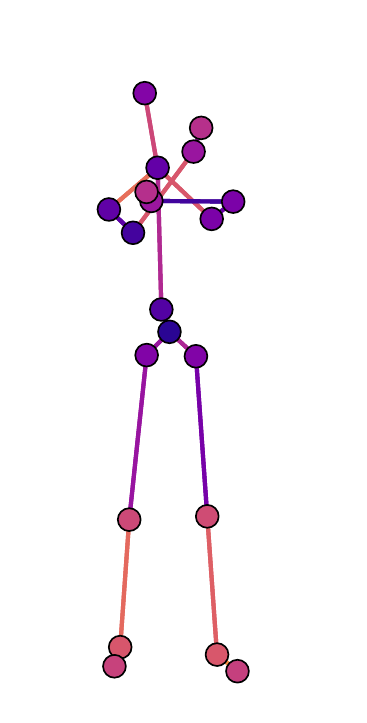}
        \caption{}
    \end{subfigure}
    \begin{subfigure}[b]{0.078\textwidth}
        \centering
        \includegraphics[width=\linewidth, trim=0.8cm 0.3cm 1cm 0, clip]{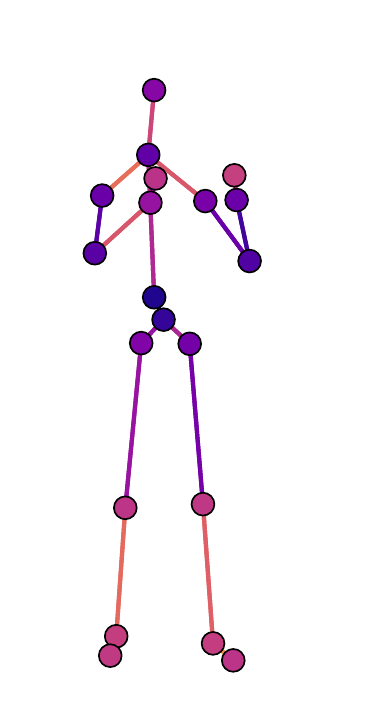}
        \caption{}
    \end{subfigure}
    \begin{subfigure}[b]{0.086\textwidth}
        \centering
        \includegraphics[width=\linewidth, trim=0.55cm 0.8cm 0.75cm 0, clip]{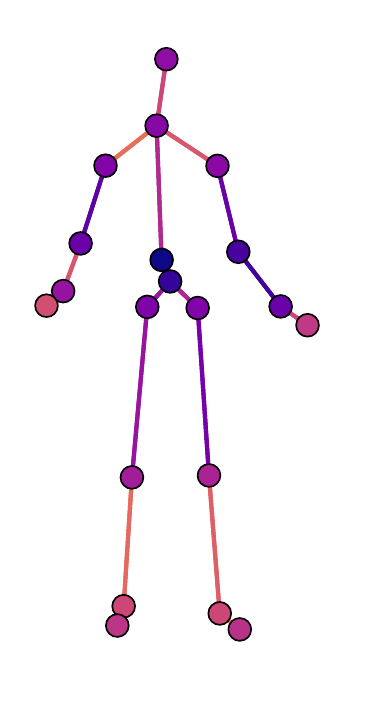}
        \caption{}
    \end{subfigure}
    \begin{subfigure}[b]{0.078\textwidth}
        \centering
        \includegraphics[width=\linewidth, trim=0.8cm 1cm 1cm 0, clip]{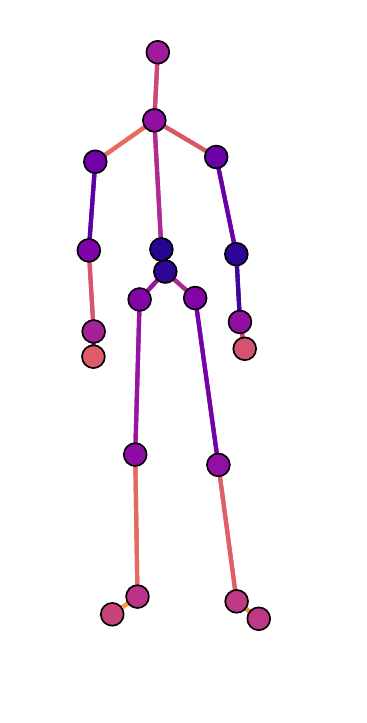}
        \caption{}
    \end{subfigure}
    \begin{subfigure}[b]{0.05\textheight}
        \centering
        \includegraphics[height=2.8cm]{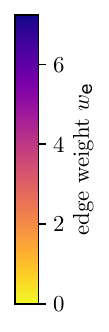}
        \caption*{}
    \end{subfigure}
    \caption{One example from the \gls{msrc12} dataset, corresponding to the class ``shoot a pistol''.}
    \label{fig:msrc12-shoot}

    \vspace{0.5cm}
    \setcounter{subfigure}{0}
    
    \begin{subfigure}[b]{0.05\textwidth}
        \centering
        \includegraphics[height=2.8cm]{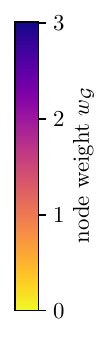}
        \caption*{}
    \end{subfigure}
    \begin{subfigure}[b]{0.076\textwidth}
        \centering
        \includegraphics[width=\linewidth, trim=1cm 1cm 1cm 0, clip]{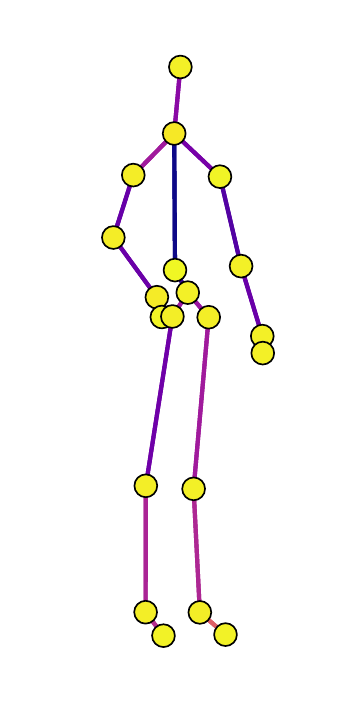}
        \caption{}
    \end{subfigure}
    \begin{subfigure}[b]{0.076\textwidth}
        \centering
        \includegraphics[width=\linewidth, trim=1cm 0.9cm 1cm 0, clip]{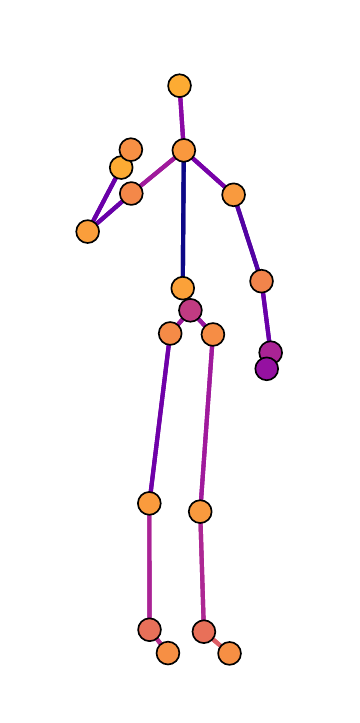}
        \caption{}
    \end{subfigure}
    \begin{subfigure}[b]{0.076\textwidth}
        \centering
        \includegraphics[width=\linewidth, trim=1cm 0.9cm 1cm 0, clip]{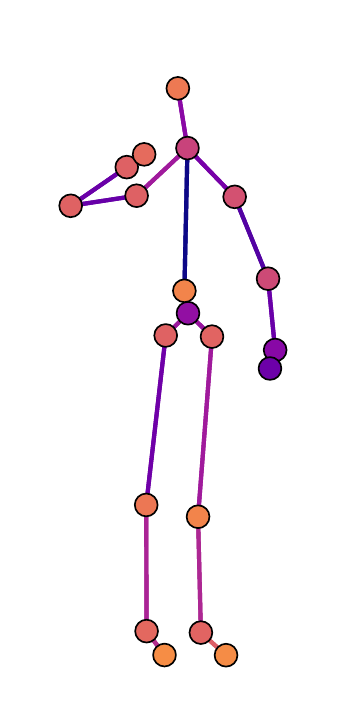}
        \caption{}
    \end{subfigure}
    \begin{subfigure}[b]{0.084\textwidth}
        \centering
        \includegraphics[width=\linewidth, trim=0.55cm 0.8cm 1cm 0, clip]{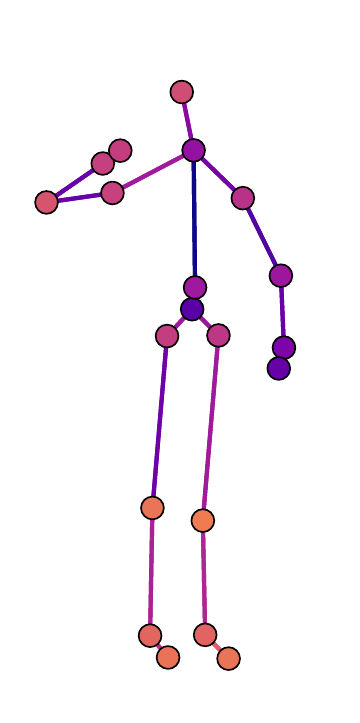}
        \caption{}
    \end{subfigure}
    \begin{subfigure}[b]{0.082\textwidth}
        \centering
        \includegraphics[width=\linewidth, trim=0.7cm 0.75cm 1cm 0, clip]{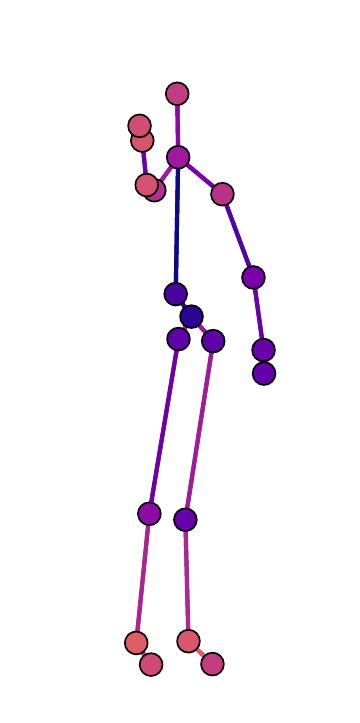}
        \caption{}
    \end{subfigure}
    \begin{subfigure}[b]{0.082\textwidth}
        \centering
        \includegraphics[width=\linewidth, trim=1cm 0.7cm 0.6cm 0, clip]{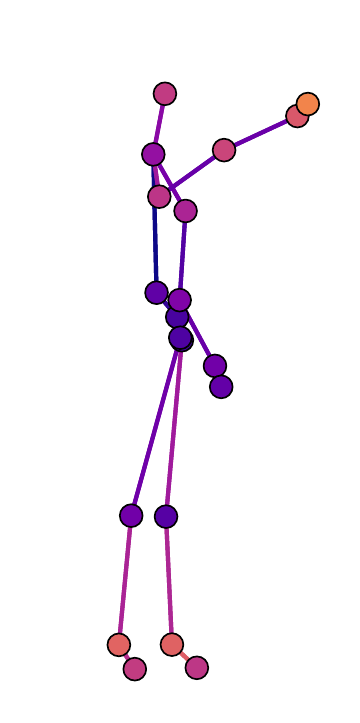}
        \caption{}
    \end{subfigure}
    \begin{subfigure}[b]{0.084\textwidth}
        \centering
        \includegraphics[width=\linewidth, trim=1cm 0.8cm 0.5cm 0, clip]{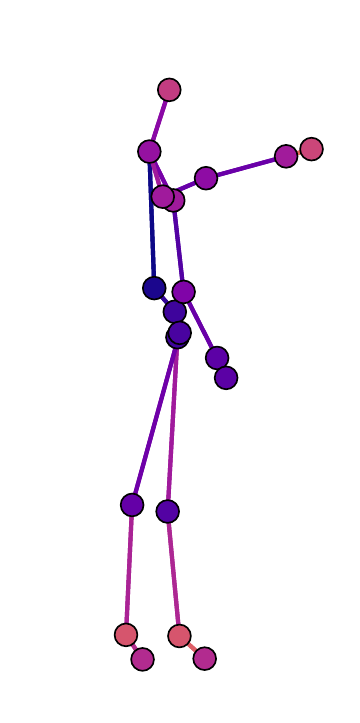}
        \caption{}
    \end{subfigure}
    \begin{subfigure}[b]{0.076\textwidth}
        \centering
        \includegraphics[width=\linewidth, trim=1cm 1.1cm 0.9cm 0, clip]{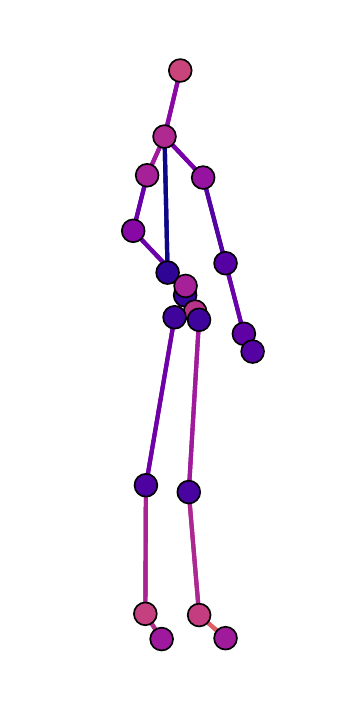}
        \caption{}
    \end{subfigure}
    \begin{subfigure}[b]{0.076\textwidth}
        \centering
        \includegraphics[width=\linewidth, trim=1cm 1.1cm 0.9cm 0, clip]{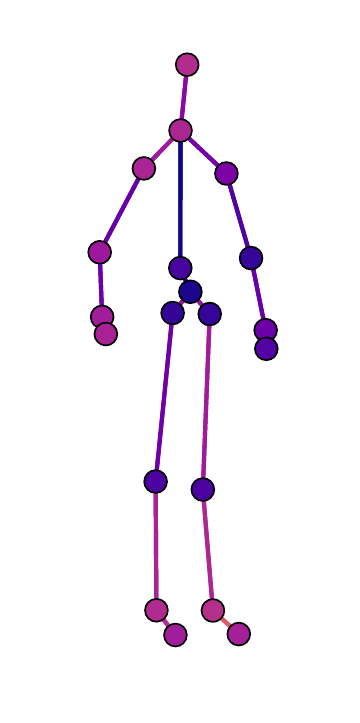}
        \caption{}
    \end{subfigure}
    \begin{subfigure}[b]{0.076\textwidth}
        \centering
        \includegraphics[width=\linewidth, trim=1cm 1.1cm 0.9cm 0, clip]{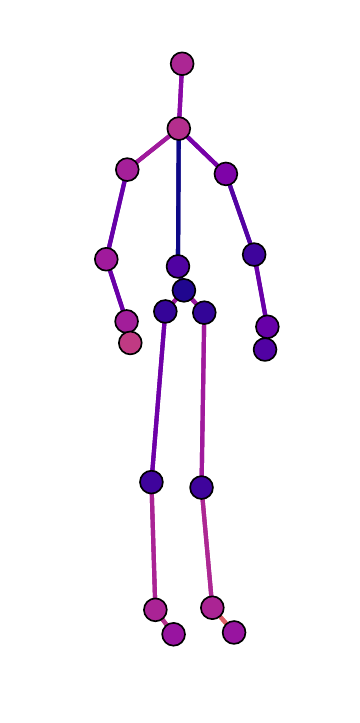}
        \caption{}
    \end{subfigure}
    \begin{subfigure}[b]{0.05\textheight}
        \centering
        \includegraphics[height=2.8cm]{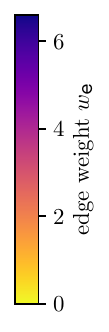}
        \caption*{}
    \end{subfigure}
    \caption{One example from the \gls{msrc12} dataset, corresponding to the class ``throw an object''.}
    \label{fig:msrc12-throw}

    \vspace{0.5cm}
    \setcounter{subfigure}{0}
    
    \begin{subfigure}[b]{0.05\textwidth}
        \centering
        \includegraphics[height=2.8cm]{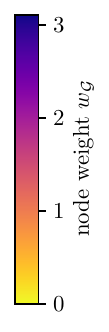}
        \caption*{}
    \end{subfigure}
    \begin{subfigure}[b]{0.078\textwidth}
        \centering
        \includegraphics[width=\linewidth, trim=1.5cm 1.4cm 1.5cm 0, clip]{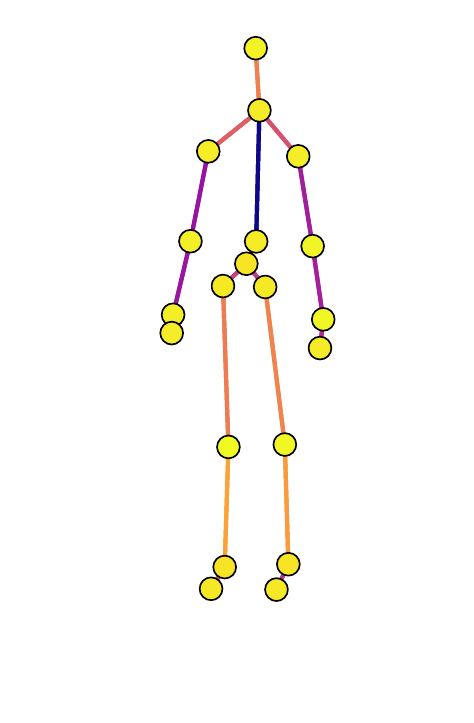}
        \caption{}
    \end{subfigure}
    \begin{subfigure}[b]{0.082\textwidth}
        \centering
        \includegraphics[width=\linewidth, trim=1.2cm 1.2cm 1.5cm 0, clip]{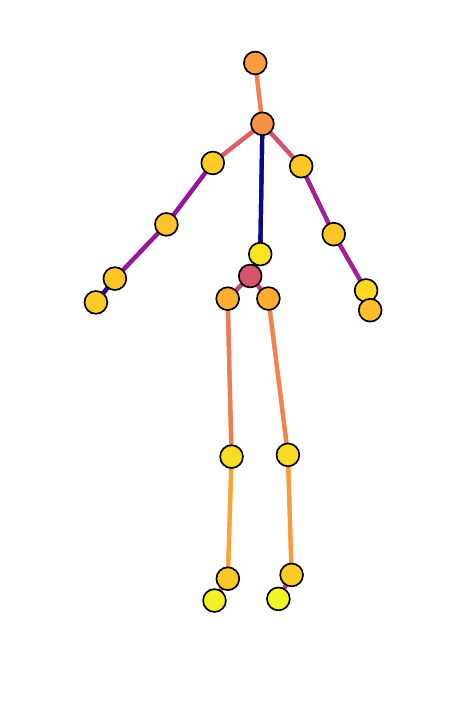}
        \caption{}
    \end{subfigure}
    \begin{subfigure}[b]{0.078\textwidth}
        \centering
        \includegraphics[width=\linewidth, trim=1.5cm 0cm 1.5cm 0, clip]{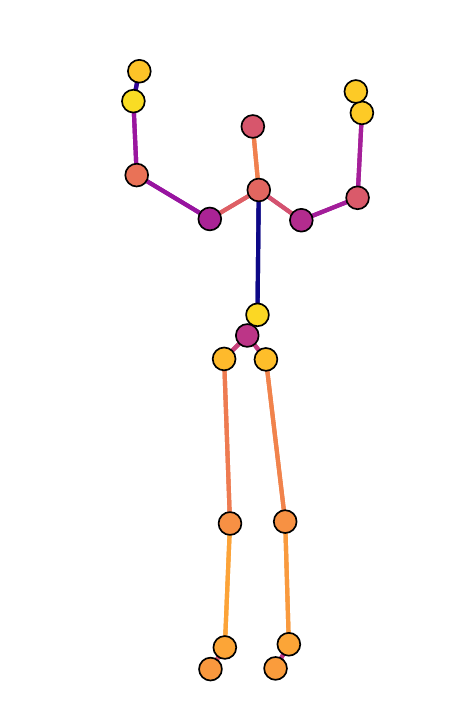}
        \caption{}
    \end{subfigure}
    \begin{subfigure}[b]{0.078\textwidth}
        \centering
        \includegraphics[width=\linewidth, trim=1.5cm 0cm 1.5cm 0, clip]{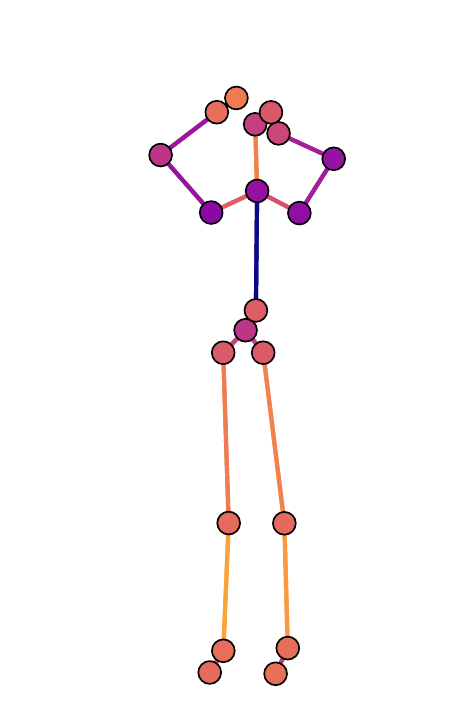}
        \caption{}
    \end{subfigure}
    \begin{subfigure}[b]{0.078\textwidth}
        \centering
        \includegraphics[width=\linewidth, trim=1.5cm 0cm 1.5cm 0, clip]{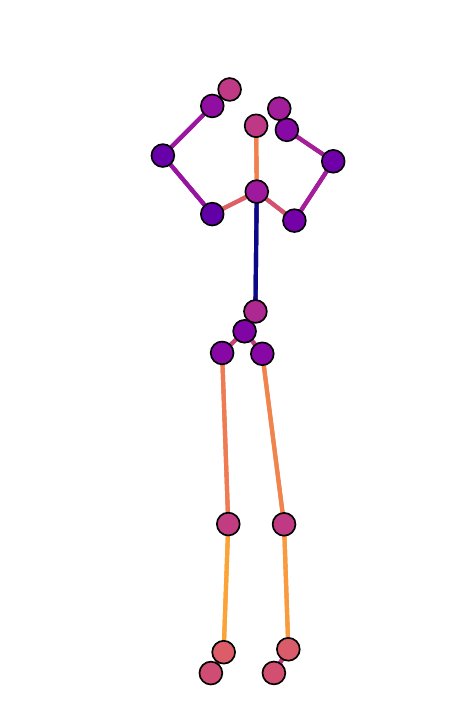}
        \caption{}
    \end{subfigure}
    \begin{subfigure}[b]{0.078\textwidth}
        \centering
        \includegraphics[width=\linewidth, trim=1.5cm 0cm 1.5cm 0, clip]{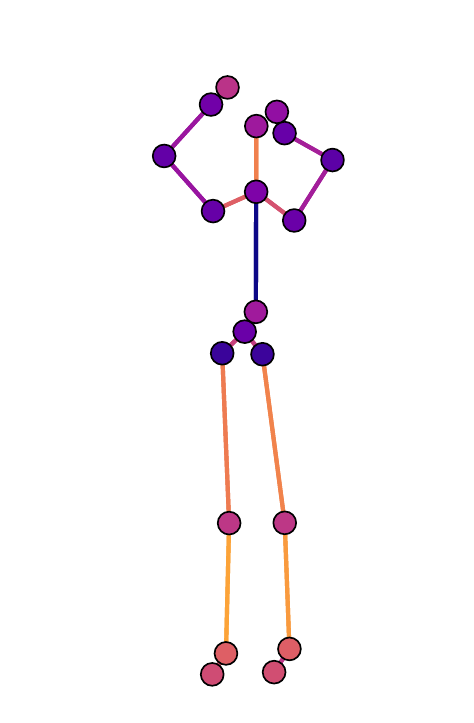}
        \caption{}
    \end{subfigure}
    \begin{subfigure}[b]{0.078\textwidth}
        \centering
        \includegraphics[width=\linewidth, trim=1.5cm 0cm 1.5cm 0, clip]{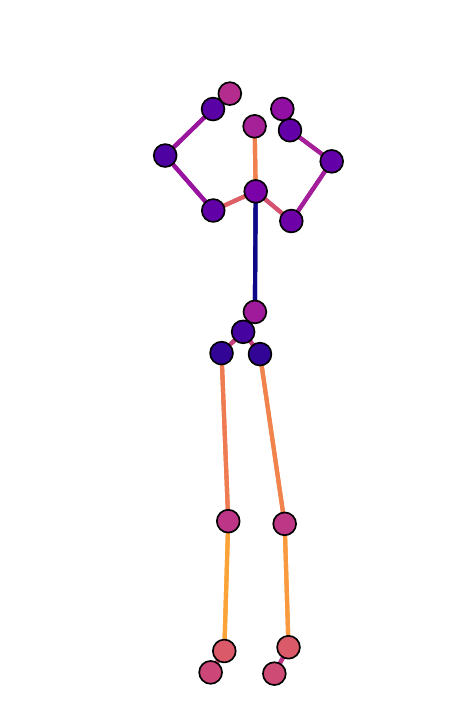}
        \caption{}
    \end{subfigure}
    \begin{subfigure}[b]{0.078\textwidth}
        \centering
        \includegraphics[width=\linewidth, trim=1.5cm 0.5cm 1.5cm 0, clip]{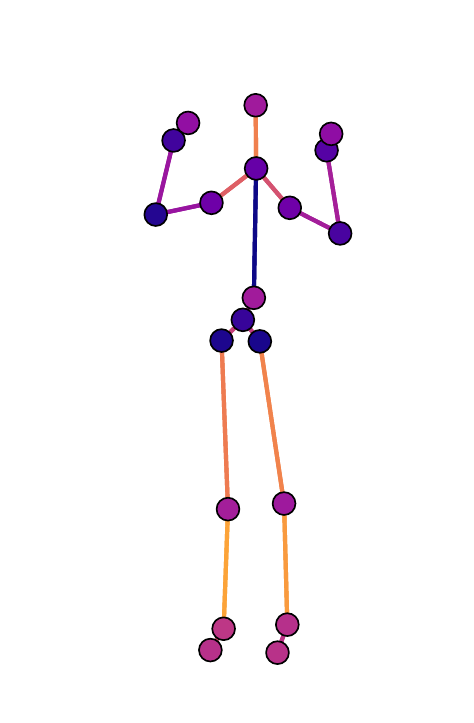}
        \caption{}
    \end{subfigure}
    \begin{subfigure}[b]{0.078\textwidth}
        \centering
        \includegraphics[width=\linewidth, trim=1.5cm 1.5cm 1.5cm 0, clip]{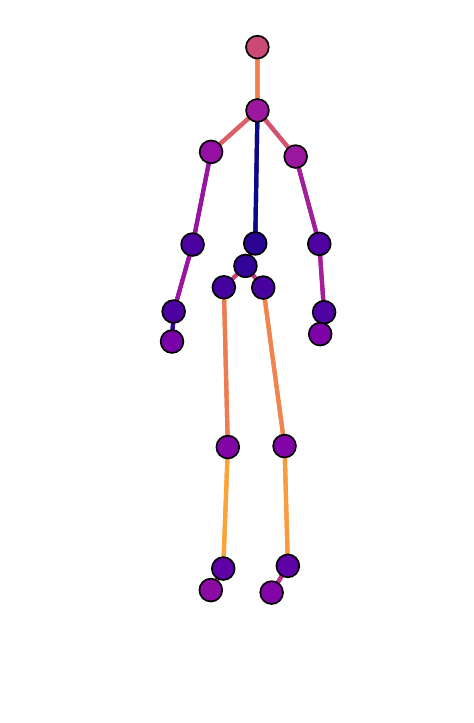}
        \caption{}
    \end{subfigure}
    \begin{subfigure}[b]{0.078\textwidth}
        \centering
        \includegraphics[width=\linewidth, trim=1.5cm 1.5cm 1.5cm 0, clip]{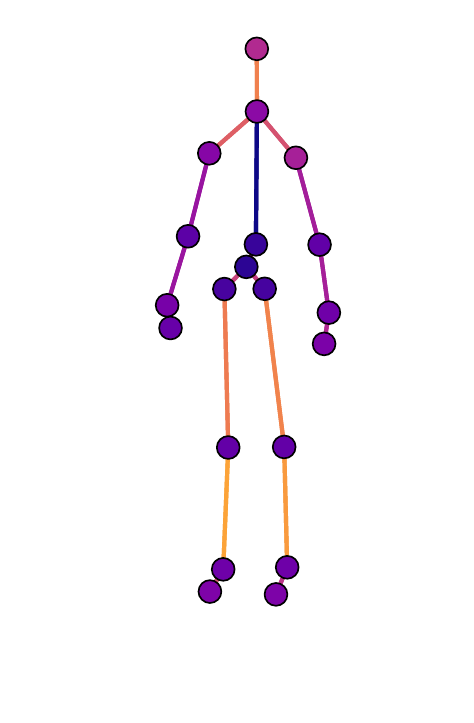}
        \caption{}
    \end{subfigure}
    \begin{subfigure}[b]{0.05\textheight}
        \centering
        \includegraphics[height=2.8cm]{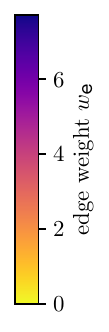}
        \caption*{}
    \end{subfigure}
    \caption{One example from the \gls{msrc12} dataset, corresponding to the class ``protest the music''.}
    \label{fig:msrc12-had-enough}

    \vspace{0.5cm}
    \setcounter{subfigure}{0}
    
    \begin{subfigure}[b]{0.05\textwidth}
        \centering
        \includegraphics[height=2.8cm]{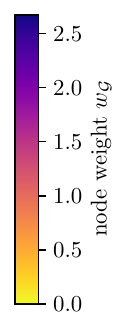}
        \caption*{}
    \end{subfigure}
    \begin{subfigure}[b]{0.064\textwidth}
        \centering
        \includegraphics[width=\linewidth, trim=1.5cm 2cm 1.5cm 0, clip]{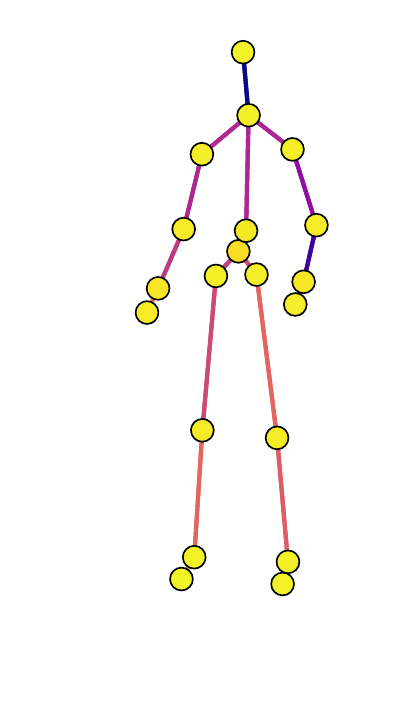}
        \caption{}
    \end{subfigure}
    \begin{subfigure}[b]{0.1\textwidth}
        \centering
        \includegraphics[width=\linewidth, trim=0 1.2cm 0.5cm 0, clip]{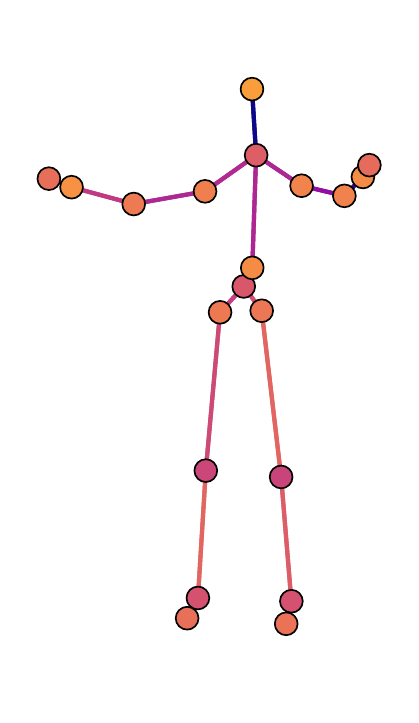}
        \caption{}
    \end{subfigure}
    \begin{subfigure}[b]{0.074\textwidth}
        \centering
        \includegraphics[width=\linewidth, trim=1.5cm 1cm 0.8cm 0, clip]{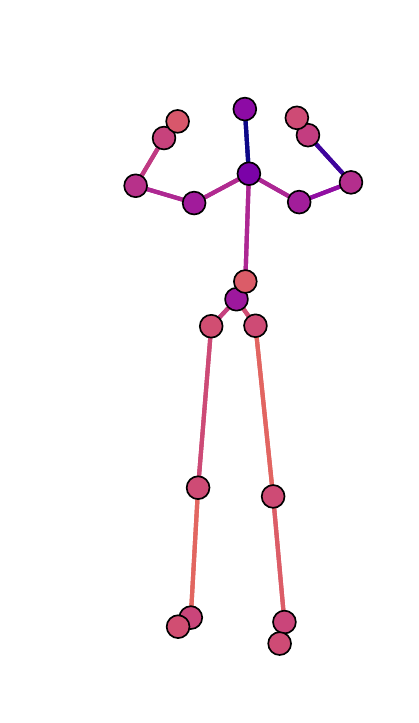}
        \caption{}
    \end{subfigure}
    \begin{subfigure}[b]{0.084\textwidth}
        \centering
        \includegraphics[width=\linewidth, trim=1cm 0.5cm 0.8cm 0, clip]{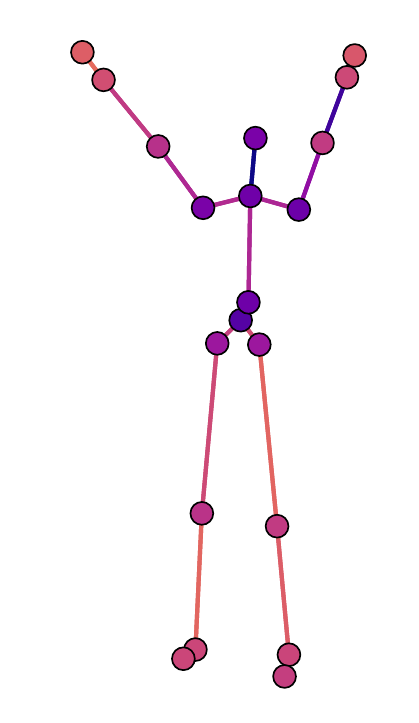}
        \caption{}
    \end{subfigure}
    \begin{subfigure}[b]{0.074\textwidth}
        \centering
        \includegraphics[width=\linewidth, trim=1.5cm 0.8cm 0.8cm 0, clip]{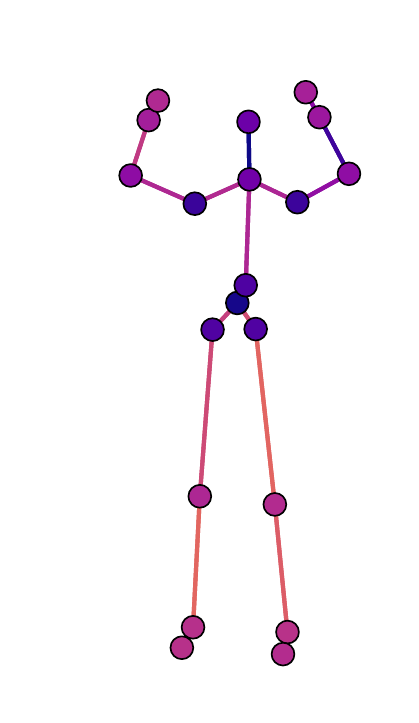}
        \caption{}
    \end{subfigure}
    \begin{subfigure}[b]{0.074\textwidth}
        \centering
        \includegraphics[width=\linewidth, trim=1.5cm 1cm 0.8cm 0, clip]{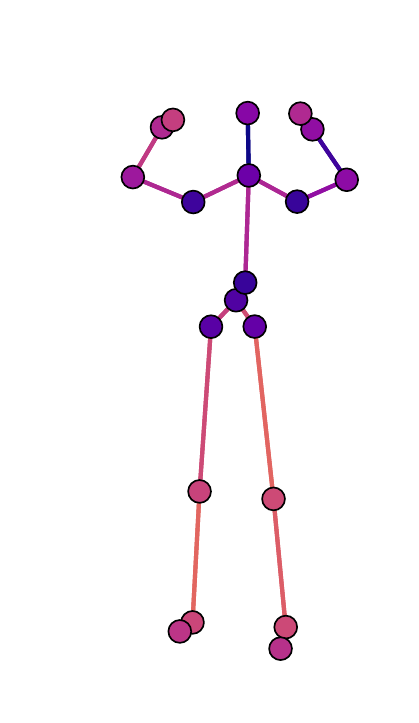}
        \caption{}
    \end{subfigure}
    \begin{subfigure}[b]{0.098\textwidth}
        \centering
        \includegraphics[width=\linewidth, trim=0 0.6cm 0.6cm 0, clip]{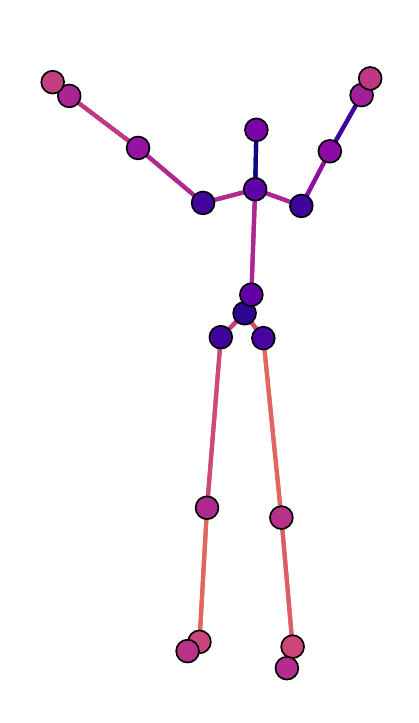}
        \caption{}
    \end{subfigure}
    \begin{subfigure}[b]{0.098\textwidth}
        \centering
        \includegraphics[width=\linewidth, trim=0 0.8cm 0.8cm 0, clip]{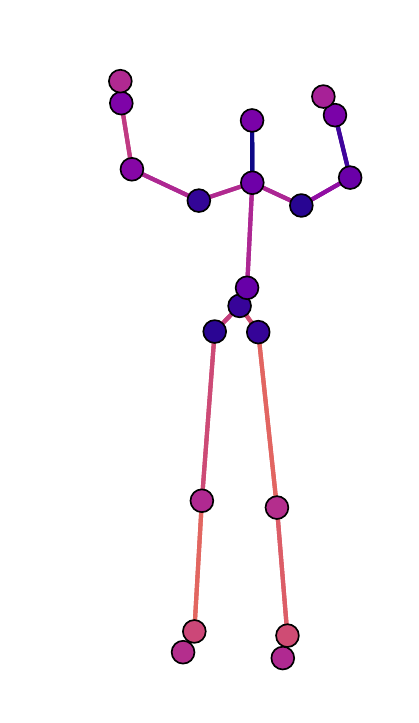}
        \caption{}
    \end{subfigure}
    \begin{subfigure}[b]{0.074\textwidth}
        \centering
        \includegraphics[width=\linewidth, trim=1.5cm 1.5cm 1cm 0, clip]{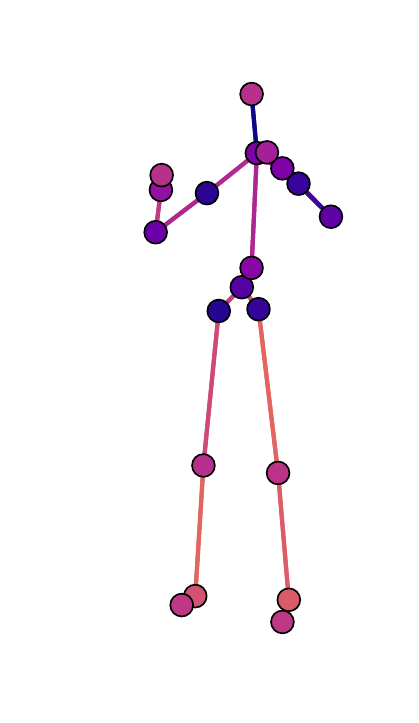}
        \caption{}
    \end{subfigure}
    \begin{subfigure}[b]{0.064\textwidth}
        \centering
        \includegraphics[width=\linewidth, trim=1.5cm 2.2cm 1.5cm 0, clip]{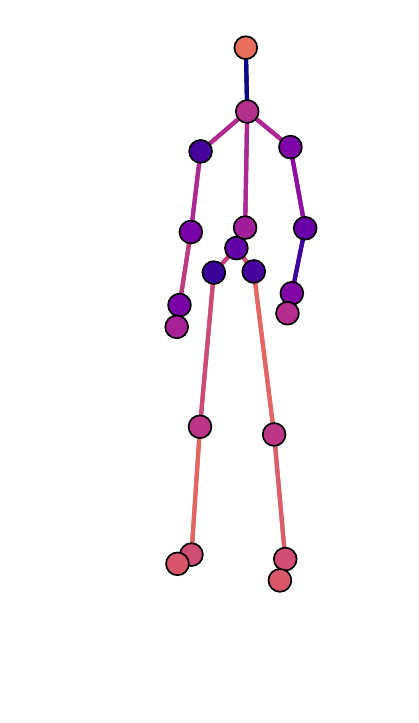}
        \caption{}
    \end{subfigure}
    \begin{subfigure}[b]{0.05\textheight}
        \centering
        \includegraphics[height=2.8cm]{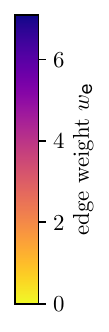}
        \caption*{}
    \end{subfigure}
    \caption{One example from the \gls{msrc12} dataset, corresponding to the class ``move up the tempo of the song''.}
    \label{fig:msrc12-move-up-tempo}
\end{figure*}

\begin{figure*}
    \renewcommand*\thesubfigure{\arabic{subfigure}}
    \centering
    \begin{subfigure}[b]{0.05\textwidth}
        \centering
        \includegraphics[height=2.8cm]{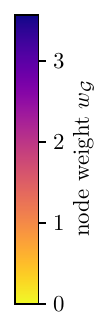}
        \caption*{}
    \end{subfigure}
    \begin{subfigure}[b]{0.072\textwidth}
        \centering
        \includegraphics[width=\linewidth, trim=1cm 1.1cm 1cm 0, clip]{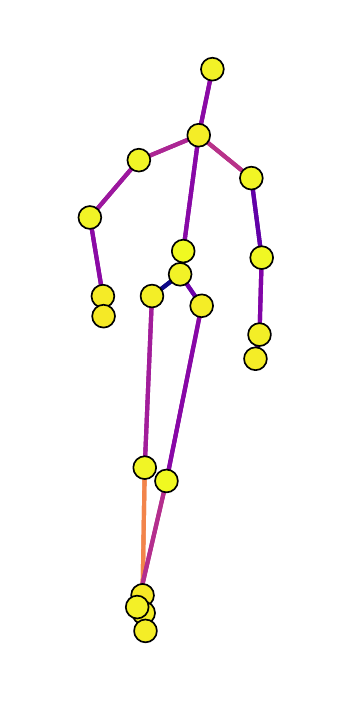}
        \caption{}
    \end{subfigure}
    \begin{subfigure}[b]{0.072\textwidth}
        \centering
        \includegraphics[width=\linewidth, trim=1cm 1.1cm 1cm 0, clip]{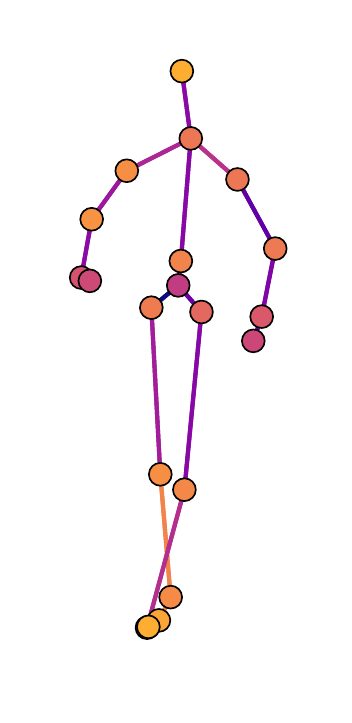}
        \caption{}
    \end{subfigure}
    \begin{subfigure}[b]{0.088\textwidth}
        \centering
        \includegraphics[width=\linewidth, trim=0.5cm 0.6cm 0.5cm 0, clip]{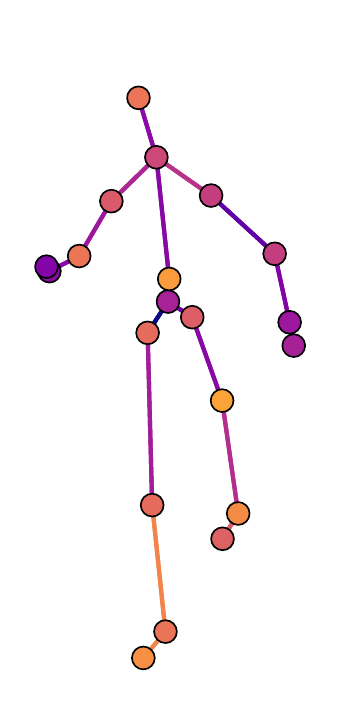}
        \caption{}
    \end{subfigure}
    \begin{subfigure}[b]{0.088\textwidth}
        \centering
        \includegraphics[width=\linewidth, trim=0.5cm 0.6cm 0.5cm 0, clip]{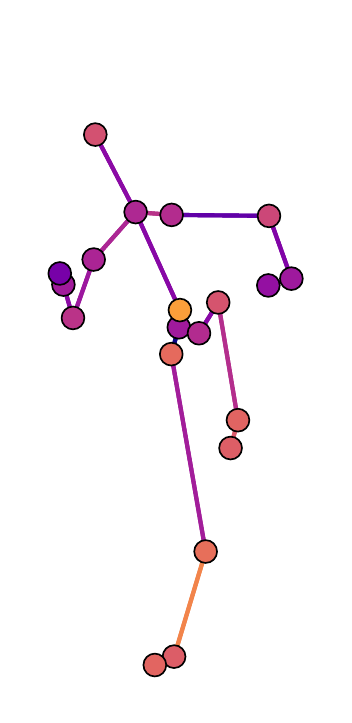}
        \caption{}
    \end{subfigure}
    \begin{subfigure}[b]{0.088\textwidth}
        \centering
        \includegraphics[width=\linewidth, trim=0.5cm 0.6cm 0.5cm 0, clip]{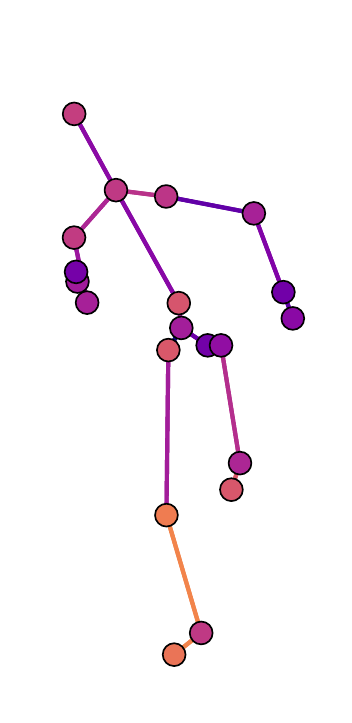}
        \caption{}
    \end{subfigure}
    \begin{subfigure}[b]{0.072\textwidth}
        \centering
        \includegraphics[width=\linewidth, trim=1cm 0.6cm 0.5cm 0, clip]{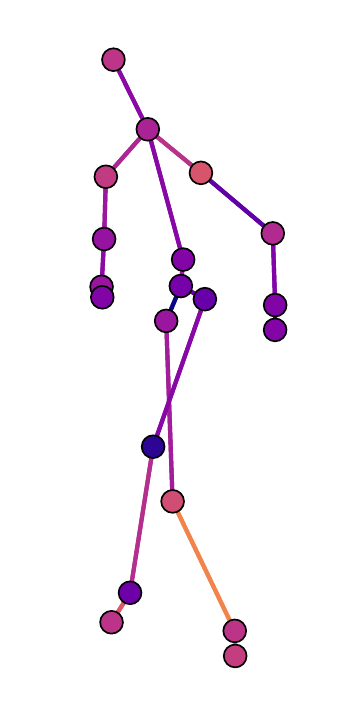}
        \caption{}
    \end{subfigure}
    \begin{subfigure}[b]{0.072\textwidth}
        \centering
        \includegraphics[width=\linewidth, trim=1cm 1cm 0.5cm 0, clip]{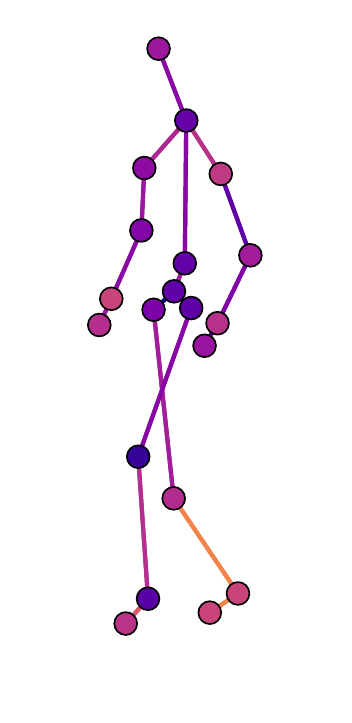}
        \caption{}
    \end{subfigure}
    \begin{subfigure}[b]{0.072\textwidth}
        \centering
        \includegraphics[width=\linewidth, trim=1cm 1cm 0.5cm 0, clip]{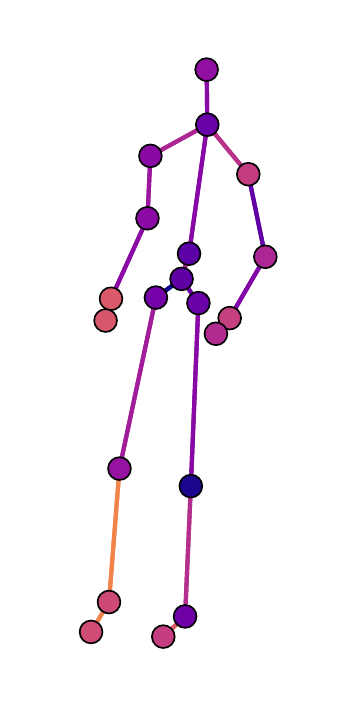}
        \caption{}
    \end{subfigure}
    \begin{subfigure}[b]{0.074\textwidth}
        \centering
        \includegraphics[width=\linewidth, trim=1cm 1cm 0.5cm 0, clip]{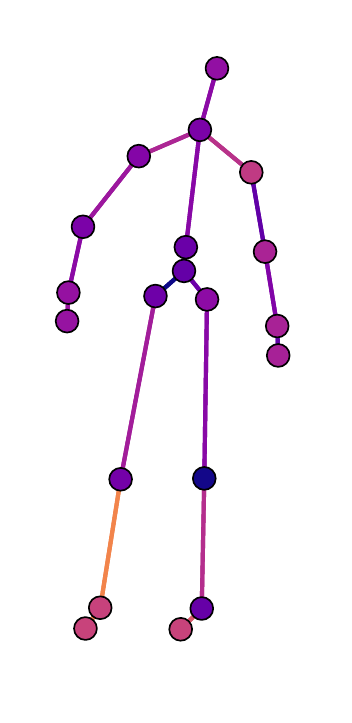}
        \caption{}
    \end{subfigure}
    \begin{subfigure}[b]{0.074\textwidth}
        \centering
        \includegraphics[width=\linewidth, trim=1cm 1cm 0.5cm 0, clip]{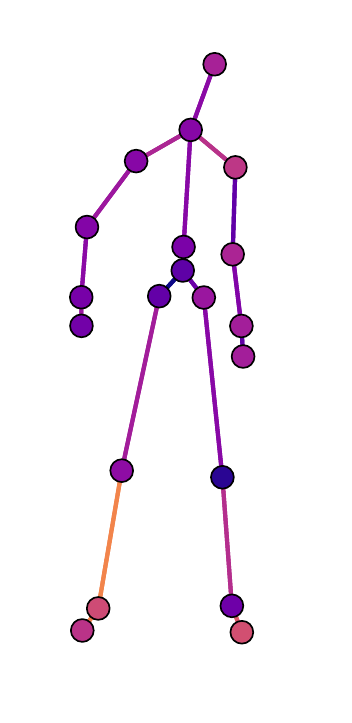}
        \caption{}
    \end{subfigure}
    \begin{subfigure}[b]{0.05\textheight}
        \centering
        \includegraphics[height=2.8cm]{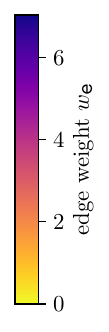}
        \caption*{}
    \end{subfigure}
    \caption{One example from the \gls{msrc12} dataset, corresponding to the class ``kick''.}
    \label{fig:msrc12-kick}
\end{figure*}

We report in this section, in \Cref{fig:msrc12-raise-volume,fig:msrc12-push-right,fig:msrc12-goggles,fig:msrc12-wind-up,fig:msrc12-shoot,fig:msrc12-throw,fig:msrc12-had-enough,fig:msrc12-move-up-tempo,fig:msrc12-kick}, one example of an explanation on \gls{msrc12} for each class not discussed in \Cref{par:msrc12-expl}. Here are some comments:
\begin{itemize}
    \item Most of the classes in the dataset involve movements of the upper body, especially the arms (e.g.~``raise volume of music'', ``put on goggles'', ``wind up the music'', etc.). For this reason, the explanations are similar, but some details serve as telltale signs of how the model can differentiate between them. For example, in \Cref{fig:msrc12-raise-volume}, more weight is given to the hands than in \Cref{fig:msrc12-goggles,fig:msrc12-wind-up}.
    \item It is interesting to compare the classes ``navigate to next menu'' in \Cref{fig:msrc12-push-right} and ``throw an object'' in \Cref{fig:msrc12-throw}. Both primarily involve the movement of one arm, but in both cases, surprisingly, the model focuses mainly on the arm that remains still: in the first case, the hand nodes; in the latter, the whole arm and the pelvis.
    \item The ``kick'' class in \Cref{fig:msrc12-kick} has a more distinct motion, making the explanation also more understandable: it highlights the spine and the leg involved in the kicking motion.
\end{itemize}

\section{Regularised \glsfmtshort{stgnn} with linear dynamics}
\label{sec-app:ablation-linear}

In general, the sequence of states $\bm{h}_{\timeidx,n}$ does not evolve according to the linear dynamics of~\eqref{eq:koopman}.
To investigate this, we introduce an internal state
\begin{equation}
    \label{eq:trainable-K}
    \tilde{\bm{h}}_{\timeidx+\tau,n} := \bm{K}^{\tau} \bm{h}_{\timeidx,n},
\end{equation}
where $\bm{K}\in\mathbb{R}^{F\times F}$ is a trainable parameter that acts as Koopman operator on $\tilde{\bm{h}}_{\timeidx,n}$.
Moreover, to train $\bm{K}$, we add the following terms to the loss:
\begin{itemize}
    \item a binary cross-entropy term $\ell_{\text{rec}}$ between the Koopman-reconstructed output $\tilde{y}=\text{MLP}(\tilde{\bm{h}})$ and the class label $\hat{y}$;
    \item an observable loss $\ell_\text{obs}$, defined as a mean-squared reconstruction loss between the~\gls{tg} embedding at time $\timeidx$, $\bm{h}_{\timeidx}$, and the corresponding Koopman-reconstructed embedding $\tilde{\bm{h}}_{\timeidx}$, together with an $\ell_2$ penalty on $\bm{K}$:
    \begin{equation}
    \label{eq:obs-loss}
        \ell_\text{obs}(\bm{h}_{\timeidx}, \tilde{\bm{h}}_{\timeidx}) = \text{MSE}(\bm{h}_{\timeidx}, \tilde{\bm{h}}_{\timeidx}) + \ell_2(\bm{K}),
    \end{equation}
    where $\ell_2$ is a weight decay regularisation term.
\end{itemize}

These two terms represent regularisation losses that push the model to represent an observable $\varphi$ that satisfies the Koopman operator definition~\eqref{eq:koopman}, as proposed by~\cite{li2017extended} and~\cite{lusch2018deep} in a deep learning setting.

We note that the sole purpose of $\bm{K}$ and the internal state $\tilde{\bm{h}}_{\timeidx,n}$ is to encourage the state $\bm{h}_{\timeidx,n}$ to follow $\tilde{\bm{h}}_{\timeidx,n}$, whose dynamics is, by construction, linear. They are not used to produce the output $y$, nor are they involved in explaining the model.

The complete loss then becomes
\begin{equation}
    \label{eq:total_loss}
    \ell = \ell_{\text{ce}} + \alpha\ell_{\text{rec}} + \beta \ell_\text{obs},
\end{equation}
where $\alpha$ and $\beta$ are hyperparameters.

\begin{figure*}[!ht]
    \centering
    \includegraphics[width=.9\textwidth]{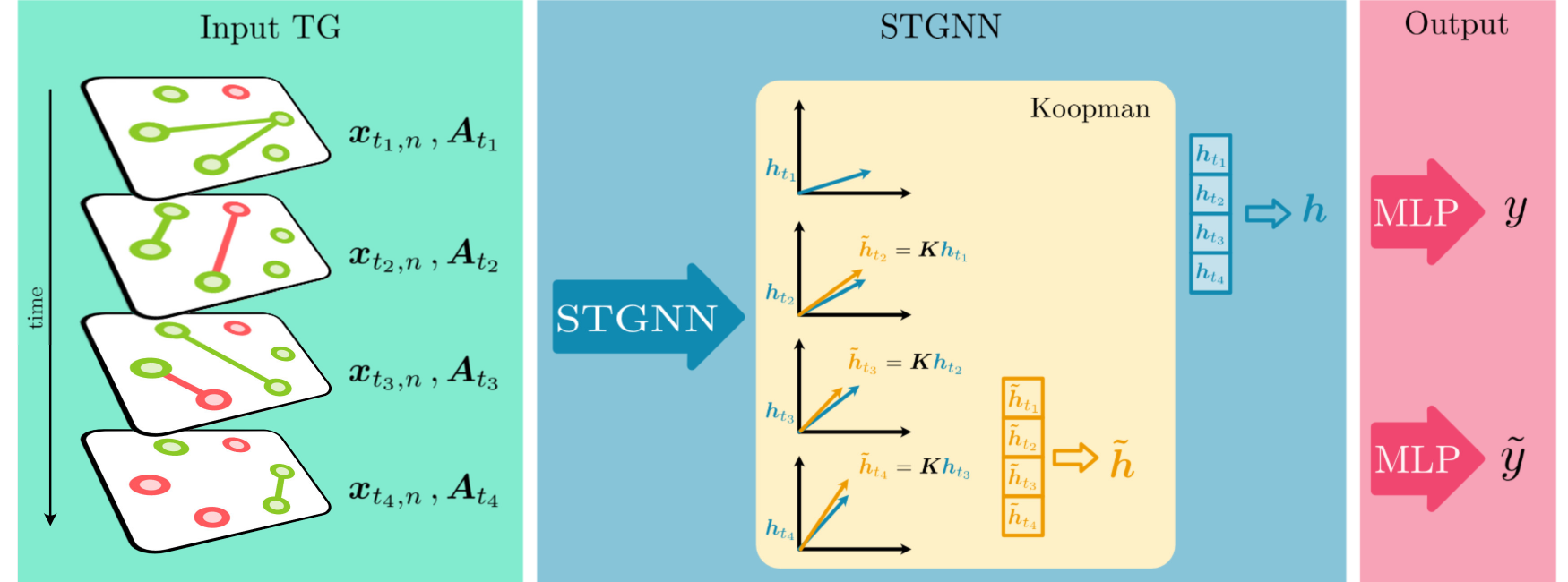}
    \caption{The left-hand side, in green, depicts an example of a \gls{tg}, with node features $\bm{x}_\timeidx$ and adjacency matrix $\bm{A}_\timeidx$. In blue, the~\gls{stgnn} processes the input and provides an embedding $\bm{h}_\timeidx$ for each time step. The inner yellow box represents the mechanism that encourages the embeddings dynamics $\bm{h}_\timeidx$ to be linear: the loss $\ell_\text{obs}$ in~\eqref{eq:obs-loss} pushes $\bm{h}_{\timeidx+1}$ to be a linear transformation of $\bm{h}_\timeidx$ (for illustration, the figure shows a 2-dimensional rotation). In red, an~\gls{mlp} produces the final output.}
    \label{fig:model-linear}
\end{figure*}

The proposed model is depicted in Figure~\ref{fig:model-linear}.

We aim to test whether the proposed regularisation in~\eqref{eq:total_loss}, which pushes the model dynamics
\begin{equation}
    \bm{h}_{\timeidx+1}=\text{STGNN}(\bm{h}_\timeidx, \bm{x}_\timeidx, \bm{A}_\timeidx)
\end{equation}
to exhibit an approximately linear behaviour, improves the performance of the proposed explainability methods.
To do so, we perform an ablation study on the parameters $\alpha$ and $\beta$.

\begin{figure*}
    \renewcommand*\thesubfigure{\arabic{subfigure}}
    \centering
    \begin{subfigure}[b]{0.24\textwidth}
        \centering
        \includegraphics[width=\linewidth, trim=0cm 0cm 0cm 0, clip]{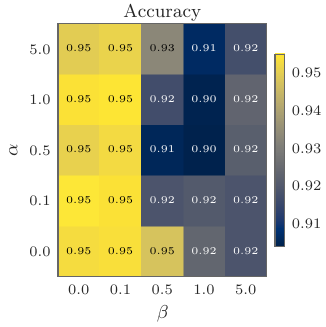}
        \caption{Facebook}
    \end{subfigure}
    \begin{subfigure}[b]{0.24\textwidth}
        \centering
        \includegraphics[width=\linewidth, trim=0cm 0cm 0cm 0, clip]{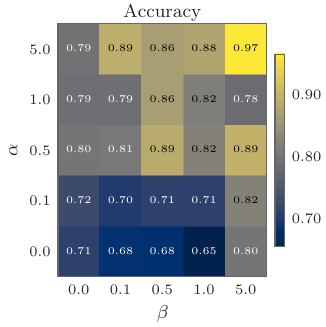}
        \caption{Infectious}
    \end{subfigure}
    \begin{subfigure}[b]{0.24\textwidth}
        \centering
        \includegraphics[width=\linewidth, trim=0cm 0cm 0cm 0, clip]{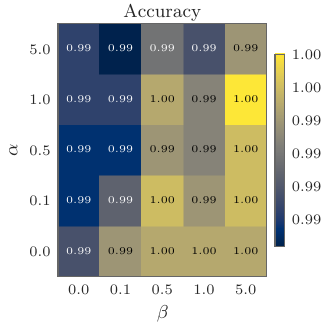}
        \caption{DBLP}
    \end{subfigure}
    \begin{subfigure}[b]{0.24\textwidth}
        \centering
        \includegraphics[width=\linewidth, trim=0cm 0cm 0cm 0, clip]{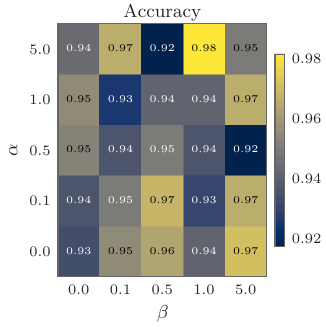}
        \caption{Tumblr}
    \end{subfigure}
    \caption{Effect of $\alpha$ and $\beta$ on accuracy.}
    \label{fig:abl-acc}

    \vspace{0.3cm}
    \setcounter{subfigure}{0}
    
    \begin{subfigure}[b]{0.24\textwidth}
        \centering
        \includegraphics[width=\linewidth, trim=0cm 0cm 0cm 0, clip]{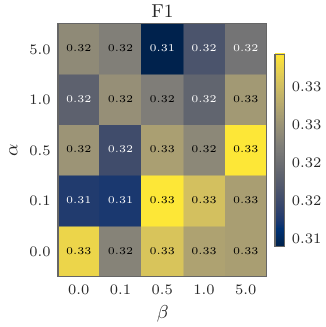}
        \caption{Facebook}
    \end{subfigure}
    \begin{subfigure}[b]{0.24\textwidth}
        \centering
        \includegraphics[width=\linewidth, trim=0cm 0cm 0cm 0, clip]{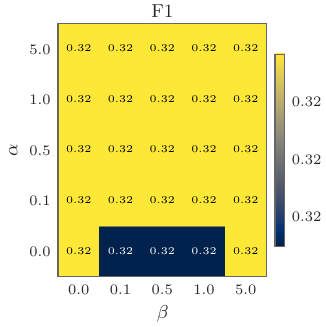}
        \caption{Infectious}
    \end{subfigure}
    \begin{subfigure}[b]{0.24\textwidth}
        \centering
        \includegraphics[width=\linewidth, trim=0cm 0cm 0cm 0, clip]{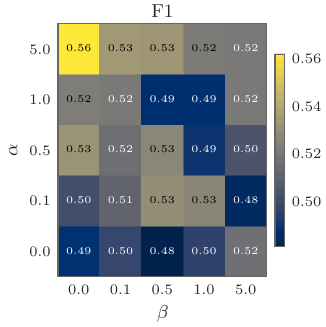}
        \caption{DBLP}
    \end{subfigure}
    \begin{subfigure}[b]{0.24\textwidth}
        \centering
        \includegraphics[width=\linewidth, trim=0cm 0cm 0cm 0, clip]{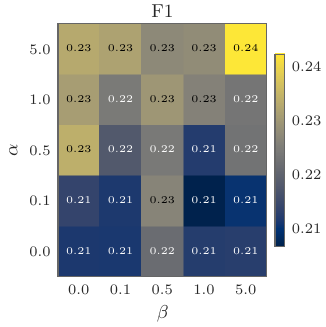}
        \caption{Tumblr}
    \end{subfigure}
    \caption{Effect of $\alpha$ and $\beta$ on F1.}
    \label{fig:abl-f1}

    \vspace{0.3cm}
    \setcounter{subfigure}{0}
    
    \begin{subfigure}[b]{0.24\textwidth}
        \centering
        \includegraphics[width=\linewidth, trim=0cm 0cm 0cm 0, clip]{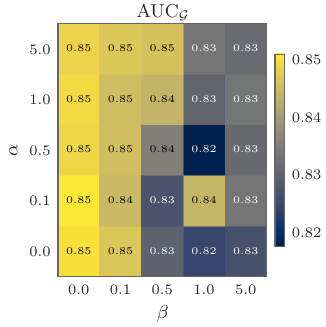}
        \caption{Facebook}
    \end{subfigure}
    \begin{subfigure}[b]{0.24\textwidth}
        \centering
        \includegraphics[width=\linewidth, trim=0cm 0cm 0cm 0, clip]{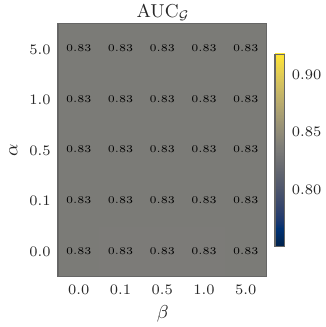}
        \caption{Infectious}
    \end{subfigure}
    \begin{subfigure}[b]{0.24\textwidth}
        \centering
        \includegraphics[width=\linewidth, trim=0cm 0cm 0cm 0, clip]{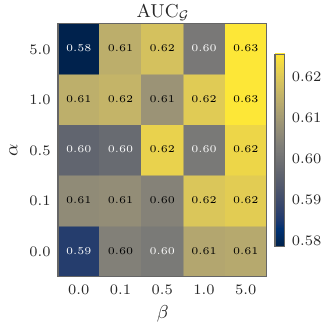}
        \caption{DBLP}
    \end{subfigure}
    \begin{subfigure}[b]{0.24\textwidth}
        \centering
        \includegraphics[width=\linewidth, trim=0cm 0cm 0cm 0, clip]{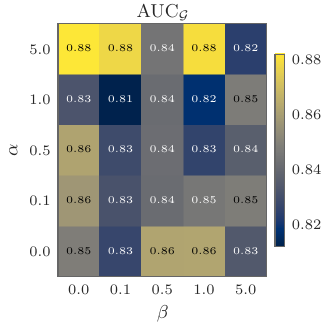}
        \caption{Tumblr}
    \end{subfigure}
    \caption{Effect of $\alpha$ and $\beta$ on $\text{AUC}_\mathcal{G}$.}
    \label{fig:abl-auc-G}

    \vspace{0.3cm}
    \setcounter{subfigure}{0}
    
    \begin{subfigure}[b]{0.24\textwidth}
        \centering
        \includegraphics[width=\linewidth, trim=0cm 0cm 0cm 0, clip]{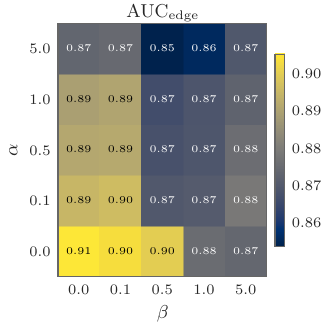}
        \caption{Facebook}
    \end{subfigure}
    \begin{subfigure}[b]{0.24\textwidth}
        \centering
        \includegraphics[width=\linewidth, trim=0cm 0cm 0cm 0, clip]{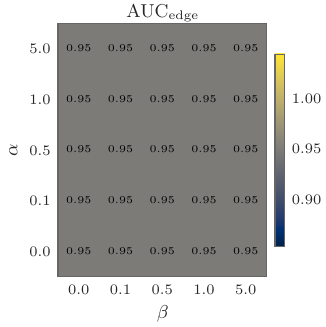}
        \caption{Infectious}
    \end{subfigure}
    \begin{subfigure}[b]{0.24\textwidth}
        \centering
        \includegraphics[width=\linewidth, trim=0cm 0cm 0cm 0, clip]{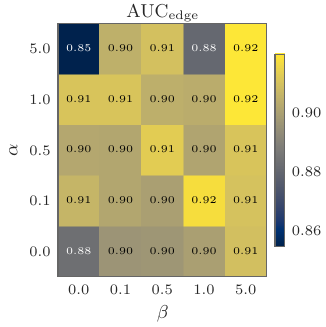}
        \caption{DBLP}
    \end{subfigure}
    \begin{subfigure}[b]{0.24\textwidth}
        \centering
        \includegraphics[width=\linewidth, trim=0cm 0cm 0cm 0, clip]{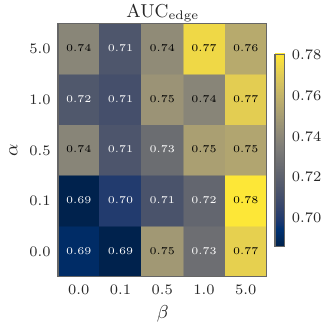}
        \caption{Tumblr}
    \end{subfigure}
    \caption{Effect of $\alpha$ and $\beta$ on $\text{AUC}_\text{edge}$.}
    \label{fig:abl-sindy}
\end{figure*}

In \Cref{fig:abl-acc,fig:abl-auc-G,fig:abl-f1,fig:abl-sindy}, we present results obtained by training a \gls{gcrn} model on four datasets\footnote{The Highschool dataset is omitted from this ablation solely because training a model on it requires substantially more time.}, varying the values of $\alpha$ and $\beta$ among $0$, $0.1$, $0.5$, $1$, and $5$, over five different seeds. Overall, the regularisation terms have little and inconsistent effects:
\begin{description}
    \item[Accuracy] In \Cref{fig:abl-acc}, there are some noticeable positive effects on the Infectious and Facebook datasets, while in the other cases, the variations are small or don't follow a clear pattern.
    \item[F1] In \Cref{fig:abl-f1}, unlike accuracy, a pattern is visible for DBLP and Tumblr, albeit still with small variations.
    \item[$\text{AUC}_\mathcal{G}$] In \Cref{fig:abl-auc-G}, there are some small improvements in DBLP and Tumblr, while the effect of the regularisation is null or detrimental in Facebook and Infectious. 
    \item[$\text{AUC}_\text{edge}$] In \Cref{fig:abl-sindy}, as for F1 and $\text{AUC}_\mathcal{G}$, there are some small improvements in DBLP and Tumblr, but zero to negative effects for Facebook and Infectious.
\end{description}

These results provide two interesting insights. 
On the one hand, they indicate that the explainability gains reported in the main text do not depend on explicitly enforcing linear latent dynamics during training: the proposed post hoc analysis can already extract meaningful structure from standard \glspl{stgnn}. 
On the other hand, they suggest that the relationship between more structured latent dynamics and better explanations is more subtle than can be captured by a simple auxiliary loss. In this sense, this ablation helps delimit the contribution of the paper and motivates future work on more targeted ways of introducing dynamical priors into temporal graph models. 
For this reason, we keep the simpler formulation in the main text and report this variant here as an exploratory extension.

%% file: sample.bib
@article{koopman1931hamiltonian,
  title={Hamiltonian systems and transformation in Hilbert space},
  author={Koopman, Bernard O.},
  journal={Proceedings of the National Academy of Sciences},
  volume={17},
  number={5},
  pages={315--318},
  year={1931},
  publisher={National Acad Sciences}
}

@article{brunton2021modern,
    author = {Brunton, Steven L. and Budi\v{s}i\'{c}, Marko and Kaiser, Eurika and Kutz, J. Nathan},
    title = {Modern Koopman Theory for Dynamical Systems},
    journal = {SIAM Review},
    volume = {64},
    number = {2},
    pages = {229-340},
    year = {2022}
}

@article{longa2023graph,
    title={Graph Neural Networks for Temporal Graphs: State of the Art, Open Challenges, and Opportunities},
    author={Longa, Antonio and Lachi, Veronica and Santin, Gabriele and Bianchini, Monica and Lepri, Bruno and Lio, Pietro and Scarselli, Franco and Passerini, Andrea},
    journal={Transactions on Machine Learning Research},
    issn={2835-8856},
    year={2023}
}

@inproceedings{naiman2023operator,
    title={An operator theoretic approach for analyzing sequence neural networks},
    author={Naiman, Ilan and Azencot, Omri},
    booktitle={Proceedings of the AAAI conference on artificial intelligence},
    volume={37},
    pages={9268--9276},
    year={2023}
}

@inproceedings{seo2018structured,
    title={Structured sequence modeling with graph convolutional recurrent networks},
    author={Seo, Youngjoo and Defferrard, Micha{\"e}l and Vandergheynst, Pierre and Bresson, Xavier},
    booktitle={Neural Information Processing: 25th International Conference, ICONIP 2018, Siem Reap, Cambodia, December 13-16, 2018, Proceedings, Part I 25},
    pages={362--373},
    year={2018},
    organization={Springer}
}

@inproceedings{viswanath2009evolution,
    title={On the evolution of user interaction in Facebook},
    author={Viswanath, Bimal and Mislove, Alan and Cha, Meeyoung and Gummadi, Krishna P.},
    booktitle={Proceedings of the 2nd ACM workshop on Online social networks},
    pages={37--42},
    year={2009}
}

@inbook{Oettershagen2020dissemination,
    author = {Lutz Oettershagen and Nils M. Kriege and Christopher Morris and Petra Mutzel},
    title = {Temporal Graph Kernels for Classifying Dissemination Processes},
    booktitle = {Proceedings of the 2020 SIAM International Conference on Data Mining (SDM)},
    chapter = {},
    pages = {496-504},
    year={2020},
    publisher={Society for Industrial and Applied Mathematics (SIAM)}
}

@article{brunton2016discovering,
    title={Discovering governing equations from data by sparse identification of nonlinear dynamical systems},
    author={Brunton, Steven L. and Proctor, Joshua L. and Kutz, J. Nathan},
    journal={Proceedings of the national academy of sciences},
    volume={113},
    number={15},
    pages={3932--3937},
    year={2016},
    publisher={National Acad Sciences}
}

@book{Brunton_Kutz_2022, 
    place={Cambridge}, 
    edition={2}, 
    title={Data-Driven Science and Engineering: Machine Learning, Dynamical Systems, and Control}, 
    publisher={Cambridge University Press}, 
    author={Brunton, Steven L. and Kutz, J. Nathan}, year={2022}
}

@article{arbabi2017ergodic,
    author = {Arbabi, Hassan and Mezi\'{c}, Igor},
    title = {Ergodic Theory, Dynamic Mode Decomposition, and Computation of Spectral Properties of the Koopman Operator},
    journal = {SIAM Journal on Applied Dynamical Systems},
    volume = {16},
    number = {4},
    pages = {2096-2126},
    year = {2017},
}

@article{melnyk2020graphkke,
    title={GraphKKE: graph Kernel Koopman embedding for human microbiome analysis},
    author={Melnyk, Kateryna and Klus, Stefan and Montavon, Gr{\'e}goire and Conrad, Tim O.F.},
    journal={Applied Network Science},
    volume={5},
    pages={1--22},
    year={2020},
    publisher={Springer}
}

@article{melnyk2023understanding,
  title={Understanding microbiome dynamics via interpretable graph representation learning},
  author={Melnyk, Kateryna and Weimann, Kuba and Conrad, Tim O.F.},
  journal={Scientific Reports},
  volume={13},
  number={1},
  pages={2058},
  year={2023},
  publisher={Nature Publishing Group UK London}
}

@article{mezic2021koopman,
  title={Koopman operator, geometry, and learning of dynamical systems},
  author={Mezi{\'c}, Igor},
  journal={Not.~Am.~Math.~Soc.},
  volume={68},
  number={7},
  pages={1087--1105},
  year={2021}
}

@article{demo2018pydmd,
  title={PyDMD: Python dynamic mode decomposition},
  author={Demo, Nicola and Tezzele, Marco and Rozza, Gianluigi},
  journal={Journal of Open Source Software},
  volume={3},
  number={22},
  pages={530},
  year={2018}
}

@article{ichinaga2024pydmd,
  title={PyDMD: A Python package for robust dynamic mode decomposition},
  author={Ichinaga, Sara M and Andreuzzi, Francesco and Demo, Nicola and Tezzele, Marco and Lapo, Karl and Rozza, Gianluigi and Brunton, Steven L and Kutz, J Nathan},
  journal={Journal of Machine Learning Research},
  volume={25},
  number={417},
  pages={1--9},
  year={2024}
}

@article{schmid2010dynamic,
  title={Dynamic mode decomposition of numerical and experimental data},
  author={Schmid, Peter J.},
  journal={Journal of fluid mechanics},
  volume={656},
  pages={5--28},
  year={2010},
  publisher={Cambridge University Press}
}

@book{kutz2016dynamic,
  title={Dynamic mode decomposition: data-driven modeling of complex systems},
  author={Kutz, J. Nathan and Brunton, Steven L. and Brunton, Bingni W. and Proctor, Joshua L.},
  year={2016},
  publisher={SIAM}
}

@article{schmid2022dynamic,
  title={Dynamic mode decomposition and its variants},
  author={Schmid, Peter J.},
  journal={Annual Review of Fluid Mechanics},
  volume={54},
  number={1},
  pages={225--254},
  year={2022},
  publisher={Annual Reviews}
}

@article{li2017extended,
  title={Extended dynamic mode decomposition with dictionary learning: A data-driven adaptive spectral decomposition of the Koopman operator},
  author={Li, Qianxiao and Dietrich, Felix and Bollt, Erik M. and Kevrekidis, Ioannis G.},
  journal={Chaos: An Interdisciplinary Journal of Nonlinear Science},
  volume={27},
  number={10},
  year={2017},
  publisher={AIP Publishing}
}

@article{micheli2022discrete,
  title={Discrete-time dynamic graph echo state networks},
  author={Micheli, Alessio and Tortorella, Domenico},
  journal={Neurocomputing},
  volume={496},
  pages={85--95},
  year={2022},
  publisher={Elsevier}
}

@article{isella2011s,
  title={What's in a crowd? Analysis of face-to-face behavioral networks},
  author={Isella, Lorenzo and Stehl{\'e}, Juliette and Barrat, Alain and Cattuto, Ciro and Pinton, Jean-Fran{\c{c}}ois and Van den Broeck, Wouter},
  journal={Journal of theoretical biology},
  volume={271},
  number={1},
  pages={166--180},
  year={2011},
  publisher={Elsevier}
}

@inproceedings{leskovec2009meme,
  title={Meme-tracking and the dynamics of the news cycle},
  author={Leskovec, Jure and Backstrom, Lars and Kleinberg, Jon},
  booktitle={Proceedings of the 15th ACM SIGKDD international conference on Knowledge discovery and data mining},
  pages={497--506},
  year={2009}
}

@article{lusch2018deep,
  title={Deep learning for universal linear embeddings of nonlinear dynamics},
  author={Lusch, Bethany and Kutz, J. Nathan and Brunton, Steven L.},
  journal={Nature communications},
  volume={9},
  number={1},
  pages={4950},
  year={2018},
  publisher={Nature Publishing Group UK London}
}

@inproceedings{mohr2021applications,
  title={Applications of Koopman Mode Analysis to Neural Networks},
  author={Mohr, Ryan and Fonoberova, Maria and Manojlovi{\'c}, Iva and Andrej{\v{c}}uk, Aleksandr and Drma{\v{c}}, Zlatko and Kevrekidis, Yannis and Mezi{\'c}, Igor},
  booktitle={Proceedings of the AAAI 2021 Spring Symposium on Combining Artificial Intelligence and Machine Learning with Physical Sciences},
  year={2021},
  organization={Aachen: CEUR}
}

@article{cini2025graph,
author = {Cini, Andrea and Marisca, Ivan and Zambon, Daniele and Alippi, Cesare},
title = {Graph Deep Learning for Time Series Forecasting},
year = {2025},
issue_date = {December 2025},
publisher = {Association for Computing Machinery},
address = {New York, NY, USA},
volume = {57},
number = {12},
issn = {0360-0300},
url = {https://doi.org/10.1145/3742784},
doi = {10.1145/3742784},
journal = {ACM Comput. Surv.},
month = jul,
articleno = {321},
numpages = {34}
}

@article{cini2023sparse,
  title={Sparse graph learning from spatiotemporal time series},
  author={Cini, Andrea and Zambon, Daniele and Alippi, Cesare},
  journal={Journal of Machine Learning Research},
  volume={24},
  number={242},
  pages={1--36},
  year={2023}
}

@inproceedings{marisca2024graph,
  title     = {Graph-based Forecasting with Missing Data through Spatiotemporal Downsampling},
  author    = {Marisca, Ivan and Alippi, Cesare and Bianchi, Filippo Maria},
  booktitle = {Proceedings of the 41st International Conference on Machine Learning},
  pages     = {34846--34865},
  year      = {2024},
  volume    = {235},
  series    = {Proceedings of Machine Learning Research},
  publisher = {PMLR}
}

@article{cini2024graph,
title        = {{Graph-based Time Series Clustering for End-to-End Hierarchical Forecasting}},
author       = {Cini, Andrea and Mandic, Danilo and Alippi, Cesare},
journal      = {International Conference on Machine Learning},
year         = 2024
}

@inproceedings{cini2023scalable,
  title={Scalable spatiotemporal graph neural networks},
  author={Cini, Andrea and Marisca, Ivan and Bianchi, Filippo Maria and Alippi, Cesare},
  booktitle={Proceedings of the AAAI conference on artificial intelligence},
  volume={37},
  pages={7218--7226},
  year={2023}
}

@article{hassija2024interpreting,
  title={Interpreting black-box models: a review on explainable artificial intelligence},
  author={Hassija, Vikas and Chamola, Vinay and Mahapatra, Atmesh and Singal, Abhinandan and Goel, Divyansh and Huang, Kaizhu and Scardapane, Simone and Spinelli, Indro and Mahmud, Mufti and Hussain, Amir},
  journal={Cognitive Computation},
  volume={16},
  number={1},
  pages={45--74},
  year={2024},
  publisher={Springer}
}

@inproceedings{
kipf2017semisupervised,
title={Semi-Supervised Classification with Graph Convolutional Networks},
author={Thomas N. Kipf and Max Welling},
booktitle={International Conference on Learning Representations},
year={2017},
}

@inproceedings{zhang2020spatio,
  title={Spatio-temporal graph structure learning for traffic forecasting},
  author={Zhang, Qi and Chang, Jianlong and Meng, Gaofeng and Xiang, Shiming and Pan, Chunhong},
  booktitle={Proceedings of the AAAI conference on artificial intelligence},
  volume={34},
  pages={1177--1185},
  year={2020}
}

@inproceedings{jain2016structural,
  title={Structural-rnn: Deep learning on spatio-temporal graphs},
  author={Jain, Ashesh and Zamir, Amir R. and Savarese, Silvio and Saxena, Ashutosh},
  booktitle={Proceedings of the ieee conference on computer vision and pattern recognition},
  pages={5308--5317},
  year={2016}
}

@article{fritz2022combining,
  title={Combining graph neural networks and spatio-temporal disease models to improve the prediction of weekly COVID-19 cases in Germany},
  author={Fritz, Cornelius and Dorigatti, Emilio and R{\"u}gamer, David},
  journal={Scientific Reports},
  volume={12},
  number={1},
  pages={3930},
  year={2022},
  publisher={Nature Publishing Group UK London}
}

@inproceedings{deng2019learning,
  title={Learning dynamic context graphs for predicting social events},
  author={Deng, Songgaojun and Rangwala, Huzefa and Ning, Yue},
  booktitle={Proceedings of the 25th ACM SIGKDD International Conference on Knowledge Discovery \& Data Mining},
  pages={1007--1016},
  year={2019}
}

@article{verdone2024explainable,
  title={Explainable Spatio-Temporal Graph Neural Networks for multi-site photovoltaic energy production},
  author={Verdone, Alessio and Scardapane, Simone and Panella, Massimo},
  journal={Applied Energy},
  volume={353},
  pages={122151},
  year={2024},
  publisher={Elsevier}
}

@inproceedings{altieri2023explainable,
  title={Explainable spatio-temporal graph modeling},
  author={Altieri, Massimiliano and Ceci, Michelangelo and Corizzo, Roberto},
  booktitle={International Conference on Discovery Science},
  pages={174--188},
  year={2023},
  organization={Springer}
}

@inproceedings{tang2023explainable,
  title={Explainable Spatio-Temporal Graph Neural Networks},
  author={Tang, Jiabin and Xia, Lianghao and Huang, Chao},
  booktitle={Proceedings of the 32nd ACM International Conference on Information and Knowledge Management},
  pages={2432--2441},
  year={2023}
}

@article{chen2025explainable,
  title={An explainable spatio-temporal graph convolutional network for the biomarkers identification of ADHD},
  author={Chen, Longyun and Yang, Yuhui and Yu, Aiju and Guo, Shuo and Ren, Kai and Liu, Qinfang and Qiao, Chen},
  journal={Biomedical Signal Processing and Control},
  volume={99},
  pages={106913},
  year={2025},
  publisher={Elsevier}
}

@inproceedings{Simonyan14a,
  author       = {Karen Simonyan and Andrea Vedaldi and Andrew Zisserman},
  title        = {Deep Inside Convolutional Networks: Visualising Image Classification Models and Saliency Maps},
  booktitle    = {Workshop at International Conference on Learning Representations},
  year         = {2014},
}

@inproceedings{
li2018diffusion,
title={Diffusion Convolutional Recurrent Neural Network: Data-Driven Traffic Forecasting},
author={Yaguang Li and Rose Yu and Cyrus Shahabi and Yan Liu},
booktitle={International Conference on Learning Representations},
year={2018},
url={https://openreview.net/forum?id=SJiHXGWAZ},
}

@article{kakkad2023survey,
  title={A Survey on Explainability of Graph Neural Networks},
  author={Kakkad, Jaykumar and Jannu, Jaspal and Sharma, Kartik and Aggarwal, Charu and Medya, Sourav},
  journal={Bulletin of the IEEE Computer Society Technical Committee on Data Engineering},
  year={2023}
}

@article{spinelli2022meta,
  title={A meta-learning approach for training explainable graph neural networks},
  author={Spinelli, Indro and Scardapane, Simone and Uncini, Aurelio},
  journal={IEEE Transactions on Neural Networks and Learning Systems},
  volume={35},
  number={4},
  pages={4647--4655},
  year={2022},
  publisher={IEEE}
}

@inproceedings{lucic2022cf,
  title={Cf-gnnexplainer: Counterfactual explanations for graph neural networks},
  author={Lucic, Ana and Ter Hoeve, Maartje A and Tolomei, Gabriele and De Rijke, Maarten and Silvestri, Fabrizio},
  booktitle={International Conference on Artificial Intelligence and Statistics},
  pages={4499--4511},
  year={2022},
  organization={PMLR}
}

@inproceedings{
    azzolin2023global,
    title={Global Explainability of {GNN}s via Logic Combination of Learned Concepts},
    author={Steve Azzolin and Antonio Longa and Pietro Barbiero and Pietro Lio and Andrea Passerini},
    booktitle={The Eleventh International Conference on Learning Representations },
    year={2023},
    url={https://openreview.net/forum?id=OTbRTIY4YS}
}

@inproceedings{
    wang2023gnninterpreter,
    title={{GNNI}nterpreter: A Probabilistic Generative Model-Level Explanation for Graph Neural Networks},
    author={Xiaoqi Wang and Han Wei Shen},
    booktitle={The Eleventh International Conference on Learning Representations },
    year={2023},
    url={https://openreview.net/forum?id=rqq6Dh8t4d}
}

@article{ying2019gnnexplainer,
  title={Gnnexplainer: Generating explanations for graph neural networks},
  author={Ying, Zhitao and Bourgeois, Dylan and You, Jiaxuan and Zitnik, Marinka and Leskovec, Jure},
  journal={Advances in neural information processing systems},
  volume={32},
  year={2019}
}

@article{luo2020parameterized,
  title={Parameterized explainer for graph neural network},
  author={Luo, Dongsheng and Cheng, Wei and Xu, Dongkuan and Yu, Wenchao and Zong, Bo and Chen, Haifeng and Zhang, Xiang},
  journal={Advances in neural information processing systems},
  volume={33},
  pages={19620--19631},
  year={2020}
}

@inproceedings{guerra2023explainability,
  title={Explainability in subgraphs-enhanced Graph Neural Networks},
  author={Guerra, Michele and Spinelli, Indro and Scardapane, Simone and Bianchi, Filippo Maria},
  booktitle={Proceedings of the Northern Lights Deep Learning Workshop},
  volume={4},
  year={2023}
}

@article{bianchi2021reservoir,
  title={Reservoir computing approaches for representation and classification of multivariate time series},
  author={Bianchi, Filippo Maria and Scardapane, Simone and L{\o}kse, Sigurd and Jenssen, Robert},
  journal={IEEE Transactions on Neural Networks and Learning Systems},
  volume={32},
  number={5},
  pages={2169--2179},
  year={2021},
  publisher={IEEE}
}

@inproceedings{spinelli2023combining,
author = {Spinelli, Indro and Guerra, Michele and Bianchi, Filippo Maria and Scardapane, Simone},
year = {2023},
month = {01},
pages = {229-234},
title = {Combining Stochastic Explainers and Subgraph Neural Networks can Increase Expressivity and Interpretability},
doi = {10.14428/esann/2023.ES2023-13},
booktitle = {ESANN 2023 proceedings},
}

@article{wu2020comprehensive,
  title={A comprehensive survey on graph neural networks},
  author={Wu, Zonghan and Pan, Shirui and Chen, Fengwen and Long, Guodong and Zhang, Chengqi and Yu, Philip S},
  journal={IEEE transactions on neural networks and learning systems},
  volume={32},
  number={1},
  pages={4--24},
  year={2020},
  publisher={IEEE}
}

@article{khemani2024review,
  title={A review of graph neural networks: concepts, architectures, techniques, challenges, datasets, applications, and future directions},
  author={Khemani, Bharti and Patil, Shruti and Kotecha, Ketan and Tanwar, Sudeep},
  journal={Journal of Big Data},
  volume={11},
  number={1},
  pages={18},
  year={2024},
  publisher={Springer}
}

@inproceedings{
    azzolin2025beyond,
    title={Beyond Topological Self-Explainable {GNN}s: A Formal Explainability Perspective},
    author={Steve Azzolin and Sagar Malhotra and Andrea Passerini and Stefano Teso},
    booktitle={Forty-second International Conference on Machine Learning},
    year={2025},
    url={https://openreview.net/forum?id=mkqcUWBykZ}
}

@inproceedings{
    azzolin2026gnn,
    title={{GNN} Explanations that do not Explain and How to find Them},
    author={Steve Azzolin and Stefano Teso and Bruno Lepri and Andrea Passerini and Sagar Malhotra},
    booktitle={The Fourteenth International Conference on Learning Representations},
    year={2026},
    url={https://openreview.net/forum?id=HBcgLe6NZD}
}

@article{longa2025explaining,
  title={Explaining the explainers in graph neural networks: a comparative study},
  author={Longa, Antonio and Azzolin, Steve and Santin, Gabriele and Cencetti, Giulia and Li{\`o}, Pietro and Lepri, Bruno and Passerini, Andrea},
  journal={ACM Computing Surveys},
  volume={57},
  number={5},
  pages={1--37},
  year={2025},
  publisher={ACM New York, NY}
}

@inproceedings{
    shi2025when,
    title={When Graph Neural Networks Meet Dynamic Mode Decomposition},
    author={Dai Shi and Lequan Lin and Andi Han and Zhiyong Wang and Yi Guo and Junbin Gao},
    booktitle={The Thirteenth International Conference on Learning Representations},
    year={2025},
    url={https://openreview.net/forum?id=duGygkA3QR}
}

@article{
    han2024from,
    title={From Continuous Dynamics to Graph Neural Networks: Neural Diffusion and Beyond},
    author={Andi Han and Dai Shi and Lequan Lin and Junbin Gao},
    journal={Transactions on Machine Learning Research},
    issn={2835-8856},
    year={2024},
    url={https://openreview.net/forum?id=fPQSxjqa2o},
    note={Survey Certification}
}

@article{
    Gravina2025, 
    title={On Oversquashing in Graph Neural Networks Through the Lens of Dynamical Systems}, 
    volume={39}, 
    url={https://ojs.aaai.org/index.php/AAAI/article/view/33858},
    DOI={10.1609/aaai.v39i16.33858}, 
    number={16}, 
    journal={Proceedings of the AAAI Conference on Artificial Intelligence}, 
    author={Gravina, Alessio and Eliasof, Moshe and Gallicchio, Claudio and Bacciu, Davide and Schönlieb, Carola-Bibiane}, 
    year={2025}, 
    month={Apr.}, 
    pages={16906-16914} 
}

@inproceedings{wu2019graph,
    author = {Wu, Zonghan and Pan, Shirui and Long, Guodong and Jiang, Jing and Zhang, Chengqi},
    title = {Graph wavenet for deep spatial-temporal graph modeling},
    year = {2019},
    isbn = {9780999241141},
    publisher = {AAAI Press},
    booktitle = {Proceedings of the 28th International Joint Conference on Artificial Intelligence},
    pages = {1907–1913},
    numpages = {7},
    location = {Macao, China},
    series = {IJCAI'19}
}

@inproceedings{fothergill2012instructing,
  title={Instructing people for training gestural interactive systems},
  author={Fothergill, Simon and Mentis, Helena and Kohli, Pushmeet and Nowozin, Sebastian},
  booktitle={Proceedings of the SIGCHI conference on human factors in computing systems},
  pages={1737--1746},
  year={2012}
}

@article{fontanesi2025bridging,
  title={Bridging XAI and spectral analysis to investigate the inductive biases of deep graph networks},
  author={Fontanesi, Michele and Micheli, Alessio and Podda, Marco and Tortorella, Domenico},
  journal={Machine Learning},
  volume={114},
  number={11},
  pages={257},
  year={2025},
  publisher={Springer}
}

@article{agarwal2023evaluating,
  title={Evaluating explainability for graph neural networks},
  author={Agarwal, Chirag and Queen, Owen and Lakkaraju, Himabindu and Zitnik, Marinka},
  journal={Scientific Data},
  volume={10},
  number={1},
  pages={144},
  year={2023},
  publisher={Nature Publishing Group UK London}
}

@inproceedings{fontanesi2024explaining,
  title={Explaining graph classifiers by unsupervised node relevance attribution},
  author={Fontanesi, Michele and Micheli, Alessio and Podda, Marco},
  booktitle={World Conference on Explainable Artificial Intelligence},
  pages={63--74},
  year={2024},
  organization={Springer}
}

@article{
    dileo2025evaluating,
    title={Evaluating explainability techniques on discrete-time graph neural networks},
    author={Manuel Dileo and Matteo Zignani and Sabrina Tiziana Gaito},
    journal={Transactions on Machine Learning Research},
    issn={2835-8856},
    year={2025},
    url={https://openreview.net/forum?id=JzmXo0rfry},
    note={}
}

@inproceedings{
    azzolin2025reconsidering,
    title={Reconsidering Faithfulness in Regular, Self-Explainable and Domain Invariant {GNN}s},
    author={Steve Azzolin and Antonio Longa and Stefano Teso and Andrea Passerini},
    booktitle={The Thirteenth International Conference on Learning Representations},
    year={2025},
    url={https://openreview.net/forum?id=kiOxNsrpQy}
}

@article{klus2018tensor,
  title={Tensor-based dynamic mode decomposition},
  author={Klus, Stefan and Gel{\ss}, Patrick and Peitz, Sebastian and Sch{\"u}tte, Christof},
  journal={Nonlinearity},
  volume={31},
  number={7},
  pages={3359--3380},
  year={2018},
  publisher={IOP Publishing}
}

@article{oseledets2011tensor,
  title={Tensor-train decomposition},
  author={Oseledets, Ivan V},
  journal={SIAM Journal on Scientific Computing},
  volume={33},
  number={5},
  pages={2295--2317},
  year={2011},
  publisher={SIAM}
}

@inproceedings{
    wagner2026formally,
    title={Formally Exploring Time-Series Anomaly Detection Evaluation Metrics},
    author={Dennis Wagner and Arjun Nair and Billy Joe Franks and Justus Arweiler and Aparna Muraleedharan and Indra Jungjohann and Fabian Hartung and Andriy Balinskyy and Saurabh Varshneya and Mayank Chetan Ahuja and Nabeel Hussain Syed and Mayank Nagda and Philipp Liznerski and Steffen Reithermann and Maja Rudolph and Sebastian Josef Vollmer and Ralf Schulz and Torsten Katz and Stephan Mandt and Michael Bortz and Heike Leitte and Daniel Neider and Jakob Burger and Fabian Jirasek and Hans Hasse and Sophie Fellenz and Marius Kloft},
    booktitle={The 29th International Conference on Artificial Intelligence and Statistics},
    year={2026},
    url={https://openreview.net/forum?id=INJj1SB5Uw}
}

@article{kim2022towards, 
    title={Towards a Rigorous Evaluation of Time-Series Anomaly Detection}, 
    volume={36}, url={https://ojs.aaai.org/index.php/AAAI/article/view/20680}, 
    DOI={10.1609/aaai.v36i7.20680}, 
    number={7}, 
    journal={Proceedings of the AAAI Conference on Artificial Intelligence}, 
    author={Kim, Siwon and Choi, Kukjin and Choi, Hyun-Soo and Lee, Byunghan and Yoon, Sungroh}, 
    year={2022}, 
    month={Jun.}, 
    pages={7194-7201} 
}
